\documentclass{article} 
\usepackage{iclr2025_conference,times}
\usepackage{xspace,placeins}
\usepackage{siunitx,tcolorbox,colortbl,array}
\usepackage{graphicx,booktabs,listings}
\usepackage[utf8]{inputenc}
\usepackage{multicol,tabularx,geometry,parskip}
\usepackage{amssymb,pifont,float}
\usepackage{fancyvrb,inconsolata,multirow}
\usepackage{subcaption,etoolbox}
\usepackage{algorithm,algpseudocode}
\tcbuselibrary{skins}

\usepackage{amsmath,amsfonts,bm}

\def\eqref#1{equation~\ref{#1}}

\def\1{\bm{1}}

\DeclareMathAlphabet{\mathsfit}{\encodingdefault}{\sfdefault}{m}{sl}
\SetMathAlphabet{\mathsfit}{bold}{\encodingdefault}{\sfdefault}{bx}{n}

\newcommand{\ix}[1]{{#1}^{(i)}}
\newcommand{\obs}[2]{\mathbf{n}_{#1}\{#2\}}
\newcommand{\ind}[1]{\mathbf{1}\{#1\}}

\newcommand{\api}{Q}
\newcommand{\local}{P}
\newcommand{\prior}{\pi}
\newcommand{\decisionrule}{\delta}
\newcommand{\sampleapi}{\mathcal D_\api}
\newcommand{\samplelocal}{\mathcal D_\local}
\newcommand{\samplelocalone}{\mathcal D_{\local_1}}
\newcommand{\samplelocaltwo}{\mathcal D_{\local_2}}

\newcommand{\llama}{Llama\xspace}
\newcommand{\mistral}{Mistral\xspace}
\newcommand{\half}{{fp16}\xspace}
\newcommand{\inteight}{{int8}\xspace}
\newcommand{\nf}{{nf4}\xspace}
\newcommand{\full}{{fp32}\xspace}
\newcommand{\humaneval}{{HumanEval}\xspace}
\newcommand{\ultrachat}{{UltraChat}\xspace}

\newcommand{\eg}{e.g.,\xspace}
\newcommand{\ie}{i.e.,\xspace}
\newcommand{\iid}{\textit{i.i.d.}\xspace}

\definecolor{ForestGreen}{RGB}{34,139,34}
\definecolor{Mahogany}{RGB}{192,64,0}
\newcommand{\cmark}{{\color{ForestGreen}\ding{51}}}%
\newcommand{\xmark}{{\color{Mahogany}\ding{55}}}%

\newcounter{samplebox}
\renewcommand{\thesamplebox}{\arabic{samplebox}}
\newtcolorbox[use counter=samplebox]{samplebox}[2][]{
  enhanced,
  colback=white, 
  colframe=black, 
  boxrule=0.5pt, 
  arc=5pt, 
  outer arc=5pt, 
  title={Box~\thesamplebox: #2}, 
  fonttitle=\color{white},
  colbacktitle=black, 
  fontupper=\small,
  #1,
}

\newcommand{\tightparagraph}[1]{\vspace{0.05in}\noindent\textbf{#1}}
\newenvironment{verbatimcode}
  {\VerbatimEnvironment
   \fontsize{8}{10}\selectfont
   \ttfamily
   \begin{Verbatim}[formatcom=\color{black}]}
  {\end{Verbatim}}

\usepackage[colorlinks = true,
            linkcolor = blue,
            urlcolor  = blue,
            citecolor = blue,
            anchorcolor = blue]{hyperref}
\usepackage{url}

\newtoggle{showcomments}
\togglefalse{showcomments}
\iftoggle{showcomments}{
    \newcommand{\pl}[1]{\textcolor{red}{[PL: #1]}}
    \newcommand{\ym}[1]{\textcolor{red}{[YM: #1]}}
    \newcommand{\nm}[1]{\textcolor{red}{[NM: #1]}}
    \newcommand{\sg}[1]{\textcolor{red}{[SG: #1]}}
    \newcommand{\todo}[1]{\textcolor{red}{[\textbf{Irena:} #1]}}
}{
    \newcommand{\pl}[1]{}
    \newcommand{\ym}[1]{}
    \newcommand{\nm}[1]{}
    \newcommand{\sg}[1]{}
    \newcommand{\todo}[1]{}
}

\title{Model Equality Testing:\\Which Model Is This API Serving?}

\author{Irena Gao, Percy Liang, Carlos Guestrin\\\texttt{irena@cs.stanford.edu, pliang@cs.stanford.edu, guestrin@stanford.edu}\\Stanford University}

\iclrfinalcopy 
\begin{document}

\maketitle

\begin{abstract}
Users often interact with large language models through black-box inference APIs, both for closed- and open-weight models 
(\eg \llama models are popularly accessed via Amazon Bedrock and Azure AI Studios).
In order to cut costs or add functionality, API providers may quantize, watermark, or finetune the underlying model, changing the output distribution --- possibly without notifying users. 
We formalize detecting such distortions as Model Equality Testing, a two-sample testing problem, 
where the user collects samples from the API and a reference distribution, and conducts a statistical test to see if the two distributions are the same. 
We find that tests based on the Maximum Mean Discrepancy between distributions are powerful for this task:
a test built on a simple string kernel achieves a median of 77.4\% power against a range of distortions, using an average of just 10 samples per prompt.
We then apply this test to commercial inference APIs from Summer 2024 for four \llama models,
finding that 11 out of 31 endpoints serve different distributions than reference weights released by Meta.

\end{abstract}


\section{Introduction}\label{sec:intro}
\begin{figure}[tb]
    \centering
    \includegraphics[width=\textwidth]{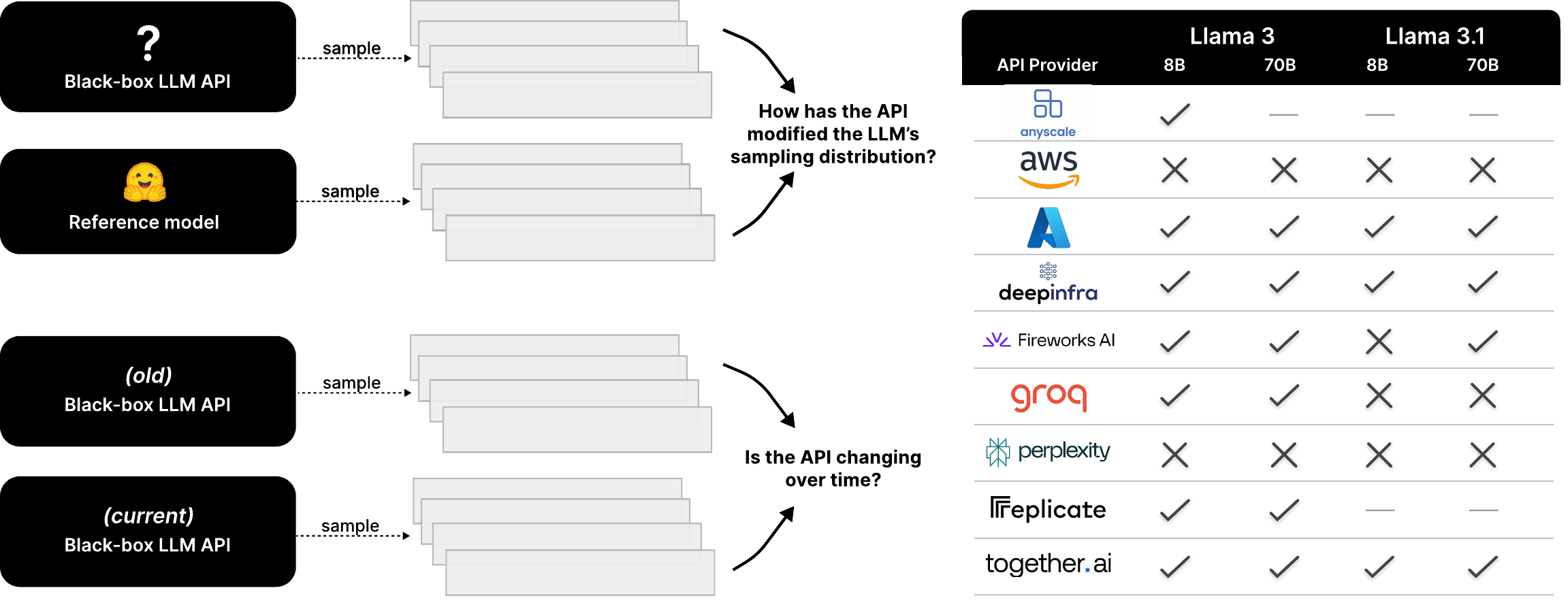}
    \caption{
        \textit{(Left)} We formalize auditing black-box language model inference APIs as Model Equality Testing. This enables us to assess an API's faithfulness to a reference distribution and its stability over time.
        \textit{(Right)} We evaluate candidate tests and apply the most powerful one to \llama model APIs from Summer 2024, finding that 11 of 31 endpoints deviate from reference weights released by Meta.
        \vspace{-1.1em}
    }
    \label{fig:hero}
\end{figure}

Since running a large language model requires compute and technical expertise,
many users rely on black-box APIs to handle inference.
This applies to both closed-weight models, like GPT and Claude, and open-weight ones:
\eg \href{https://aws.amazon.com/bedrock/}{Amazon Bedrock}, \href{https://azure.microsoft.com/en-us/products/machine-learning/generative-ai}{Microsoft Azure}, and 
the seven other companies in Figure \ref{fig:hero} all compete to offer \llama models as a service.
While users can sample from black-box APIs, they have little to no insight into the underlying implementation of the model, including into questions like:
\begin{enumerate}
    \item \textbf{How has the API modified the language model's distribution?} 
    To drive down costs, API providers may quantize or prune large model weights; 
    they may also watermark outputs or incorrectly implement some decoding parameters.
    These changes distort the resulting distribution of completions.
    The problem is when such distortions are undisclosed: 
    users assume that calling a third-party API is exactly equivalent to working with the original model.
    For example, benchmarks like HELM \citep{liang2022holistic} evaluate models through third-party APIs, 
    but quantized or watermarked models may be less capable than the intended model. 
    \item \textbf{Is the API changing over time?} Language model inference endpoints may also drift over time without notifying users \citep{chen2023chatgpt,eyuboglu2024model},
    \eg due to finetuning or updates to the inference stack.
    Unstable APIs affect research reproducibility \citep{pozzobon2023challenges} and can disrupt user productivity in human-AI teams \citep{bansal2019updates}.
\end{enumerate}
Under the status quo, neither users nor regulators have a way to rigorously answer these questions for themselves.\footnote{
These problems are already experienced by users: \eg see threads \href{https://www.reddit.com/r/OpenAI/comments/zoez3p/is_it_just_me_or_is_chatgpt_getting_worse_by_the/}{1}, \href{https://www.reddit.com/r/LocalLLaMA/comments/1ebzd5a/warning_the_quality_of_hosted_llama_31_may_vary/}{2} and \href{https://x.com/pandaashwinee/status/1816966288905998829?s=46&t=fmzCzaSrcpp2jNTx6-oaKw}{3}. 
}
These concerns are important to address: 
tens of thousands of developers already rely on black-box inference APIs for applications \citep{amazonbedrockusage}, 
and this dependence will increase as LLMs --- and the corresponding infrastructure costs for hosting --- grow larger.
For example, most users \textit{must} rely on third-party APIs to use \llama 3.1 405B because of its size.

The current approach to this problem is for an outside auditor to monitor APIs' accuracies on multiple-choice or short-answer benchmarks \citep{artificialanalysis};  
these studies typically decode from the language model greedily.
Such audits can be a poor match for user needs.
Greedy decoding only checks that the modes of the next-token distributions match, rather than the overall distribution over completions,
which is problematic because users often \textit{sample} from models.
Short-answer benchmarks cover only a small slice of possible prefixes, which may significantly differ from a particular user's task:
\eg common applications like code generation, dialogue, and summarization are longform tasks.
Ideally, users could personally audit APIs on their custom tasks.
Such a method should be sample-efficient, apply to tasks without automated verifiers, and assess with confidence whether the \textit{overall} distribution of completions has shifted in a statistically significant way.

We provide such a method.
Suppose a user wishes to audit a an API on their task of interest.
The user collects two samples: one from a reference distribution $\local$ and one from the test API's distribution $\api$.
For example, to answer if an API has modified the distribution of an open-weight model, $\local$ might be from reference model weights released on Hugging Face.
To answer if the API is changing over time, $\local$ might be from the API at an earlier point in time (Figure \ref{fig:hero} left).
The user then conducts a two-sample test for whether $\local = \api$ to examine if the API's distribution is statistically indistinguishable from the reference. 

Our setting is challenging because the distributions being compared are high-dimensional:
they are defined over multi-token completions from large vocabularies.
Two-sample kernel tests based on estimating the Maximum Mean Discrepancy (MMD) between $\local$ and $\api$ \citep{gretton2012kernel} are flexible tools for this setting,
as they allow us to specify a featurization to reduce dimensionality. 
We find that a simple string kernel based on the Hamming distance between completions is particularly sample-efficient.
In simulations (\S \ref{sec:simulations}), this test achieves a median of 77.4\% power against a wide range of distortions --- \eg quantization, watermarking, and finetuning --- using an average of just 10 samples per prompt for distributions over 20--25 prompts. 
We then apply this test to nine commercial inference API providers across four \llama models (\S \ref{sec:apis}, Figure \ref{fig:hero} right).
Our test flags 11 out of these 31 endpoints, with each audit costing less than \$1.

Because our test statistic is an estimate of a distance, we also explore how this machinery can quantify statistical distances between black-box endpoints.
In \S\ref{sec:model_model}, we estimate pairwise distances between the output distributions of 13 language models --- without requiring log probability access --- and find that models within 
the same family (\eg the Llama family or GPT-3.5 family) output more similar distributions than models within the same size range (\eg 7B or 70B models).
In \S\ref{sec:apis}, we estimate the effect size of deviations between API endpoints and reference weights, finding that some implementations are further from the reference weights than if the provider had substituted in an entirely different language model.


\paragraph{Summary of contributions.} 
We unify several API auditing tasks under the formalization of Model Equality Testing, a two-sample distribution testing problem,
and empirically validate kernel-based tests for this problem.
We then apply this test to audit popular commercial inference APIs. 
To enable users to audit APIs for custom applications, we open-source a Python package.
We also encourage future research in Model Equality Testing by releasing a dataset of 1 million LLM completions from five models.\footnote{Package, experiment code, and dataset: \url{https://github.com/i-gao/model-equality-testing}.}

\section{The Model Equality Testing problem}\label{sec:problem}
Suppose an auditor is interested in a task parameterized by a distribution $\prior$ over $m$ prompts and a maximum completion length of $L$ tokens.
The auditor has sample access to a reference distribution $\local$ and API distribution $\api$, both operating on the same vocabulary $\mathcal V$ with the same decoding parameters.
The auditor samples $N$ prompt-completion pairs\footnote{In Appendix \ref{app:asymmetric}, we discuss how to extend this setup to unequal sample sizes.} $z := (x, y)$ from each distribution:
\begin{equation}
\begin{aligned}
    \samplelocal &:= \{\ix{z} : \ix{x} \sim \prior, \ix{y} \sim \local(\cdot \mid \ix{x}) \}_{i=1}^N, \\
    \sampleapi &:= \{\ix{z} : \ix{x} \sim \prior, \ix{y} \sim \api(\cdot \mid \ix{x}) \}_{i=1}^N.
\end{aligned}
\label{eqn:samples}
\end{equation}
We wish to use these samples to test if $\local = \api$, \ie distinguish between the hypotheses
\begin{equation}
\begin{aligned}
H_0:& \quad\prior(x) \local(y \mid x) = \prior(x) \api(y \mid x), \\
H_1:& \quad\prior(x) \local(y \mid x) \neq \prior(x) \api(y \mid x).
\end{aligned}
\label{eqn:hypotheses}
\end{equation}
We require that the Type-1 error rate is controlled at $\alpha$.
A good test will maximize power against unknown $\api$ and generalize across several language models and prompt distributions $\prior$.
We are particularly interested in \textit{sample-efficient} tests that are cheap to run:
such tests are powerful when $N$ is small compared to the size of the vocabulary $|\mathcal V|$ and the maximum completion length $L$.\footnote{As a concrete example, \llama-3 uses a vocabulary size of $|\mathcal V| = \num{128256}$, and users often sample $L=250$ tokens for longform generation tasks.}
The latter parameters modulate the size of the space that the joint distributions are defined over:
the set of all prompt-completion pairs has size $m|\mathcal V|^L$, where $m$ is the number of prompts captured in $\prior$.
Effective tests must navigate this high-dimensional space well.
Fortunately, we expect the distributions in practice to be lower-dimensional, as language typically only places significant mass on a small number of tokens at each position.

Throughout the paper, we will use $\obs{s}{S}$ to denote the count of object $s$ in string or sample $S$, $\prior \local$ to denote the joint distribution of prompts and completions under $\local$, and $\prior \api$ to denote the joint distribution under $\api$.

\section{Method}\label{sec:method}
To tackle the problem, we employ a two-sample kernel test from \citet{gretton2012kernel}.
This test uses samples $\samplelocal$ and $\sampleapi$ to estimate the Maximum Mean Discrepancy (MMD) between $\local$ and $\api$, which is a measure of the distance between the two distributions.
Intuitively, if the estimated MMD is large, we reject the null hypothesis that $\local = \api$.

The MMD is defined with respect to a unit-norm kernel function $k$ and its associated feature map $\phi$.
For our two joint distributions $\prior\local$ and $\prior\api$, the MMD is defined as the squared distance between the expected features from each distribution:
\begin{equation}
    \begin{aligned}
    \text{MMD}_k\left(\prior \local, \prior \api\right)
    &= \left\| \mathbb E_{z \sim \prior \local}\left[\phi(z)\right] - \mathbb E_{z \sim \prior \api}\left[\phi(z)\right] \right\|^2,\\
    &= \mathbb E_{z, z' \sim \prior\local}\left[k(z, z')\right] + \mathbb E_{z, z' \sim \prior \api}\left[k(z, z')\right] - 2\mathbb E_{z \sim \prior\local, z' \sim \prior\api}\left[k(z, z')\right] .
    \end{aligned}
    \label{eqn:mmd}
\end{equation}
For simplicity, we select kernels of the form $k(z,z') = \ind{x = x'} \tilde k(y, y')$, where $\tilde k$ is a prompt-agnostic kernel over completions.
Then $\text{MMD}_k(\prior\local, \prior\api) = \mathbb E_{x \sim \prior} \left[ \text{MMD}_{\tilde k}\left(\local (y \mid x), \api (y \mid x)\right) \right]$.

To conduct a two-sample test with samples $\sampleapi$ and $\samplelocal$, the test statistic is the empirical estimator of (\ref{eqn:mmd}):
\begin{equation}
    \begin{aligned}
    \widehat{\text{MMD}}(\samplelocal, \sampleapi)
    &= \frac{1}{N(N-1)} 
    \left[
        \sum_{z, z' \in \samplelocal} k(z, z')
        + \sum_{z, z' \in \sampleapi} k(z, z')
    \right]
    - \frac{2}{N^2} \sum_{z \in \samplelocal}\sum_{z' \in \sampleapi} k(z, z').
    \end{aligned}
    \label{eqn:mmd_estimator}
\end{equation}
We can compute p-values by simulating the test statistic's distribution under the null, \ie by repeatedly sampling both $\sampleapi$ and $\samplelocal$ from $\local$ and computing (\ref{eqn:mmd_estimator}).
Alternatively, to avoid drawing extra samples from $\local$, we can use the permutation procedure \citep{lehmann1986testing}, at a potential cost to power.
This procedure repeatedly shuffles samples between $\sampleapi$ and $\samplelocal$ to recompute the test statistic (Appendix \ref{app:pvalues}).

\paragraph{Kernel choice.}
The choice of kernel $(k, \phi)$ determines the test's semantics and power.
For example, setting $\phi(z)$ to be an indicator of whether $y$ passes an automated verifier for $x$, when one is available, leads to rejecting the null when $\local, \api$ result in different task accuracies.
However, because $\phi$ is relatively coarse, the MMD may be zero even when $\local \neq \api$, limiting the test's power.
At the other extreme are \textit{universal} kernels, which guarantee that the MMD is zero if and only if $\local = \api$ \citep{gretton2012kernel}.
One such universal kernel for strings is the computationally expensive all-substrings kernel \citep{borgwardt2006integrating}:
\begin{equation}
    \tilde k_\text{all}(y, y') = \sum_{s \in \mathcal V^{\le L}} \obs{s}{y} \cdot \obs{s}{y'},
    \label{eqn:all_substrings}
\end{equation}
where $\obs{s}{y}$ is the number of times $s$ appears in $y$. Another universal kernel is the one-hot kernel: 
\begin{equation}
    \tilde k_\text{one-hot}(y, y') = \ind{y = y'},
    \label{eqn:one_hot}
\end{equation}
which results in a classical two-sample multinomial test between the joint distributions.
While universal kernels can eventually detect differences between any $\local, \api$ with enough samples,
they may have low power in the small-sample regime.
For example, the one-hot MMD measures if there are more exact match collisions within $\samplelocal$ or $\sampleapi$ than between them,
but in small samples, we may see no duplicate completions at all.

We posit that other string kernels, though not universal, provide more powerful features for testing with small samples.
Specifically, we investigate a fast kernel related to the Hamming distance between completions:
\begin{equation}
    \tilde k_\text{hamming}(y, y') = \sum_{i=1}^{L} \ind{y_i = y'_i},
    \label{eqn:hamming}
\end{equation}
where $y$ shorter than $L$ is right-padded with a special token.
Intuitively, a test based on this kernel rejects if a significantly larger number of substitutions are needed to align completions between $\samplelocal$ and $\sampleapi$ than within each sample.
The associated Hamming MMD is a pseudo-metric, as it is zero when $\local = \api$ and obeys the triangle inequality, but may not separate all distributions (Appendix \ref{app:metrics}).
Despite this limitation, we find in the following sections that this kernel is empirically effective and well-suited to common distortions we encounter with language models: quantization, watermarking, finetuning, and related distortions tend to result in detectable inter-sample Hamming distances.

\section{Evaluating tests in simulations}\label{sec:simulations}
In this section, we evaluate our test's power using different kernels at checking equivalence between known pairs of distributions.
Specifically, we evaluate if tests can detect when a language model has been quantized or watermarked (\S\ref{sec:quantization_watermarking}), finetuned (\S\ref{sec:finetuning}), or swapped out for a different model altogether (\S\ref{sec:model_model}).

All experiments in this section are run on a longform language modeling task.
The prompt distribution $\prior$ is a uniform distribution over $m=25$ random 100-character strings sampled from English, German, Spanish, French, and Russian Wikipedia (Box \ref{box:wikipedia}).
The maximum completion length is $L=50$, and we sample using temperature 1.
Power is computed from \num{100} Monte Carlo simulations.
We estimate p-values by simulating the empirical distribution of the test statistic under the null \num{1000} times; in Appendix \ref{app:additional_results}, we validate that the permutation procedure results in the same trends.

\begin{samplebox}[label=box:wikipedia]{Sample prompt for the Wikipedia language modeling task}
Continue the paragraph. Do not output anything except the continuation to the paragraph. Start the continuation immediately.\\
``The British Arab Commercial Bank PLC (BACB) is an international wholesale bank incorporated in the U...''
\end{samplebox}

To evaluate tests' generalization across prompts, we repeat power experiments over ten different prompt distributions, where we resample 25 Wikipedia strings for each $\prior$.
All tests are conducted at a significance level of $\alpha=0.05$.
Additional details can be found in Appendix \ref{app:experiment_details}.

\subsection{Detecting quantization and watermarking}\label{sec:quantization_watermarking}
In our first experiments, the reference distribution $\local$ represents full-precision weights published on Hugging Face.
We evaluate if tests can distinguish $\local$ from alternative distributions $\api$:
\begin{itemize}
    \item \textbf{Quantized models.} These alternatives represent the model inferenced at lower precisions: \nf \citep{dettmers2024qlora}, \inteight \citep{dettmers2022gpt3}, and \half.
    Some accounts suggest that quantization particularly degrades \llama-3 models on longform tasks \citep{llama3reddit,quantizationtwitter}.
    \item \textbf{Watermarked models.} Some providers may watermark outputs so that they are later detectable as having been generated by the platform. We apply the watermarking algorithm from \citet{kirchenbauer2023watermark} with default bias of 2.5. \todo{add distortion-free?}
\end{itemize}
We repeat evaluations for 5 instruction-tuned models:
\mistral 7B Instruct \citep{jiang2023mistral}, \llama-3 8B and 70B Instruct, and \llama-3.1 8B and 70B Instruct \citep{meta2024introducing}. Additional models are in Appendix \ref{app:additional_results_4_1}.


\tightparagraph{Tests.}
We compare three choices of kernels: the Hamming kernel (\ref{eqn:hamming}), the all-substrings kernel (\ref{eqn:all_substrings}), and the one-hot kernel (\ref{eqn:one_hot}).
We also evaluate two tests from the multinomial testing literature \citep{balakrishnan2018hypothesis,bhattacharya2015testing}:
\begin{equation}
    \text{L1}(\samplelocal, \sampleapi)
    = \sum_{z \in \samplelocal \cup \sampleapi} \left|
        \frac{\obs{z}{\samplelocal} -\obs{z}{\sampleapi}}{N}
    \right|
\end{equation}
\begin{equation}
    \chi^2(\samplelocal, \sampleapi)
    = N^2 \sum_{z\in \samplelocal \cup \sampleapi} \frac{
        \left(\obs{z}{\samplelocal} -\obs{z}{\sampleapi}\right)^2
        - \obs{z}{\samplelocal} - \obs{z}{\sampleapi}
    }{\obs{z}{\samplelocal} + \obs{z}{\sampleapi}}.
\end{equation}

\begin{figure}[bt]
    \centering
    \includegraphics[width=\textwidth]{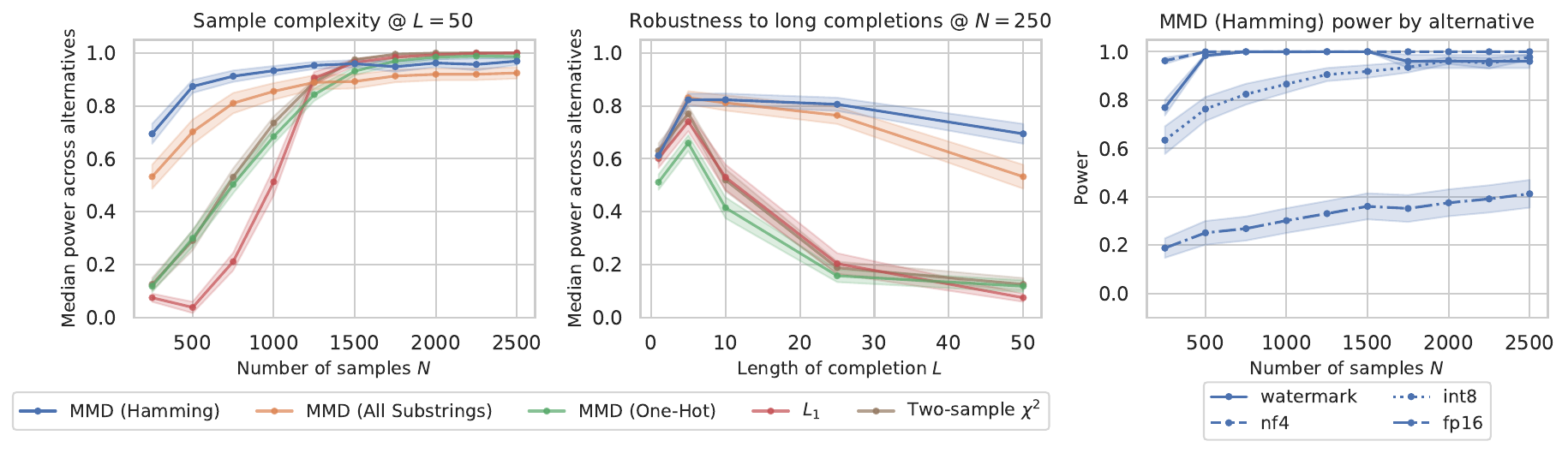}
    \caption{
        \textit{(Left)} Sample complexity of tests. At an average of just 10 samples per prompt, the Hamming MMD test is able to detect quantization and watermarking with nontrivial power. 
        Curves first median power across alternative distributions $Q$, averaged over language models and prompt distributions, with shaded standard errors. 
        Results stratified by language model and alternative are in Appendix \ref{app:additional_results_4_1}.
        \textit{(Middle)} While other tests rapidly degrade in power when the user is interested in longer completions, the Hamming MMD test maintains power best across completion lengths.
        \textit{(Right)} Power of the Hamming MMD test, stratified by alternative distribution. The test is significantly less powerful against the \half alternative.
        \vspace{-0.1in}
    }
    \label{fig:sample_complexity_length}
\end{figure}

\tightparagraph{Results.}
Figure \ref{fig:sample_complexity_length} (left) compares the empirical sample complexities of each test.
To draw out a sample complexity curve, we vary the number of samples from $N=10m$ to $N=100m$, where $m=25$ is the number of prompts in the prompt distribution.
We observe that the Hamming MMD test attains the highest power with the fewest samples:
at an average of 10 samples per prompt, this test has a median power of 77.4\% across alternatives.
In Figure \ref{fig:sample_complexity_length} (right), we break down power by alternative distribution.
The Hamming test is strong on all alternatives except \half, 
where the initial power at $N=10m$ is much smaller.
This suggests that \half and \full differ in ways that are not captured by the Hamming kernel.

To accommodate user tasks which require very long completions, it is important that tests retain power as the completion length $L$ increases, even though the size of the sample space grows exponentially with $L$.
In Figure \ref{fig:sample_complexity_length} (middle), we fix $N=250$ (\ie $N=10m$) and vary the completion length $L$ from 1 to 50 tokens.
We observe that the Hamming MMD and all substring tests are more robust to increasing completion length than the other tests.
This result is consistent with the intuition that a clever string kernel --- as opposed to a one-hot kernel --- can help MMD tests generalize to high-dimensional spaces.


\subsection{Detecting finetuning}\label{sec:finetuning}
Given their effectiveness in detecting quantized and watermarked models, we next ask if MMD-based tests can detect when a model has been finetuned.
We finetune \llama-3 8B Instruct on two datasets: 
a disjoint, \iid split of the testing Wikipedia task,
and an out-of-distribution code dataset \citep{codealpaca}.
We use a small learning rate of \num{1e-6} with AdamW \citep{loshchilov2017decoupled}.
We then use the Hamming MMD test to compare finetuned checkpoints $\api$ to the original model as the reference distribution $\local$.

Figure \ref{fig:model_model} (upper left) plots power against the checkpoint number of the finetuned model.
The Hamming MMD test is always able to detect finetuning with nontrivial (greater than 50\%) power, even after a single epoch (42 optimization steps).
One might expect that finetuning on the out-of-distribution code dataset would not affect the model's distribution on the Wikipedia testing task,
but we find this is not the case.
Finetuning affects the model on other distributions enough to be detectable by statistical tests.
These results suggest that it is challenging to isolate the effects of full finetuning to any single distribution, which may have implications for tasks such as unlearning or model editing \citep{hase2024fundamental}.

\subsection{Distinguishing if samples come from different language models}\label{sec:model_model}
\begin{figure}[tb]
    \centering
    \begin{subfigure}[t]{0.37\textwidth}
        \centering
        \vspace{0.1em}  
        \includegraphics[width=\textwidth]{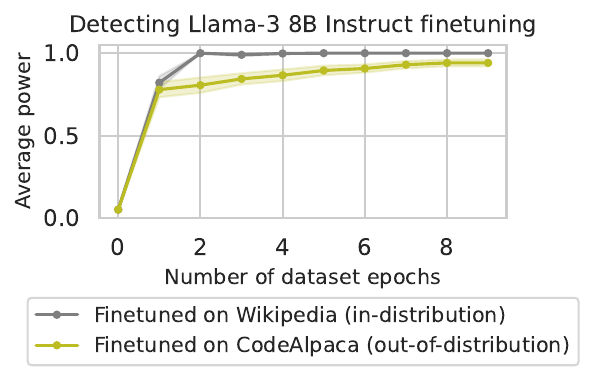}\\
        \vspace{0.1in}
        \scriptsize
        \begin{tabular}{ll}
            \toprule
            & Power  \\
            Alternative $\api$ &  ($\local$ = Llama-3.1 8B) \\
            \midrule
            Llama-3 8B & 0.76 (0.00) \\
            Llama-3 70B & 1.00 (0.00) \\
            Llama-3.1 70B & 0.75 (0.07) \\
            OLMo 7B & 1.00 (0.00) \\
            gpt-4o-mini & 1.00 (0.00) \\
            Llama-3.1 8B (watermark) & 0.32 (0.04) \\
            Llama-3.1 8B (\inteight) & 0.07 (0.01) \\
            Llama-3.1 8B (\nf) & 1.00 (0.00) \\
            \bottomrule
        \end{tabular}        
    \end{subfigure}
    \hfill
    \begin{subfigure}[t]{0.62\textwidth}
        \centering
        \vspace{0pt}  
        \includegraphics[width=\textwidth]{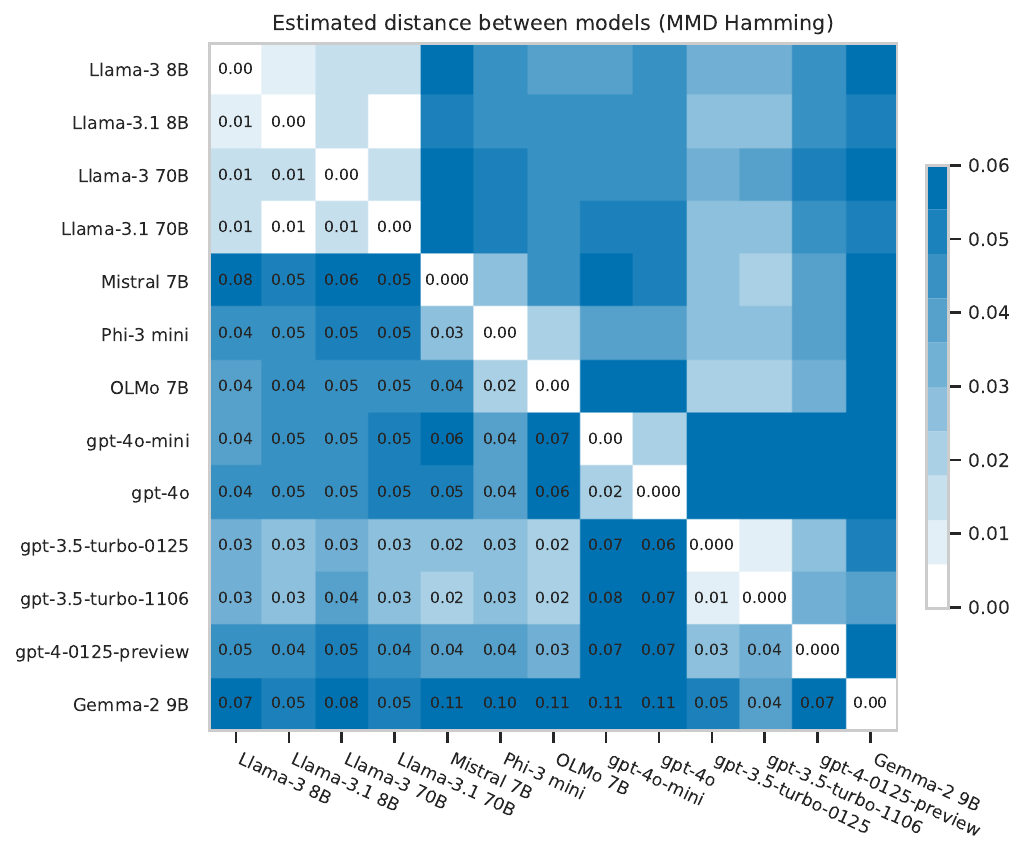}
    \end{subfigure}%
    \caption{
        \textit{(Upper left)} The Hamming MMD test is able to detect when \llama-3 8B has been finetuned on datasets of \num{1000} samples, even after a single epoch. Power is higher, earlier, when the finetuning distribution is \iid with the testing distribution.
        \textit{(Lower left)} The Hamming MMD test can also detect when two models are different with near-perfect power. Standard errors are over prompt distributions. Full results are in Appendix \ref{app:higher_dimensional}.
        \textit{(Right)} The MMD framework allows us to estimate statistical distance between any models from which we can draw samples.
        The cells show average estimated MMDs over \num{10} bootstraps. Rows are sorted using spectral clustering with two components. 
        Models within a family are typically clustered together, suggesting that factors like training data, rather than scale, determine model similarity.
        \vspace{-0.15in}
    }
    \label{fig:model_model}
\end{figure}

We next explore if the Hamming MMD test can distinguish whether two bodies of text are generated from the same language model.
In this setting, $\local$ and $\api$ are two different language models, drawn from a pool of 13 instruction-tuned models (Figure \ref{fig:model_model} right),
including eight open-weight models \citep{abdin2024phi,groeneveld2024olmo,team2024gemma} and and five OpenAI closed-weight models.
In order to compare models with different tokenizers, samples must be compared in character space.
We sample $L=50$-token completions to the Wikipedia language modeling task as before,
and then we decode these to characters, ignoring special tokens.
The vocabulary of interest is now all Unicode characters ($|\mathcal V| = \num{155063}$), the maximum completion length is $L \approx \num{1000}$ characters, and we test with $N=10m$ samples.

Switching to character space makes the testing problem significantly more challenging, 
as the sample space is higher dimensional:
Figure \ref{fig:model_model} (lower left) shows that power to detect quantization and watermarking is lower in this setting than in the token space setting (\S\ref{sec:quantization_watermarking}).
However, even in this more challenging setting, we find that all model swaps are detectable with 100\% power, except for pairs within the \llama family, \eg \llama-3.1 8B and \llama-3.1 70B (75\% power) or \llama-3 8B and \llama-3.1 8B (76\% power; Figure \ref{fig:model_model} lower left). 
Model swaps are significantly more detectable than watermarking or quantization.
Appendix \ref{app:samples_models_vs_models} shows qualitative examples of completions from different models.

\paragraph{Estimating distances between models.}
A useful feature of the MMD tests is that the test statistic is an estimator of a distance.
As a result, we can reuse the machinery to quantify the \textit{degree} to which two models differ by estimating $\mathbb E_{x \sim \prior}[\text{MMD}(\local(\cdot \mid x), \api(\cdot \mid x))]$.\footnote{Note that the Hamming MMD is a pseudometric: $\text{MMD}_\text{Hamming}(\local, \api) = 0$ does not imply $\local = \api$ (Appendix \ref{app:metrics}).}
Figure \ref{fig:model_model} (right) estimates the Hamming MMD between all pairs of models.
To decrease estimator error, we increase the sample size to $N=100m$ samples from each model and report the average $\widehat{\text{MMD}}$ over \num{10} simulations, along with standard errors.
We observe that models within a family are typically clustered together, suggesting that training data, rather than scale, determines model similarity.
Surprisingly, while generations of \llama models (3 and 3.1) are close in distance, some generations of GPT models (\eg 4-preview and 3.5-turbo) are not.
This result suggests the use of different training data or procedures between these models.

\section{Auditing inference API providers}\label{sec:apis}
As a case study, we now apply our test to 31 commercial inference endpoints for four of Meta's \llama models (Figure \ref{fig:hero}).
These endpoints are distributed across nine API providers from Summer 2024:
\href{https://aws.amazon.com/bedrock/}{Amazon Bedrock},
\href{https://anyscale.com/}{Anyscale},\footnote{We collected samples from Anyscale's serverless endpoints from before they were deprecated in August 2024.}
\href{https://azure.microsoft.com/en-us/products/machine-learning/generative-ai}{Azure AI Studio},
\href{https://deepinfra.com/}{Deepinfra},
\href{https://fireworks.ai/}{Fireworks AI},
\href{https://groq.com/}{Groq},
\href{https://www.perplexity.ai/}{Perplexity},
\href{https://replicate.com/}{Replicate},
and \href{https://together.ai/}{Together.ai}.\footnote{
At the time of writing, only two providers disclosed distribution-altering optimizations:
\href{https://fireworks.ai/}{Fireworks AI noted semantic caching}, and \href{https://web.archive.org/web/20240723005159/https://www.together.ai/blog/together-inference-engine-2}{Together.ai noted quantization}.}
We are interested in whether endpoints differ from weights published by Meta on Hugging Face inferenced at commonly accepted precisions.
Specifically, we consider two possible null distributions: the full-precision weights ($\local_1$) and the \half-precision weights ($\local_2$).
The null and alternative hypotheses are
\begin{equation}
    \begin{aligned}
    H_0:& \quad\prior(x) \local_1(y \mid x) = \prior(x) \api(y \mid x)\quad \text{OR} \quad \prior(x) \local_2(y \mid x) = \prior(x) \api(y \mid x), \\
    H_1:& \quad\prior(x) \local_1(y \mid x) \neq \prior(x) \api(y \mid x) \quad \text{AND} \quad \prior(x) \local_2(y \mid x) \neq \prior(x) \api(y \mid x).
    \end{aligned}
    \label{eqn:composite_hypotheses}
\end{equation}
To test this composite hypothesis, we collect three samples: $\samplelocalone$ and $\samplelocaltwo$ from the two null distributions, and $\sampleapi$ from the API.
We then conduct 2 two-sample tests, one for $(\samplelocalone, \sampleapi)$ and another for $(\samplelocaltwo, \sampleapi)$, and obtain p-values $p_1$ and $p_2$.
We set the overall rejection rule to 
\begin{equation}
    \decisionrule(\samplelocalone, \samplelocaltwo, \sampleapi) = \ind{p_1 < \alpha  \land  p_2 < \alpha}.
    \label{eqn:composite_decision_rule}
\end{equation}
This rule continues to control the FPR at $\alpha$ under the composite null hypothesis: without loss of generality, suppose $Q = P_1$. Since $P_{Q=P_1}(p_1 < \alpha) \le \alpha$, we have $P_{Q=P_1}(p_1 < \alpha  \land  p_2 < \alpha) = P_{Q=P_1}(p_1 < \alpha)P_{Q=P_1}(p_2 < \alpha \mid p_1 < \alpha) \le \alpha$.
Note that this rule may be pessimistic, reducing power.

\paragraph{Experiment details.} We consider testing with three prompt distributions $\pi$.
For all models, we test with one set of the Wikipedia completion task from \S\ref{sec:simulations},
where $\prior$ is uniform over $m=25$ prompts, and $L=\num{50}$ tokens or around \num{1000} characters.
For the smaller \llama-3 8B and \llama-3.1 8B models, we also test with the
coding task \humaneval \citep{humaneval} and
instruction task \ultrachat \citep{ultrachat}.
Both $\prior$ are uniform over $m=20$ prompts, and $L=\num{250}$ tokens or \num{3000} characters. 
Because APIs often return decoded completions, rather than individual tokens, we conduct all tests in character space, as in \S\ref{sec:model_model}.
We explicitly requested all samples at temperature 1.
Tests are conducted at level $\alpha=0.01$ using $N=10m$ samples.
To reduce variance, we repeat tests over ten samples and fail endpoints if the average rejection rate is $\ge 0.5$.
For the most expensive endpoint (Azure's \llama-3 70B), a Wikipedia audit costs \$0.14, \humaneval \$0.83, and \ultrachat \$0.93.
For the cheapest endpoint (Fireworks' \llama-3 8B), all three audits cost less than \$0.02.
For additional details, including the dates we collect API samples, see Appendix \ref{app:experiment_details}.

\begin{table}[bt]
    \centering
    \small
    \begin{tabular}{lllllllll}
        \toprule
         & \multicolumn{4}{l}{Wikipedia} & \multicolumn{2}{l}{\humaneval} & \multicolumn{2}{l}{\ultrachat} \\
         & 3 8B & 3.1 8B & 3 70B & 3.1 70B & 3 8B & 3.1 8B & 3 8B & 3.1 8B \\
        \midrule
        Amazon & \cmark & \xmark & \cmark & \xmark & \cmark & \xmark & \xmark & \xmark \\
        Anyscale & \cmark & --- & --- & --- & --- & --- & --- & --- \\
        Azure & \cmark & \cmark & \cmark & \cmark & \cmark & \cmark & \cmark & \cmark \\
        Deepinfra & \cmark & \cmark & \cmark & \cmark & \cmark & \cmark & \cmark & \cmark \\
        Fireworks & \cmark & \cmark & \cmark & \cmark & \cmark & \xmark & \cmark & \xmark \\
        Groq & \cmark & \cmark & \cmark & \xmark & \cmark & \xmark & \cmark & \cmark \\
        Perplexity & \xmark & \xmark & \xmark & \xmark & \xmark & \xmark & \xmark & \xmark \\
        Replicate & \cmark & --- & \xmark & --- & \cmark & --- & \cmark & --- \\
        Together & \cmark & \cmark & \cmark & \cmark & \cmark & \cmark & \cmark & \cmark \\
        \bottomrule
        \end{tabular}
    \caption{
        Audit results for 31 API endpoints across nine companies and four language models. \xmark\xspace denotes an endpoint failure, \ie the average rejection rate over ten samples is $\ge$ 50\%.
    }
    \label{tab:api_pass_fail}
\end{table}

\begin{figure}[tb]
    \centering
    \includegraphics[width=\textwidth]{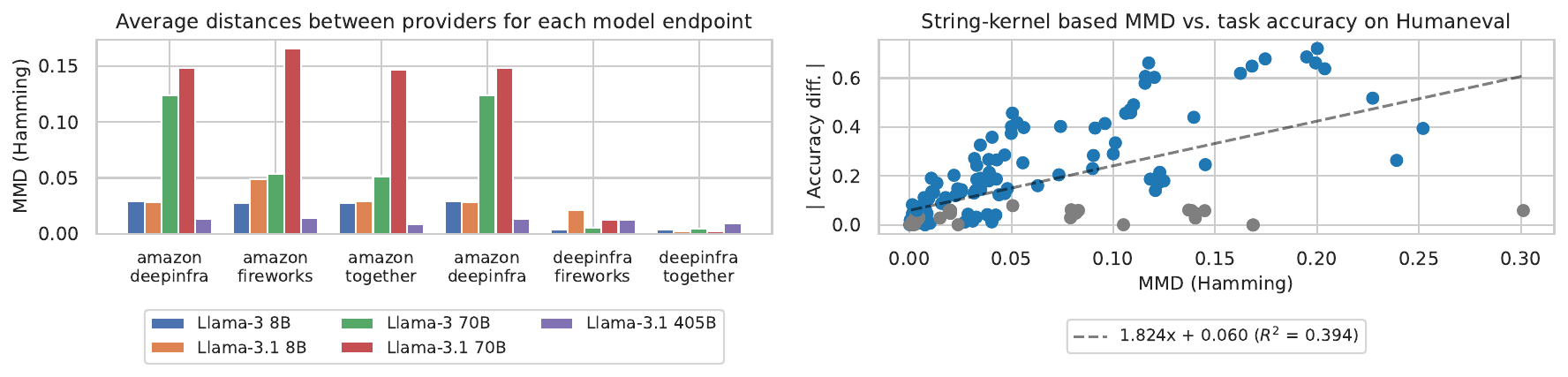}
    \caption{
        \textit{(Left)} Average MMD (Hamming) between providers for each model. Amazon Bedrock's \llama-3 and -3.1 70B models are the most different from the other providers.
        \textit{(Right)} Absolute difference in \humaneval average accuracy vs. the MMD (Hamming). There is a moderate positive correlation between MMD and task accuracy. Gray points indicate pairs where both distributions have accuracy $<$ 10\%. There are multiple ways to be wrong for a task, and the MMD captures these differences.
        \vspace{-0.8em}
    }
    \label{fig:api_mmd_humaneval}
\end{figure}

\paragraph{Results.}
Despite power being generally reduced due to the composite decision rule, 
the test flags several endpoints (Table \ref{tab:api_pass_fail}).\footnote{In Figure \ref{fig:hero}, we combine the results of the tests on the prompt distributions via a Bonferroni correction,
setting the level of each test to $\alpha = 0.01/3$ and rejecting if the endpoint fails on any of the three prompt distributions.
}
Notably, Amazon Bedrock and Perplexity have the most endpoints flagged, with the latter failing all tests.
In Box \ref{box:perplexity_qualitative}, we include an example comparing samples from the \full null and Perplexity; these samples suggest that Perplexity serves a lower entropy distribution than the full-precision model.
Additional qualitative samples can be found in Appendix \ref{app:samples_models_vs_apis}.
The \humaneval and \ultrachat prompt distributions elicit more failures than the Wikipedia distribution; this may be because these knowledge-intensive distributions are more sensitive to changes such as quantization.

To estimate the effect size of deviations, we estimate the MMD between providers and the nulls using ten bootstraps of $N=100m$ samples each.
We find that some deviations are quite large:
some APIs' implementations are further from the reference weights than if the provider had substituted in an entirely different language model.
For example, the deviation between Perplexity's \llama-3 8B and the \full null on the Wikipedia testing task is 0.03.
This is comparable to the deviation between \llama-3 8B and GPT-3.5-Turbo (0.03; Figure \ref{fig:model_model} right), Phi-3 mini (0.04), or OLMo 7B (0.04).

\paragraph{Correlating MMD with task accuracy.}
We now ask how well the Hamming MMD correlates with task accuracy when available.
Automated verifiers exist for one of our prompt distributions, \humaneval.
We find a moderate positive correlation between the absolute average accuracy difference and the Hamming MMD (Figure \ref{fig:api_mmd_humaneval} right,  $R^2 = 0.392$).
In several cases, the MMD is high but the accuracy difference is low.
These are often when both task accuracies are low: 
the gray points in the figure highlight pairs where both distributions have accuracy $<$ 10\%.
In these cases, the MMD captures that there are multiple ways to be wrong for a task.
For example, the bottom-right-most point in the figure contrasts \llama-3.1 8B from Perplexity vs. the \nf version; 
Perplexity answers with incorrect code snippets, while \nf outputs nonsense completions (\eg ``simulation!!!!!!!!!![...]'').
Samples from this comparison can be found in Appendix \ref{app:samples_models_vs_apis}.

\paragraph{Which models do providers differ on most?}
Finally, we ask whether providers differ on some more models more than others.
Figure \ref{fig:api_mmd_humaneval} (left) estimates the MMD between providers for each model, and the large \llama-3.1 405B model.
We find that deviations are larger on the earlier \llama models than the recent \llama-3.1 405B model. 

\begin{samplebox}[label=box:perplexity_qualitative]{Samples from \full vs. Perplexity for \llama-3 8B (Instruct)}
    \textbf{Prompt:} Continue the paragraph. Do not output anything except the continuation to the paragraph. Start the continuation immediately.\\
    ``Weaver, Tom. I Talked with a Zombie: Interviews with 23 Veterans of Horror and Sci-Fi Films and Tele...''\\

    {\scriptsize
    \def\arraystretch{1.5}
    \begin{tabularx}{\linewidth}{>{\raggedright\arraybackslash}X>{\raggedright\arraybackslash}X}
        \textit{\full} & \textit{Perplexity} \\
        \arrayrulecolor[gray]{0.9}\hline
        ``...vision, Some of Whom Are Still Alive. New York: Pocket Books, 1981. According to Tom Weaver, Vincent Price was known for being a bit of a practical joker, and he would often play harmless pranks on the crew & ``...visions. New York: St. Martin's Press, 1988. This collection of interviews with horror and science fiction professionals is a fascinating look at the creative process and the people behind some of the most iconic films and TV shows of the 
        \\
        \arrayrulecolor[gray]{0.9}\hline
        ``...vision, voodoo, and the Apocalypse. New York: St. Martin's Press, 1990. The book features interviews with notable horror and science fiction personalities, including Greg Cannom, makeup artist for The Lost Boys and RoboCop	 & ``...visions. New York: St. Martin's Press, 1988. In this collection of interviews, Weaver delves into the experiences of some of the most iconic figures in the horror and science fiction genres, including George Romero, John Carpenter,
        \\
        \arrayrulecolor[gray]{0.9}\hline
        ``...vision. Written by Tom Weaver, a renowned authority on classical horror cinema, this book brings together intimate and revealing conversations with some of the most legendary figures in the genre, including Vincent Price, Boris Karloff, and Bela Lugosi. Interviews	 & ``...vision. New York: St. Martin's Press, 1988. 224 pp. This collection of interviews with horror and science fiction professionals is a fascinating look at the people behind the genre's most iconic films. Weaver's conversational style
        \\
        \arrayrulecolor[gray]{0.9}\hline
    \end{tabularx}
    }
\end{samplebox}

\section{Related work}\label{sec:related_work}
\tightparagraph{Auditing ML APIs.}
Prior work either only applies to the classifier setting \citep{chen2021did,chen2022hapi}
or audit by monitoring performance on low-temperature multiple-choice or short-answer benchmarks \citep{artificialanalysis,chen2023chatgpt,eyuboglu2024model}.
Concurrent work studies a general language model setting, but does not provide guarantees of audit correctness \citep{sam2025predicting}. 
In contrast, we formalize auditing as a hypothesis testing problem, allowing us to provide FPR guarantees, and we study the general setting of sampling completions from language models, including for tasks without automated verifiers.
In cryptography, a body of work has aimed to verify ML API predictions through proof protocols \citep{ghodsi2017safetynets,feng2021zen,liu2021zkcnn,kang2022scaling,weng2023pvcnn,lee2024vcnn}.
These methods require APIs to cooperate and opt-in to providing proofs of valid inference alongside each prediction; public verifiers then check these proofs for correctness with perfect accuracy.
Unfortunately, these methods scale too poorly to apply to large language models:
in \citet{kang2022scaling}, generating a proof takes 41 minutes \emph{per prediction} for a 68M parameter MobileNet \citep{sandler2018mobilenetv2}.

\tightparagraph{Two-sample testing.}
Testing whether two samples come from the same distribution is an established problem in statistics.
A line of work focuses on testing multinomial distributions \citep{batu2013testing,chan2014optimal,canonne2020survey}, including when samples have unequal sizes \citep{bhattacharya2015testing,diakonikolas2016new,balakrishnan2018hypothesis}.
MMD-based tests are a general approach that do not assume a specific distributional form \citep{gretton2012kernel}.
These tests can be applied to structured data when paired with appropriate kernels \citep{lodhi2002text,borgwardt2006integrating}.


\section{Conclusion}\label{sec:conclusion}
As the public grows more dependent on black-box APIs to interact with language models,
tools for auditing these APIs are increasingly important.
In this work, we unified several API auditing tasks under the formalization of Model Equality Testing, a two-sample distribution testing problem, 
and we extensively validated candidate tests for this problem.
Future work could explore stronger tests than those we have presented, or explore how to adapt these tests to other modalities, such as image generation models.
To help facilitate this research, we open-source the dataset of 1 million LLM completions used for this work at \url{https://github.com/i-gao/model-equality-testing}.
This dataset contains completions from five models across quantized, watermarked, and API alternatives; also see Appendix \ref{app:dataset}.

\FloatBarrier
\section*{Acknowledgements}
The authors would like to thank Mert Yuksekgonul, Steven Cao, Chenchen Gu, Chenglei Si, Nicole Meister, Yifan Mai, Simon Guo, Teddi Worledge, Yuhui Zhang, and other members of the Guestrin and P-Lambda labs for feedback on this paper.

\section*{Conflicts of interest}
PL is a co-founder of Together AI; this work was done in his Stanford capacity.  The topic, research, and results of this work were not shared with Together, or any other API provider evaluated, until the public release of the paper.


\bibliography{references}
\bibliographystyle{iclr2025_conference}

\clearpage
\appendix
\section{Additional notes on tests}\label{app:methods}
\subsection{Which MMD kernels lead to valid metrics?}\label{app:metrics}
Recall that a metric on probability distributions satisfies 
(1) symmetry,
(2) the triangle inequality,
and (3) $d(\local, \api) = 0$ if and only if $\local = \api$.
On the other hand, a pseudo-metric satisfies the first two properties and has $d(\local, \local) = 0$.
Regardless of the kernel, the MMD as defined in Equation \ref{eqn:mmd} is clearly symmetric. 
It also satisfies triangle inequality, since $\text{MMD}(P_1, P_3) = \text{MMD}(P_1, P_2) + \text{MMD}(P_2, P_3)$ for any $P_1, P_2, P_3$.
The question left is whether our kernels make the MMD injective as in Condition 3.

\begin{itemize}

\item \textbf{One-hot kernel}. \citep{gretton2012kernel} prove that universal kernels \citep{steinwart2001influence} result in an injective MMD.
\citep{borgwardt2006integrating} show that a kernel defined on a finite domain $\mathcal X$ is universal if $k$ satisfies \textit{strict positive definiteness}:
\ie $k$ induces a nonsingular Gram matrix for any finite set of points $X \subseteq \mathcal X$. 
This is true if $\phi(x^{(1)}), \cdots, \phi(x^{(n)})$ are linearly independent for any set of distinct points $x^{(1)}, \cdots, x^{(n)} \in \mathcal X$.
For the one-hot kernel, the associated feature map $\phi_\text{one-hot}$ is of length $|V|^L$, where the $i$th entry is an indicator for whether $x$ is equal to the $i$th string.
Since all $\phi(x)$ is one-hot and for distinct sets $X$, no two $\phi(x)$ are both 1 at the same index, the $\phi(x)$ are linearly independent.
Therefore this kernel is universal, and $\text{MMD}_\text{one-hot}$ is a valid metric.

\item \textbf{All-substrings kernel}. \citet{borgwardt2006integrating} (Theorem 2.7) prove that this kernel is universal, and thus the MMD is a metric.

\item \textbf{Hamming kernel}. We will show that the MMD is not injective by showing that the mean embedding $\mathbb E_P[\phi(x)]$ is not injective, 
\ie there exist $P \neq Q$ with $\mathbb E_P[\phi(x)]=\mathbb E_Q[\phi(x)]$.
For the Hamming kernel, the associated feature map is of length $|\mathcal V| \times L$:
$$
\phi(x) = \begin{bmatrix}
\ind{x_1 = v_1} \\ \ind{x_1 = v_2} \\ \cdots \\ \ind{x_1 = v_{|\mathcal V|}} \\ \cdots \\ \ind{x_L = v_1} \\ \ind{x_L = v_2} \\ \cdots \\ \ind{x_L = v_{|\mathcal V|}}
\end{bmatrix}
\implies
\mathbb E[\phi(x)] = \begin{bmatrix}
P(x_1 = v_1) \\ P(x_1 = v_2) \\ \cdots \\ P(x_1 = v_{|\mathcal V|}) \\ \cdots \\ P(x_L = v_1) \\ P(x_L = v_2) \\ \cdots \\ P(x_L = v_{|\mathcal V|})
\end{bmatrix}
$$
\ie the mean embedding stacks all marginal distributions of $\local$.
But this shows the mean embedding is not injective: we know that multiple joint distributions $P \neq Q$ can map to 
the same marginal distributions.
Thus the Hamming MMD is not injective, and it is only a pseudo-metric.
\end{itemize}

\subsection{Simulating p-values}\label{app:pvalues}
P-values for MMD tests may be simulated in two ways:

\begin{enumerate}
    \item \textbf{Simulating the test statistic under the null (Algorithm \ref{alg:simulate}).}
    This is done by repeatedly sampling $\sampleapi$ and $\samplelocal$ from $\local$ and caching $\widehat{\text{MMD}}(\samplelocal, \sampleapi)$.
    The p-value is then the proportion of times the test statistic is greater than or equal to the observed test statistic.
    We conduct tests using this method in the main text, reusing the same cached empirical distribution of the test statistic under the null for all alternatives at that sample size.
    Note that this method requires significant sampling access to $\local$.
    \item \textbf{Permutation procedure (\citet{lehmann1986testing}; Algorithm \ref{alg:permutation}).}
    Given samples $\sampleapi$ and $\samplelocal$, the permutation procedure randomly shuffles the labels of the samples and computes the test statistic on the permuted samples.
    This process is repeated many times to estimate the null distribution of the test statistic.
    The p-value is then the proportion of times the permuted test statistic is greater than or equal to the observed test statistic.
    This method does not require additional sampling access to $\local$ but may have lower power.
    We conduct experiments using this method in Appendix \ref{app:permutation}.
\end{enumerate}

\begin{algorithm}[H]
    \caption{Simulating the test statistic under the null}
    \label{alg:simulate}
    \begin{algorithmic}[1]
        \State \textbf{Input:} Number of simulations $B$, null distribution $\local$, test samples $\sampleapi$ and $\samplelocal$
        \State Initialize a list $\mathcal{T} = []$ to store simulated test statistics
        \For{$i = 1, 2, \dots, B$}
            \State Sample $\sampleapi^{(i)} \sim \local$ and $\samplelocal^{(i)} \sim \local$
            \State Compute $\widehat{\text{MMD}}(\sampleapi^{(i)}, \samplelocal^{(i)})$ and append to $\mathcal{T}$
        \EndFor
        \State Compute p-value as the proportion of $\mathcal T$ greater than or equal to $\widehat{\text{MMD}}(\samplelocal, \sampleapi)$
        \State \textbf{Return} p-value
    \end{algorithmic}
\end{algorithm}

\begin{algorithm}[H]
    \caption{Permutation testing}
    \label{alg:permutation}
    \begin{algorithmic}[1]
        \State \textbf{Input:} Number of permutations $B$, test samples $\sampleapi$ and $\samplelocal$
        \State Concatenate $\sampleapi$ and $\samplelocal$ into a single dataset $\mathcal{D} = [\sampleapi, \samplelocal]$
        \State Initialize a list $\mathcal{T} = []$ to store permuted test statistics
        \For{$i = 1, 2, \dots, B$}
            \State Randomly shuffle $\mathcal{D}$ and split into two sets: $\sampleapi^{(i)}$ and $\samplelocal^{(i)}$
            \State Compute $\widehat{\text{MMD}}(\sampleapi^{(i)}, \samplelocal^{(i)})$ and append to $\mathcal{T}$
        \EndFor
        \State Compute p-value as the proportion of $\mathcal T$ greater than or equal to $\widehat{\text{MMD}}(\samplelocal, \sampleapi)$
        \State \textbf{Return} p-value
    \end{algorithmic}
\end{algorithm}

\FloatBarrier
\section{Experiment details}\label{app:experiment_details}

\subsection{Sampling and dataset details}\label{app:dataset}
All experiments were conducted by sampling with replacement from a pre-collected dataset of language model completions,
which we release alongside this paper at \url{https://github.com/i-gao/model-equality-testing}.
The dataset consists of completions from five models:
\texttt{mistralai/Mistral-7B-Instruct-v0.3}, 
\texttt{meta-llama/Meta-Llama-3-8B-Instruct},
\texttt{meta-llama/Meta-Llama-3.1-8B-Instruct},
\texttt{meta-llama/Meta-Llama-3-70B-Instruct}, and
\texttt{meta-llama/Meta-Llama-3.1-70B-Instruct}.
We collected multiple completions per prompt for prompts from Wikipedia \citep{wikipedia}, \humaneval \citep{humaneval}, and \ultrachat \citep{ultrachat}.
In total, our dataset contains 440 prompts:
80 from Wikipedia in each of English, Spanish, French, German, and Russian, 
as well as 20 from \humaneval and 20 from \ultrachat.
We repeated the collection process for each model at each precision (\full, fp16, \half, \inteight, and \nf) and when watermarked using the method in \citet{kirchenbauer2023watermark}, as well as from each API audited in the paper.
In total, this dataset size is 1.1M completions.

\begin{samplebox}[label=box:humaneval]{Sample prompt for \humaneval}
Complete the code. Do not output anything except the completion. Start the continuation immediately.\\
{\scriptsize 
\begin{verbatimcode}
from typing import List

def has_close_elements(numbers: List[float], threshold: float) /> bool:
    """ Check if in given list of numbers, are any two numbers closer 
    to each other than given threshold.
    >>> has_close_elements([1.0, 2.0, 3.0], 0.5)
    False
    >>> has_close_elements([1.0, 2.8, 3.0, 4.0, 5.0, 2.0], 0.3)
    True
    """
\end{verbatimcode}
}
\end{samplebox}

\begin{samplebox}[label=box:ultrachat]{Sample prompt for \ultrachat}\label{box:ultrachat}
Explain how the invention and widespread use of digital music formats such as MP3s and streaming services like Spotify have impacted the way music is distributed, consumed, and monetized in the music industry, and how this has affected the relationship between artists and their fans. Provide specific examples of how technological advancements have changed the production and consumption of music, including changes to the format and length of songs, the role of record labels, and the use of social media to promote artists and their work. Additionally, discuss possible future developments in music technology and their potential impact on the industry and consumer habits.
\end{samplebox}

For the goodness-of-fit experiments in Appendix \ref{app:gof},
we collected completions for a disjoint set of 20 additional prompts from each of language of Wikipedia for Mistral 7B and \llama-3 8B.
This set also includes the log probabilities of each sample under the full precision model.
All other experiments outside of this goodness-of-fit experiment were conducted on the previous, larger dataset.

\paragraph{Details about local sampling (\full, \half, \inteight, \nf4, and watermark).}
We collected \num{15000} completions per prompt for the \full precision,
and \num{5000} completions per prompt for the other precisions and watermarked alternatives.
Local sampling was performed on a mix of RTX 6000, RTX 3090, Quadro RTX 8000, and A100 GPUs using a mixture of the Transformers \citep{wolf2020transformers} and VLLM \citep{kwon2023efficient} libraries.
The watermarking, \nf, and \inteight implementations are from the Transformers library.
We use the default watermarking parameters of 2.5 bias and context width 1.
All samples were collected with vanilla decoding parameters of temperature 1 without top-k or top-p sampling, and parameter \texttt{max\_new\_tokens} set to $L=50$ for Wikipedia and $L=250$ for \humaneval and \ultrachat.

We used the default chat templates from the Transformers library for models.
It was important for us to match the chat templates during local sampling to those we believe the APIs to use, as the chat templates can affect the completions generated.
We confirmed that the number of tokens in our local rendering of the prompts matched the returned number of prompt tokens from API calls. 
The exception was for the \llama-3.1 models, where the default Transformers chat template includes the current date.
We found that this template did not match the number of prompt tokens returned by APIs;
however, the \llama-3 chat template did.\footnote{In October 2024, this behavior has changed for Together AI, which now uses the \llama-3.1 chat template that includes the current date. At the time we collected samples with the \llama-3 template, this was not the case.}
As a result, we used the \llama-3 chat template for \llama-3.1 models in our local sampling.

\paragraph{Details about API sampling.}
API samples are collected by repeatedly querying endpoints for one completion at a time.
We aimed to collect 250 completions per prompt for each API within a 24 hour window, but due to rate limits and request failures, some prompts had samples collected over multiple days.
The dates we query each endpoint are listed in Tables \ref{tab:3_8b_api_dates} -- \ref{tab:3.1_70b_api_dates}.
We query serverless endpoints offered by providers and use the same decoding parameters as for local sampling.
When available, we called the providers using their Python packages; otherwise, we made raw HTTP requests.

Below are provider-specific details:

\begin{itemize}
\item \textit{Anyscale.} Because Anyscale deprecated its endpoints during our data collection in August 2024, we were only able to collect samples from their \llama-3 8B endpoint for Wikipedia.
\item \textit{Together.}  We collected \llama-3 8B and 70B samples from Together.ai before they introduced separate reference, turbo, and lite endpoints.
We collected \llama-3.1 8B, 70B, and 405B from the turbo endpoints, which was the only option available at the time of collection.
\end{itemize}

\begin{table}[h]
    \centering
    \scriptsize
    \begin{tabular}{lllc}
        \toprule
        \textbf{Model} & \textbf{Dataset} & \textbf{Provider} & \textbf{Dates queried} \\
        \midrule
        \multirow{9}{*}{3 8B}  
            & \multirow{9}{*}{Wikipedia} 
            & Anyscale & 7/4 \\
            &  & Amazon & 7/8, 8/1 \\
            &  & Azure & 8/19-20, 8/24 \\
            &  & Deepinfra & 7/4, 8/1 \\
            &  & Fireworks & 7/4, 7/19 \\
            &  & Groq & 7/4, 8/1-4, 8/8 \\
            &  & Perplexity & 7/4 \\
            &  & Replicate & 7/4, 7/19 \\
            &  & Together & 7/4 \\
        \cmidrule(lr){2-4}
        & \multirow{8}{*}{\humaneval} 
            & Amazon & 7/29 \\
            &  & Azure & 8/24 \\
            &  & Deepinfra & 8/1 \\
            &  & Fireworks & 8/1, 8/12 \\
            &  & Groq & 8/4, 8/24 \\
            &  & Perplexity & 8/12 \\
            &  & Replicate & 8/1 \\
            &  & Together & 8/1, 8/12 \\
        \cmidrule(lr){2-4}
        & \multirow{8}{*}{\ultrachat} 
            & Amazon & 7/29 \\
            &  & Azure & 8/24 \\
            &  & Deepinfra & 8/1 \\
            &  & Fireworks & 8/1, 8/12 \\
            &  & Groq & 8/4, 8/24 \\
            &  & Perplexity & 8/12 \\
            &  & Replicate & 8/1-2 \\
            &  & Together & 8/1, 8/12 \\
        \bottomrule
    \end{tabular}
    \caption{
        Dates we queried \llama-3 8B inference endpoints.
    }
    \label{tab:3_8b_api_dates}
\end{table}

\begin{table}[h]
    \centering
    \scriptsize
    \begin{tabular}{lllc}
        \toprule
        \textbf{Model} & \textbf{Dataset} & \textbf{Provider} & \textbf{Dates queried} \\
        \midrule
        \multirow{7}{*}{3.1 8B} 
            & \multirow{7}{*}{Wikipedia} 
            & Amazon & 8/1 \\
            &  & Azure & 8/19-21, 8/23-24 \\
            &  & Deepinfra & 8/1-2 \\
            &  & Fireworks & 7/26-27, 8/6 \\
            &  & Groq & 8/1-4, 8/8-11, 8/24-26 \\
            &  & Perplexity & 8/1-2 \\
            &  & Together & 7/26-27, 8/6-8 \\
        \cmidrule(lr){2-4}
        & \multirow{7}{*}{\humaneval} 
            & Amazon & 8/1 \\
            &  & Azure & 8/24 \\
            &  & Deepinfra & 7/31 \\
            &  & Fireworks & 7/30 \\
            &  & Groq & 7/31, 8/24, 8/27 \\
            &  & Perplexity & 7/30 \\
            &  & Together & 7/30 \\
        \cmidrule(lr){2-4}
        & \multirow{7}{*}{\ultrachat} 
            & Amazon & 7/26-27, 8/1, 8/6 \\
            &  & Azure & 8/24 \\
            &  & Deepinfra & 7/31 \\
            &  & Fireworks & 7/26, 7/30, 8/6 \\
            &  & Groq & 7/31-8/1, 8/27 \\
            &  & Perplexity & 7/30 \\
            &  & Together & 7/26-27, 7/30, 8/6 \\
        \bottomrule
    \end{tabular}
    \caption{
        Dates we queried \llama-3.1 8B inference endpoints.
    }
    \label{tab:3.1_8b_api_dates}
\end{table}

\begin{table}[h]
    \centering
    \scriptsize
    \begin{tabular}{lllc}
        \toprule
        \textbf{Model} & \textbf{Dataset} & \textbf{Provider} & \textbf{Dates queried} \\
        \midrule
        \multirow{8}{*}{3 70B} 
            & \multirow{8}{*}{Wikipedia} 
            & Amazon & 7/8, 8/1 \\
            &  & Azure & 8/19-21, 8/24 \\
            &  & Deepinfra & 7/4, 8/5-6 \\
            &  & Fireworks & 7/4, 7/31-8/1 \\
            &  & Groq & 7/4, 7/31, 8/2-13 \\
            &  & Perplexity & 7/4, 7/8, 8/5-6 \\
            &  & Replicate & 7/4, 7/19, 7/31-8/1 \\
            &  & Together & 7/4, 7/31-8/1 \\
        \cmidrule(lr){2-4}
        & \multirow{8}{*}{\humaneval} 
            & Amazon & 7/29 \\
            &  & Azure & 8/24 \\
            &  & Deepinfra & 8/1 \\
            &  & Fireworks & 8/6 \\
            &  & Groq & 8/1 \\
            &  & Perplexity & 8/6 \\
            &  & Replicate & 8/6 \\
            &  & Together & 8/6, 8/24 \\
        \cmidrule(lr){2-4}
        & \multirow{8}{*}{\ultrachat} 
            & Amazon & 8/24 \\
            &  & Azure & 8/25-26 \\
            &  & Deepinfra & 8/1-2 \\
            &  & Fireworks & 8/6-7, 8/24 \\
            &  & Groq & 8/1-2, 8/4, 8/24 \\
            &  & Perplexity & 8/6, 8/24 \\
            &  & Replicate & 8/6 \\
            &  & Together & 8/6, 8/24 \\
        \bottomrule
    \end{tabular}
    \caption{
        Dates we queried \llama-3 70B inference endpoints.
    }
    \label{tab:3_70b_api_dates}
\end{table}

\begin{table}[h]
    \centering
    {\scriptsize
    \begin{tabular}{lllc}
        \toprule
        \textbf{Model} & \textbf{Dataset} & \textbf{Provider} & \textbf{Dates queried} \\
        \midrule
        \multirow{7}{*}{3.1 70B} 
            & \multirow{7}{*}{Wikipedia} 
            & Amazon & 8/2 \\
            &  & Azure & 8/24 \\
            &  & Deepinfra & 8/2 \\
            &  & Fireworks & 7/27-28, 8/2, 8/5-6 \\
            &  & Groq & 8/2-5, 8/8-11, 8/21-24 \\
            &  & Perplexity & 8/2, 8/6 \\
            &  & Together & 7/27-28, 8/6 \\
        \cmidrule(lr){2-4}
        & \multirow{7}{*}{\humaneval} 
            & Amazon & 8/24 \\
            &  & Azure & 8/24, 8/26 \\
            &  & Deepinfra & 7/31 \\
            &  & Fireworks & 7/30 \\
            &  & Groq & 7/31, 8/1-2, 8/24 \\
            &  & Perplexity & 7/30-31 \\
            &  & Together & 7/30, 8/6 \\
        \cmidrule(lr){2-4}
        & \multirow{7}{*}{\ultrachat} 
            & Amazon & 8/24-25 \\
            &  & Azure & 8/25-26 \\
            &  & Deepinfra & 7/31 \\
            &  & Fireworks & 7/30 \\
            &  & Groq & 7/31-8/1, 8/24 \\
            &  & Perplexity & 7/30-31 \\
            &  & Together & 7/30, 8/6 \\
        \bottomrule
    \end{tabular}}
    \caption{
        Dates we queried \llama-3.1 70B inference endpoints.
    }
    \label{tab:3.1_70b_api_dates}
\end{table}

\begin{table}[h]
    \centering
    {\scriptsize
    \begin{tabular}{lllc}
        \toprule
        \textbf{Model} & \textbf{Dataset} & \textbf{Provider} & \textbf{Dates queried} \\
        \midrule
        \multirow{7}{*}{3.1 405B} 
            & \multirow{7}{*}{Wikipedia} 
            & Amazon & 8/16-17, 8/23-24 \\
            &  & Deepinfra & 8/16, 8/23-24 \\
            &  & Fireworks & 8/16, 8/23-24 \\
            &  & Together & 8/16, 8/20, 8/23-24 \\
        \cmidrule(lr){2-4}
        & \multirow{7}{*}{\humaneval} 
            & Amazon & 8/24-25 \\
            &  & Deepinfra & 8/24 \\
            &  & Fireworks & 8/24 \\
            &  & Together & 8/24 \\
        \cmidrule(lr){2-4}
        & \multirow{7}{*}{\ultrachat} 
            & Amazon & 8/24-25 \\
            &  & Deepinfra & 8/24-25 \\
            &  & Fireworks & 8/24 \\
            &  & Together & 8/24 \\
        \bottomrule
    \end{tabular}}
    \caption{
        Dates we queried \llama-3.1 405B inference endpoints.
    }
    \label{tab:3.1_405b_api_dates}
\end{table}

\begin{table}[h]
    \centering
    \scriptsize
    \begin{tabular}{lc}
        \toprule
        \textbf{Model} & \textbf{Dates queried} \\
        \midrule
        \texttt{gpt-4o-mini} & 8/21, 8/23-24 \\
        \texttt{gpt-4o} & 8/29 \\
        \texttt{gpt-3.5-turbo-0125} & 8/29 \\
        \texttt{gpt-3.5-turbo-1106} & 8/29 \\
        \texttt{gpt-4-0125-preview} & 8/29 \\
        \bottomrule
    \end{tabular}
    \caption{
        Dates we queried OpenAI endpoints (Wikipedia task).
    }
    \label{tab:openai_api_dates}
\end{table}

\subsection{Monte carlo simulations}
To construct the ten Wikipedia prompt distributions in \S\ref{sec:simulations}, we randomly sampled 25 prompts per distribution from the Wikipedia prompts in our dataset.
The \humaneval and \ultrachat prompt distributions were constructed by using all available prompts from those sources.

In most experiments, we estimated power as the average rejection rate over 100 simulations, 
where we sample a fresh $\samplelocal$ and $\sampleapi$ each time.
We simulated p-values by sampling \num{1000} datasets $\samplelocal$ and $\sampleapi$ from $\local$ and computing the test statistic on each pair,
and then we reused this empirical distribution when testing against all alternatives for the same $\prior\local$.
The exception is for the MMD all-substrings test statistic:
because this was exceptionally slow to compute, we simulated p-values using 100 samples instead of 1000, and we estimated power from 20 simulations instead of 100.

\FloatBarrier
\section{Additional results}\label{app:additional_results}
\subsection{Additional results from \S\ref{sec:quantization_watermarking}}\label{app:additional_results_4_1}
In \S\ref{sec:quantization_watermarking}, we evaluated two-sample tests on their ability to distinguish samples (in token space, $L \le 50$) from full-precision models from those of quantized models and watermarked models.
Concretely, we evaluated tests' power on pairs of distributions $(\local, \api)$, 
where $\local$ represents samples from the \full model and $\api$ represents samples from one of \{\half, \inteight, \nf, or watermarked\} versions of the same model.
To ensure generalizable results, we experimented with five language models (\mistral 7B Instruct, \llama-3 8B and 70B Instruct, and \llama-3.1 8B and 70B Instruct) and ten Wikipedia-based prompt distributions (Box \ref{box:wikipedia}).

Figure \ref{fig:sample_complexity_length} in the main text plotted sample complexity and length results, \textit{averaging across language models and prompt distributions}.
In this appendix, we show plots for the same experiments, but \textit{stratified by language model and prompt distribution}.

\begin{figure}[h]
    \centering
    \includegraphics[width=\textwidth]{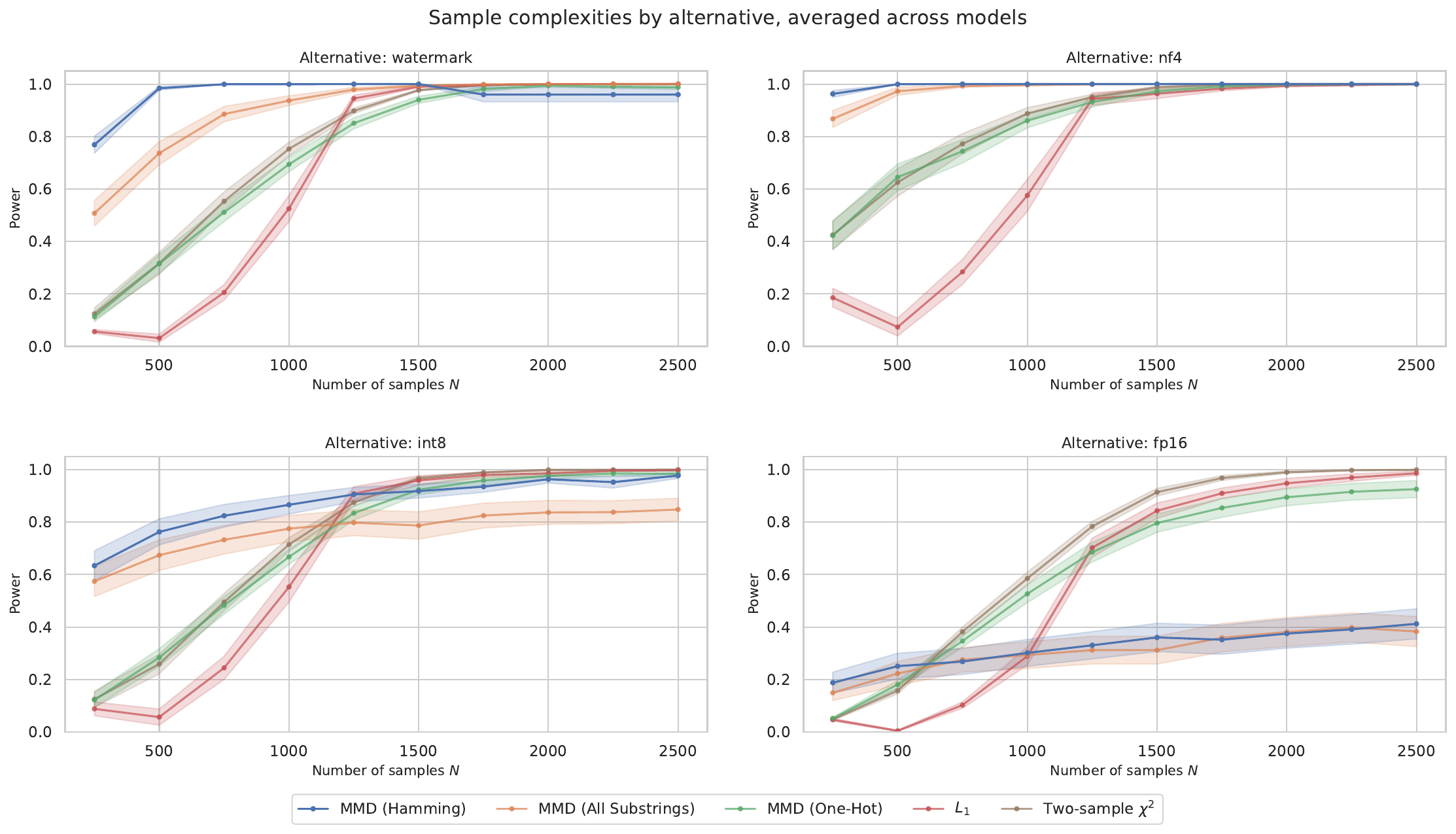}
    \caption{
        Sample complexities for different two-sample tests. Each subplot represents power for an alternative distribution $\api$ averaged over five language models and ten prompt distributions $\prior$.
        Tests are conducted with sample sizes ranging from $N=10m$ to $N=100m$, where $m=25$ is the number of prompts in the prompt distribution.
        The completion length is fixed to $L=50$ tokens.
        The Hamming MMD test is significantly more sample-efficient for all alternatives except \half quantization: for this alternative, while other two-sample tests can 
        attain perfect power with enough samples, the kernel test increases power slowly. This suggests that \half and \full differ in ways that are difficult to capture with the Hamming kernel.
    }
    \label{fig:sample_complexity}
\end{figure}

\begin{figure}[h]
    \centering
    \includegraphics[width=\textwidth]{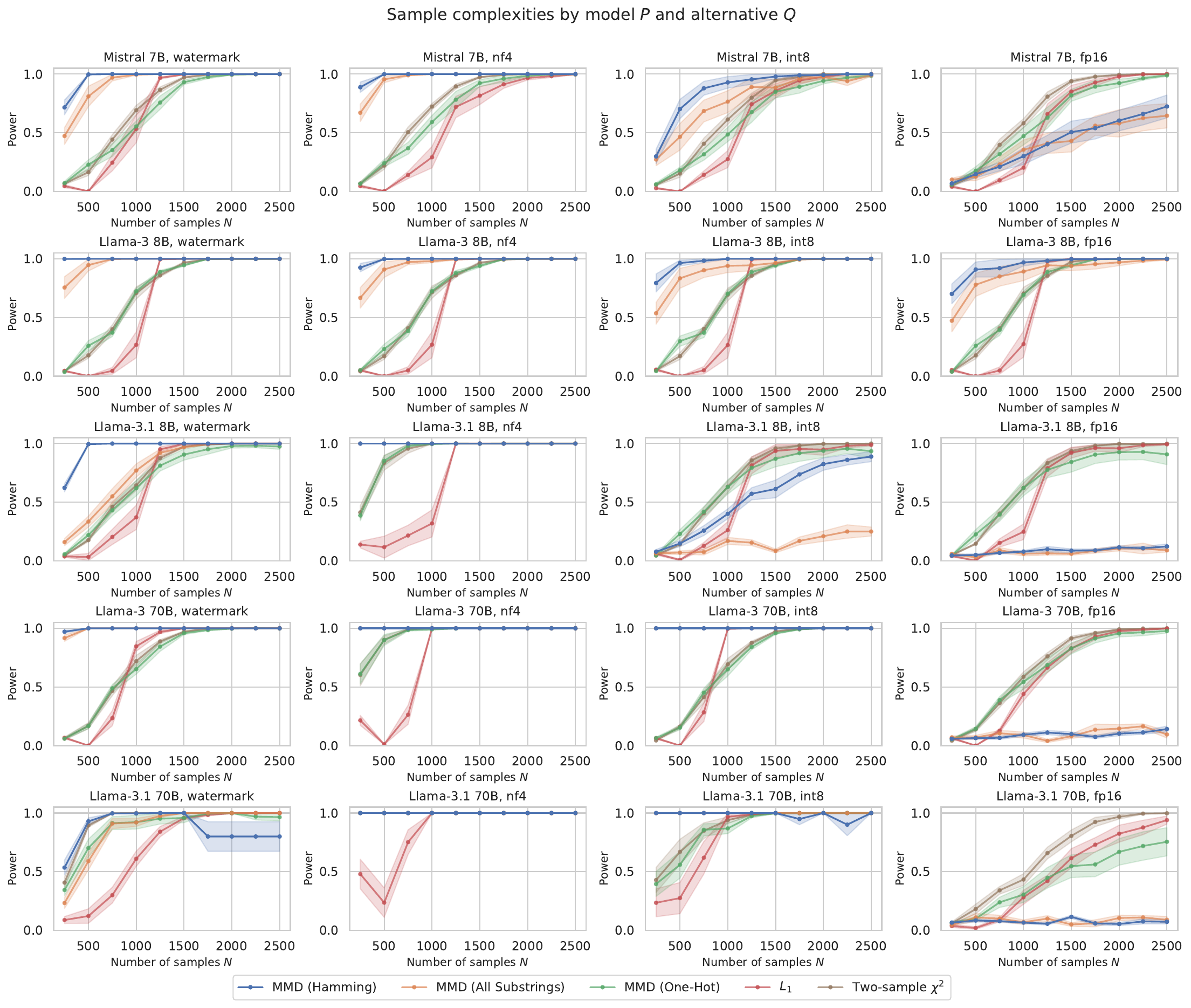}
    \caption{
        Sample complexities for different two-sample tests. Each subplot represents power for a particular alternative distribution $\api$ and model, but averaged over ten prompt distributions $\prior$. 
        Tests are conducted with sample sizes ranging from $N=10m$ to $N=100m$, where $m=25$ is the number of prompts in the prompt distribution.
        The completion length is fixed to $L=50$ tokens.
        Some model and alternative combinations are more difficult to detect than others.
    }
    \label{fig:sample_complexity_stratified}
\end{figure}

\begin{figure}[h]
    \centering
    \includegraphics[width=\textwidth]{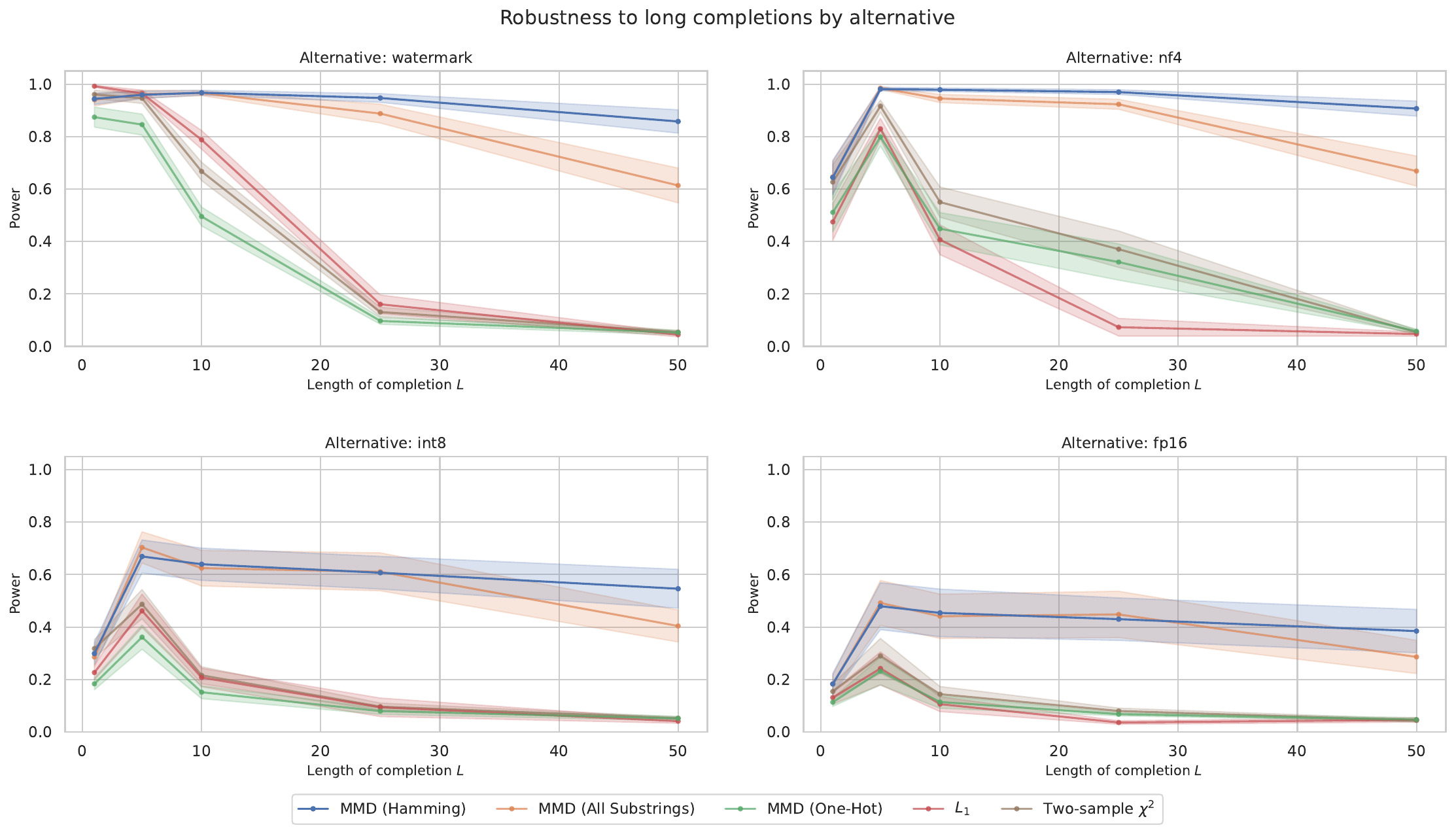}
    \caption{
        Simulated powers for different completion lengths $L \in \{1, \cdots, 50\}$, stratified by the alternative distribution $\api$. 
        The sample size is fixed to $N=10m$, where $m=25$ is the number of prompts in the prompt distribution.
        Across all alternatives, kernel tests suffer less drop in power as the length of completions increases.
        Traditional two-sample tests observe a slight power increase from $L=1$ tokens to $L=5$, tokens, but power
        dramatically degrades afterwards.
    }
    \label{fig:length_stratified}
\end{figure}

\begin{itemize}
    \item Figure \ref{fig:sample_complexity} shows the sample complexities for different two-sample tests, stratified by the alternative distribution $\api$, but averaged over the five language models and ten prompt distributions. 
    \item Figure \ref{fig:sample_complexity_stratified} shows the sample complexities for different two-sample tests, stratified by the alternative distribution $\api$ and model, but averaged over ten prompt distributions.
    \item Figure \ref{fig:length_stratified} shows the simulated powers for different completion lengths $L$, stratified by the alternative distribution $\api$.
\end{itemize}

Additionally, Table \ref{tab:other_llms} provides the power of the Hamming MMD test to distinguish $(\local, \api)$ for other language models in $L=50$ token space.
This table adds Phi-3 Mini (Instruct), OLMo 7B (Instruct), and Gemma-2 9B (Instruct).
Table \ref{tab:lev} evaluates a Levenshtein MMD test, which uses the Levenshtein distance as the kernel function.
The Levenshtein kernel is defined as $k(x, y) = \exp(-\text{Levenshtein}(x, y))$.
It is less powerful than the Hamming kernel, but still outperforms traditional two-sample tests.

\paragraph{Discussion: the Hamming MMD is the most powerful test.} 
Across Figure \ref{fig:sample_complexity_length} (left), Figure \ref{fig:sample_complexity}, and Figure \ref{fig:sample_complexity_stratified}, we observe that the Hamming MMD test is significantly more sample-efficient for all alternatives except \half quantization at completion length $L=50$ tokens. 
We also observe that the Hamming MMD is more robust to changes in completion length than other tests (Figure \ref{fig:sample_complexity_length} middle, Figure \ref{fig:length_stratified}).

\paragraph{Discussion: tests' powers depend on the particular $(\local, \api)$ pair.}
In Figure \ref{fig:sample_complexity_stratified} and Table \ref{tab:other_llms}, we observe that tests' powers are heterogeneous across different $(\local, \api)$ pairs.
The choice of model affects both the reference distribution $\local$ and the effect of interventions like quantization, \ie the alternative $\api$. 
For example, 4-bit quantization of the Llama and OLMo models is consistently noticeable. 
On the other hand, 8-bit quantization and watermarking have more inconsistent effects.
The impact of quantization is heterogeneous even within the same model family: for example, the Llama-3 8B model is more affected by \inteight quantization than the Llama-3.1 8B model.
On average however, as discussed above, our results above still support our claim that the Hamming MMD is more powerful than other kernel choices.

\begin{table}[t]
    \centering
    \footnotesize
    \begin{tabular}{lcccc}
        \toprule
        Model & watermark & nf4 & int8 & FPR (\full) \\
        \midrule
        Llama-3 8B & 1.00 (0.00) & 0.93 (0.04) & 0.79 (0.08) & 0.05 (0.01) \\
        Llama-3.1 8B & 0.62 (0.04) & 1.00 (0.00) & 0.08 (0.02) & 0.05 (0.01) \\
        Llama-3 70B & 0.97 (0.01) & 1.00 (0.00) & 1.00 (0.00) & 0.07 (0.01) \\
        Llama-3.1 70B & 0.54 (0.06) & 1.00 (0.00) & 1.00 (0.00) & 0.06 (0.01) \\
        Mistral 7B & 0.72 (0.06) & 0.89 (0.04) & 0.30 (0.06) & 0.04 (0.01) \\
        OLMo 7B & 0.47 (0.08) & 0.99 (0.01) & 0.36 (0.08) & 0.06 (0.01) \\
        Gemma-2 9B & 0.43 (0.02) & 0.12 (0.01) & 0.06 (0.01) & 0.06 (0.01) \\
        Phi-3 Mini & 0.75 (0.04) & 0.63 (0.05) & 0.24 (0.01) & 0.06 (0.01) \\
        \bottomrule
    \end{tabular}
    \caption{
        Power of the Hamming MMD test to distinguish the \full model from alternatives in $L=50$ token space at $N=10m$ samples, where $m=25$ is the number of prompts in the prompt distribution. 
        Averages and standard errors are reported over ten prompt distributions.
        The column represents the alternative distribution $\api$. The FPR (\full) column represents the false positive rate when comparing the full-precision model to itself.
    }
    \label{tab:other_llms}
\end{table}

\begin{table}[t]
    \centering
    \footnotesize
    \begin{tabular}{lcccc}
        \toprule
        Model & watermark & nf4 & int8 & FPR (\full) \\
        \midrule
        Llama-3 8B & 0.14 (0.01) & 0.15 (0.01) & 0.16 (0.01) & 0.07 (0.01) \\
        Llama-3 70B & 0.13 (0.02) & 1.00 (0.00) & 0.12 (0.01) & 0.04 (0.01) \\
        Mistral 7B & 0.11 (0.01) & 0.19 (0.01) & 0.12 (0.01) & 0.05 (0.01) \\
        \bottomrule
    \end{tabular}
    \caption{
        Power of the Levenshtein MMD test to distinguish the \full model from alternatives in $L=50$ token space at $N=10m$ samples, where $m=25$ is the number of prompts in the prompt distribution. 
        Averages and standard errors are reported over ten prompt distributions.
        The column represents the alternative distribution $\api$. The FPR (\full) column represents the false positive rate when comparing the full-precision model to itself.
    }
    \label{tab:lev}
\end{table}

\subsection{Additional results from \S\ref{sec:model_model}}\label{app:higher_dimensional}
In \S\ref{sec:model_model}, we moved from evaluating tests in an easier token space to a higher-dimensional character space.
Specifically,  in \S\ref{sec:quantization_watermarking} and \S\ref{sec:finetuning}, we evaluated tests in $L=50$ token space, where $|\mathcal V| = \num{128256}$.
In \S\ref{sec:model_model} and \S\ref{sec:apis}, we evaluated tests in $L=\num{1000}$ or $L=\num{5000}$ character space, where $|\mathcal V| = \num{155063}$.
In general, since the set of all prompt-completion pairs has size $O(m|\mathcal V|^L)$, we expect testing to be harder in the latter, \textit{significantly} higher-dimensional space.
One can come to the same conclusion by extrapolating from Figure \ref{fig:sample_complexity_length} (middle) in the main text; 
this figure also shows that the Hamming MMD test is more robust to this increase in dimensionality than other tests.

\paragraph{Discussion: power decreases in higher-dimensional space.}
Table \ref{tab:char_power} shows the power of the Hamming MMD test to distinguish between pairs of distributions $(\local, \api)$ in $L=\num{1000}$ character space.
As expected, moving to this higher-dimensional space decreases the power of the test; 
the powers in the last three rows are lower than those in Table \ref{tab:other_llms} for the same models in $L=50$ token space.
These power losses can be compensated for by increasing the sample size:
Table \ref{tab:char_power_more_samples} repeats the experiments in Table \ref{tab:char_power} with $N=50m$ instead of $N=10m$. 
If we were to use this sample size in \S\ref{sec:apis}, each audit would still cost $<$ \$5.

Note that even in the reduced power setting, model swaps are easy to detect (Table \ref{tab:char_power}, top rows).

\begin{table}[t]
    \centering
    \footnotesize
    \begin{tabular}{llllll}
        \toprule
        model & Mistral 7B & Llama-3 8B & Llama-3.1 8B & Llama-3 70B & Llama-3.1 70B \\
        alternative &  &  &  &  &  \\
        \midrule
        Mistral 7B & \textit{0.06} (\textit{0.01}) & 1.00 (0.00) & 1.00 (0.00) & 1.00 (0.00) & 1.00 (0.00) \\
        Llama-3 8B & 1.00 (0.00) & \textit{0.05} (\textit{0.01}) & 0.76 (0.00) & 0.98 (0.00) & 0.95 (0.00) \\
        Llama-3.1 8B & 1.00 (0.00) & 0.83 (0.05) & \textit{0.07} (\textit{0.01}) & 1.00 (0.00) & 0.53 (0.00) \\
        Llama-3 70B & 1.00 (0.00) & 0.99 (0.00) & 1.00 (0.00) & \textit{0.07} (\textit{0.01}) & 0.89 (0.00) \\
        Llama-3.1 70B & 1.00 (0.00) & 0.98 (0.00) & 0.75 (0.07) & 0.99 (0.00) & \textit{0.06} (\textit{0.01}) \\
        Phi-3 mini & 1.00 (0.00) & 1.00 (0.00) & 1.00 (0.00) & 1.00 (0.00) & 1.00 (0.00) \\
        Gemma-2 9B & 1.00 (0.00) & 1.00 (0.00) & 1.00 (0.00) & 1.00 (0.00) & 1.00 (0.00) \\
        OLMo 7B & 1.00 (0.00) & 1.00 (0.00) & 1.00 (0.00) & 1.00 (0.00) & 1.00 (0.00) \\
        gpt-4o-mini & 1.00 (0.00) & 1.00 (0.00) & 1.00 (0.00) & 1.00 (0.00) & 1.00 (0.00) \\
        gpt-4o & 1.00 (0.00) & 1.00 (0.00) & 1.00 (0.00) & 1.00 (0.00) & 1.00 (0.00) \\
        gpt-3.5-turbo-0125 & 1.00 (0.00) & 1.00 (0.00) & 1.00 (0.00) & 1.00 (0.00) & 1.00 (0.00) \\
        gpt-3.5-turbo-1106 & 1.00 (0.00) & 1.00 (0.00) & 1.00 (0.00) & 1.00 (0.00) & 1.00 (0.00) \\
        gpt-4-0125-preview & 1.00 (0.00) & 1.00 (0.00) & 1.00 (0.00) & 1.00 (0.00) & 1.00 (0.00) \\
        watermark & 0.23 (0.02) & 0.62 (0.05) & 0.32 (0.04) & 0.57 (0.06) & 0.26 (0.03) \\
        int8 & 0.15 (0.02) & 0.30 (0.06) & 0.07 (0.01) & 1.00 (0.00) & 0.99 (0.01) \\
        nf4 & 0.44 (0.05) & 0.38 (0.05) & 1.00 (0.00) & 1.00 (0.00) & 1.00 (0.00) \\
        \bottomrule
    \end{tabular}
    \caption{Power of the Hamming MMD test to distinguish between pairs of models (and other alternatives) in $L=\num{1000}$ character space at $N=10m$, where $m=25$ is the number of prompts covered by $\pi$. The column represents the null distribution $\local$. Italicized entries are FPRs. Note that unlike \S\ref{sec:quantization_watermarking}, the test is now run in higher-dimensional character space, which generally reduces power.}
    \label{tab:char_power}
\end{table}

\begin{table}[t]
    \centering
    \footnotesize
    \begin{tabular}{lcccc}
        \toprule
        Sample size & watermark & nf4 & int8 & FPR (\full) \\
        \midrule
        $N=10m$ & 0.62 (0.05) & 0.38 (0.05) & 0.30 (0.06) & 0.05 (0.01) \\
        $N=50m$ & 1.00 (0.00) & 1.00 (0.01) & 0.96 (0.02) & 0.07 (0.01) \\
        \bottomrule
    \end{tabular}
    \caption{Power of the Hamming MMD test to distinguish between the \full \llama-3 8B and quantized or watermarked alternatives in $L=\num{1000}$ character space. We compare power at $N=10m$ and $N=50m$, where $m=25$ is the number of prompts covered by $\pi$. Increasing the sample size can compensate for losses of power.}
    \label{tab:char_power_more_samples}
\end{table}

\subsection{Detecting incorrect decoding parameters}
Our problem formulation in \S\ref{sec:problem} compares two distributions in general:
this includes cases where the API $\api$ samples from the same {model} as the reference distribution $\local$, but with different decoding parameters (\eg temperature $\tau$).
Table \ref{tab:temperature} shows that detecting differences in the temperature parameter is quite easy; 
this parameter significantly affects the distribution of completions.

\begin{table}[t]
    \centering
    \footnotesize
    \begin{tabular}{lcc}
        \toprule
        Model & $\tau=0.5$ & $\tau=1.5$ \\
        \midrule
        \llama-3 8B & 1.00 (0.00) & 1.00 (0.00) \\
        \llama-3.1 8B & 1.00 (0.00) & 1.00 (0.00) \\
        Mistral 8B & 1.00 (0.00) & 0.97 (0.02) \\
        \bottomrule
    \end{tabular}
    \caption{Power of the Hamming MMD test to distinguish between models sampled at temperature $\tau=1$ ($\local$) vs. other temperatures ($\api$). This decoding parameter mismatch can be detected with near-perfect power.}
    \label{tab:temperature}
\end{table}

\FloatBarrier

\subsection{Effect of the prompt distribution}
In \S\ref{sec:simulations}, we evaluated tests on a prompt distribution $\pi$ drawn from a Wikipedia language modeling task and supported on $m=25$ prompts. 
In this appendix, we extend these evaluations to different prompt distributions.

\begin{figure}[h]
    \centering
    \includegraphics[width=0.7\textwidth]{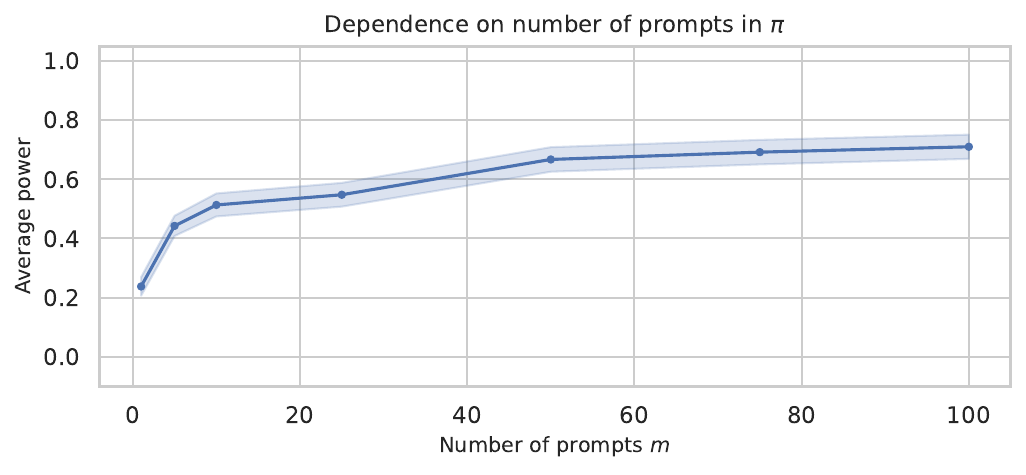}
    \caption{
        Power of the Hamming MMD test for different numbers of prompts $m$ in the prompt distribution.
        Results are averaged over ten random samples of $m$ prompts (for the Wikipedia task).
        The test is run with $N=10m$ and $L=50$ tokens, with $m$ varying from 1 to 100.
        Power increases with $m$, suggesting that users benefit from testing many prompts together.
    }
    \label{fig:prompt_distribution_m}
\end{figure}

\paragraph{Effect of the number of prompts $m$.}
In Figure \ref{fig:prompt_distribution_m}, we continue to use the Wikipedia task, but vary the number of prompts $m$ in the prompt distribution.
Larger $m$ values increase the power of the test, suggesting that users benefit from testing many prompts together, so long as the sample size is increased proportionally.

\paragraph{Effect of the task.}
We explored whether the ``open-endedness'' of the task affects power: 
intuitively, one might expect that more creative tasks lead to completion distributions that are higher entropy, which might make testing harder.
To test this, we evaluated power across several prompt distributions:
on the most constrained side of the spectrum, we experimented with \humaneval (code, Box \ref{box:humaneval}); on the most creative side, \ultrachat (chatbot dialogues, Box \ref{box:ultrachat}), and in the middle, language modeling with Wikipedia (Box \ref{box:wikipedia}).
We additionally experimented with tightly concentrated prompts (just Wikipedia in English, just Wikipedia in French, etc.) vs. diverse prompt distributions (Wikipedia mixed with \humaneval and \ultrachat). 

Table \ref{tab:prompt_distribution_task} lists the Hamming MMD's power against the local alternatives (nf4, int8, watermark) at $N=10m$ with tests conducted in token space ($L=50$ token completions). We make several observations:

\begin{itemize}
\item We do not observe that creative tasks are harder to test than constrained tasks: if this were the case, we would expect the power to be lowest for \ultrachat and highest for \humaneval across all alternatives.
\item We also do not observe that diverse prompt distributions are harder to test than concentrated prompt distributions: if this were the case, we would expect the power to be higher for the single-language Wikipedia prompt distributions than the mixed-task prompt distributions. 
\item The prompt distribution mainly affects the difficulty of detecting quantization, not watermarking. 
\item For comparison, we also provide the results of the one-hot kernel in Table \ref{tab:prompt_distribution_task_onehot}. The Hamming kernel consistently outperforms the one-hot baseline on all prompt distributions, matching our main conclusion of \S\ref{sec:quantization_watermarking} about the relative strength of kernels.
\end{itemize}

\begin{table}[t]
    \centering
    \footnotesize
    \begin{tabular}{llccccc}
        \toprule
        & & $m$ & watermark & \nf & \inteight & FPR (\full) \\
        \midrule
        \humaneval (constrained) & & 20 & 0.99 (0.00) & 1.00 (0.00) & 0.32 (0.00) & 0.06 (0.00) \\
        \midrule
        \multirow{6}{*}{Wikipedia (language modeling)} 
        & English only & 25 & 0.98 (0.01) & 0.49 (0.04) & 0.11 (0.01) & 0.06 (0.01) \\
        & French only & 25 & 1.00 (0.00) & 1.00 (0.00) & 1.00 (0.00) & 0.06 (0.01) \\
        & German only & 25 & 0.98 (0.01) & 0.71 (0.08) & 0.27 (0.06) & 0.05 (0.00) \\
        & Spanish only & 25 & 1.00 (0.00) & 1.00 (0.00) & 1.00 (0.00) & 0.05 (0.01) \\
        & Russian only & 25 & 0.98 (0.01) & 0.71 (0.05) & 0.24 (0.04) & 0.05 (0.01) \\
        & All languages & 25 & 1.00 (0.00) & 0.93 (0.04) & 0.79 (0.08) & 0.05 (0.01) \\
        \midrule
        \ultrachat (creative) & & 20 & 0.89 (0.00) & 1.00 (0.00) & 0.04 (0.00) & 0.04 (0.00) \\
        \midrule
        Wikipedia + \ultrachat & & 25 & 0.81 (0.05) & 1.00 (0.00) & 0.10 (0.02) & 0.06 (0.01) \\
        Wikipedia + \ultrachat + \humaneval & & 25 & 0.95 (0.01) & 1.00 (0.00) & 0.35 (0.05) & 0.04 (0.01) \\
        \bottomrule
    \end{tabular}    
    \caption{Hamming MMD's power against the local alternatives (\nf, \inteight, watermark) at $N=10m$ with tests conducted in token space ($L=50$ token completions).}
    \label{tab:prompt_distribution_task}
\end{table}

\begin{table}[t]
    \centering
    \footnotesize
    \begin{tabular}{llccccc}
        \toprule
        & & $m$ & watermark & \nf & \inteight & FPR (\full) \\
        \midrule
        \humaneval (constrained) & & 20 & 0.46 (0.00) & 0.77 (0.00) & 0.16 (0.00) & 0.05 (0.00) \\
        \midrule
        \multirow{6}{*}{Wikipedia (language modeling)} 
        & English only & 25 & 0.06 (0.01) & 0.05 (0.01) & 0.04 (0.01) & 0.01 (0.00) \\
        & French only & 25 & 0.04 (0.01) & 0.06 (0.01) & 0.05 (0.00) & 0.01 (0.00) \\
        & German only & 25 & 0.04 (0.01) & 0.04 (0.01) & 0.04 (0.01) & 0.01 (0.00) \\
        & Spanish only & 25 & 0.04 (0.01) & 0.06 (0.01) & 0.04 (0.01) & 0.02 (0.00) \\
        & Russian only & 25 & 0.04 (0.01) & 0.05 (0.01) & 0.03 (0.01) & 0.01 (0.00) \\
        & All languages & 25 & 0.04 (0.00) & 0.05 (0.01) & 0.04 (0.01) & 0.01 (0.00) \\
        \midrule
        \ultrachat (creative) & & 20 & 0.07 (0.00) & 0.48 (0.00) & 0.04 (0.00) & 0.01 (0.00) \\
        \midrule
        Wikipedia + \ultrachat & & 25 & 0.09 (0.01) & 0.05 (0.01) & 0.07 (0.01) & 0.04 (0.01) \\
        Wikipedia + \ultrachat + \humaneval & & 25 & 0.22 (0.02) & 0.32 (0.03) & 0.05 (0.01) & 0.05 (0.01) \\
        \bottomrule
    \end{tabular}
    \caption{One-hot MMD's power against the local alternatives (\nf, \inteight, watermark) at $N=10m$ with tests conducted in token space ($L=50$ token completions).}
    \label{tab:prompt_distribution_task_onehot}
\end{table}

\FloatBarrier
\subsection{comparing two-sample and approximate goodness-of-fit tests}\label{app:gof}
In \S\ref{sec:problem}, we assumed only sample access to both the reference distribution $\local$ and API $\api$. 
An alternative problem setup might give the auditor privileged access to $\local$ to evaluate probabilities $P(y\mid x)$ for arbitrary (prompt, completion) pairs.
Here, we compare the performance of two-sample tests to goodness-of-fit tests that leverage this privileged setting.

In an extreme case, evaluating probabilities is free -- the auditor can fully describe $P(y\mid x)$ for all completions $y \in \mathcal V^{\le L}$.
Then goodness-of-fit tests, like the one-sample $L_1$ statistic, can be used to compare observed counts in $\sampleapi$ to expected counts under $\local$:
\begin{equation}
    \text{L1}_\text{gof}(\local, \sampleapi)
    = \sum_{z \in \mathcal X \times \mathcal V^{\le L}} \left|
        \obs{z}{\sampleapi} - N \cdot \local(y|x)\prior(x)
    \right|.
\end{equation}
However, in practice, evaluating $\local (y \mid x)$ for all $y \in \mathcal V^{\le L}$ and all $x$ is intractable: 
as a concrete example, for our language modeling task on \llama-3, $m \cdot |\mathcal V^{\le L}| = 25 \cdot \num{128000}^{50} \approx 5 \times 10^{256}$.
A more realistic scenario is that the auditor can only evaluate $\local(y \mid x)$ for the observed $(x,y)$ in $\sampleapi$.
This leads to an approximation of the goodness-of-fit test statistic:
\begin{equation}
    \widehat{\text{L1}_\text{gof}}(\local, \sampleapi)
    = \sum_{z \in \sampleapi} \left|
        \obs{z}{\sampleapi} - N \cdot \local(y|x)\prior(x)
    \right|.
\end{equation}
We take a similar strategy for the one-sample $L_2$ test
\begin{equation}
    \begin{aligned}
    \text{L2}_\text{gof}(\local, \sampleapi)
    = \sum_{z \in \mathcal X \times \mathcal V^{\le L}} \left(
        \obs{z}{\sampleapi} - N \cdot \local(y|x)\prior(x)
    \right)^2
    \\
    \widehat{\text{L2}_\text{gof}}(\local, \sampleapi)
    = \sum_{z \in \sampleapi} \left(
        \obs{z}{\sampleapi} - N \cdot \local(y|x)\prior(x)
    \right)^2,
    \end{aligned}
\end{equation}
the Pearson $\chi^2$ test
\begin{equation}
    \begin{aligned}
    \chi^2_\text{pearson}(\local, \sampleapi)
    = \sum_{z \in \mathcal X \times \mathcal V^{\le L}} \frac{\left(
        \obs{z}{\sampleapi} - N \cdot \local(y|x)\prior(x)
    \right)^2}{N \cdot \local(y|x)\prior(x)}
    \\
    \widehat{\chi^2_\text{pearson}}(\local, \sampleapi)
    = \sum_{z \in \sampleapi} \frac{\left(
        \obs{z}{\sampleapi} - N \cdot \local(y|x)\prior(x)
    \right)^2}{N \cdot \local(y|x)\prior(x)},
    \end{aligned}
\end{equation}
and the truncated $\chi^2$ test \citep{balakrishnan2018hypothesis}
\begin{equation}
    \begin{aligned}
    \chi^2_\text{truncated}(\local, \sampleapi)
    = \sum_{z \in \mathcal X \times \mathcal V^{\le L}} \frac{\left(
        \obs{z}{\sampleapi} - N \cdot \local(y|x)\prior(x)
    \right)^2 - \obs{z}{\sampleapi}}{\max \left( \local(y|x)\prior(x), \frac{1}{|\mathcal V^{\le L}|} \right)}
    \\
    \widehat{\chi^2_\text{truncated}}(\local, \sampleapi)
    = \sum_{z \in \sampleapi} \frac{\left(
        \obs{z}{\sampleapi} - N \cdot \local(y|x)\prior(x)
    \right)^2- \obs{z}{\sampleapi}}{\max \left( \local(y|x)\prior(x), \frac{1}{|\mathcal V^{\le L}|} \right)}.
    \end{aligned}
\end{equation}
The only goodness-of-fit test that we consider which is unaffected by the approximation is the likelihood ratio test:
\begin{equation}
\text{LRT}(\local, \sampleapi)
= -2 \sum_{z \in \sampleapi} \obs{z}{\sampleapi} \log \left(
    \frac{\obs{z}{\sampleapi}}{N \cdot \local(y|x)\prior(x)}
\right).
\end{equation}

\begin{figure}[h]
    \centering
    \includegraphics[width=\textwidth]{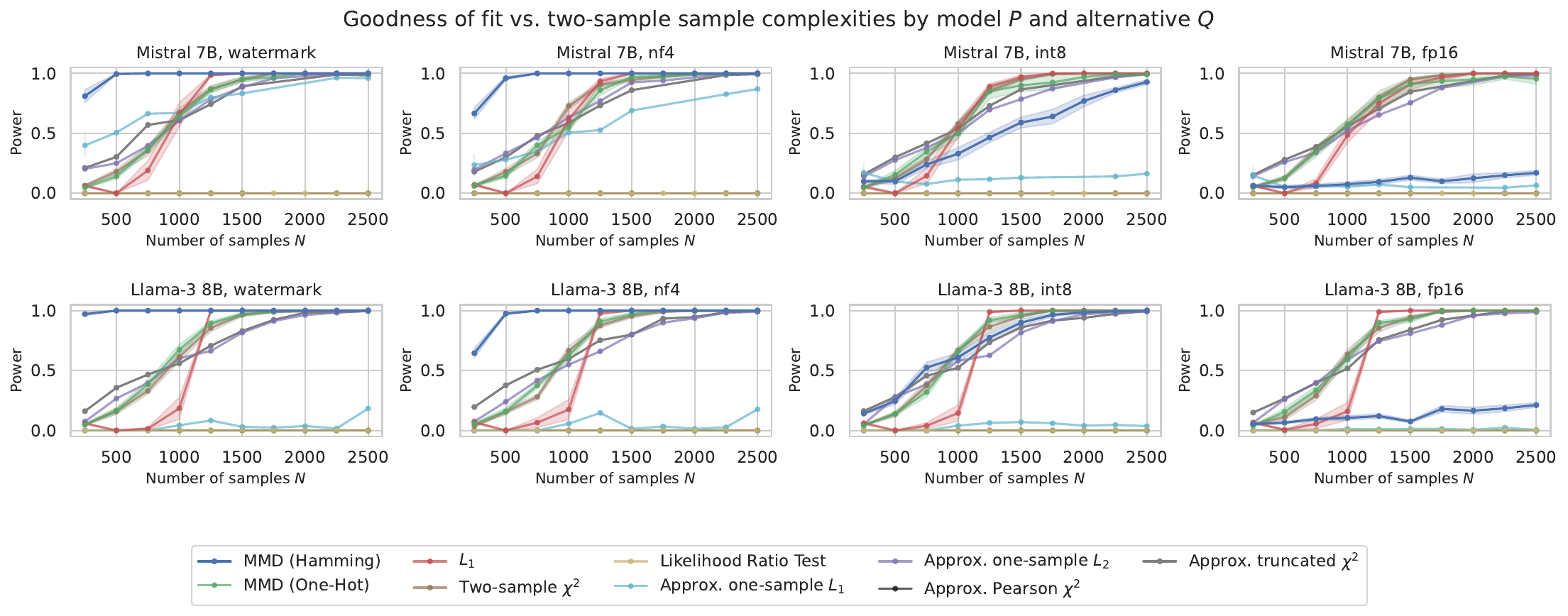}
    \caption{
        Power of two-sample MMD tests and goodness-of-fit tests. Each subplot represents a particular language model and alternative distribution $\api$.
        The goodness-of-fit tests are approximated by evaluating $\local(y \mid x)$ only for observed $(x,y)$ in $\sampleapi$.
        The truncated chi-squared and one-sample $L_2$ tests perform best out of the goodness-of-fit tests, while the likelihood ratio test and one-sample $L_1$ tests perform worst.
        Note that experiments were run on a different set of 10 prompt distributions than the main text; these prompts specifically had model probabilities saved (see Appendix \ref{app:dataset}).
    }
    \label{fig:gof}
\end{figure}

Figure \ref{fig:gof} plots sample complexities of these (approximate) goodness-of-fit tests alongside the two-sample tests evaluated in the main text.
The best goodness-of-fit tests outperform their two-sample counterparts in the extremely low-sample regime ($N < \num{1000}$), but this trend reverses as $N$ increases.
This is surprising --- in theory, we would expect probability access to only increase power.
These results suggest that the approximations compensating for limited evaluation budget can introduce bias, reducing the power of goodness-of-fit tests.
We leave to future work ideas for the correction of this bias.

\FloatBarrier
\subsection{asymmetric sampling costs}\label{app:asymmetric}
\begin{figure}[h]
    \centering
    \includegraphics[width=\textwidth]{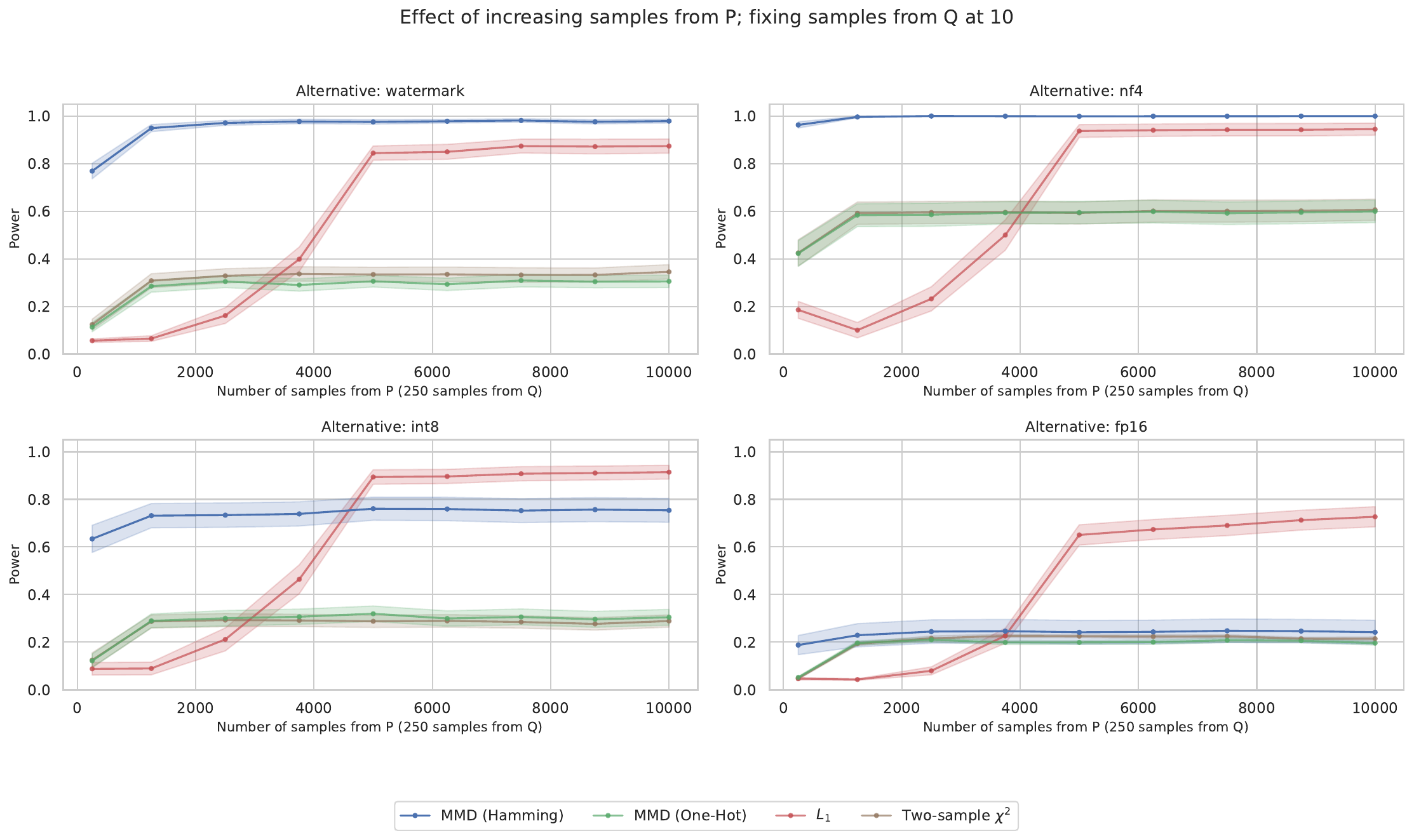}
    \caption{
        Power when the sample size from $\local$ increases, when the sample size from $\api$ is fixed to 250. There are slight power gains for all test statistics, with the $L_1$ test seeing the largest gains.
    }
    \label{fig:asymmetric}
\end{figure}

In some cases, it may be significantly less expensive to sample from one distribution than the other.
For example, the auditor may have unlimited compute to sample from the null distribution $\local$, but limited monetary budget to sample from the API $\api$.
In these cases, we show that it is possible to achieve slight power gains by increasing the sample size of the cheaper distribution, even while keeping the sample size of the expensive distribution fixed.
Figure \ref{fig:asymmetric} fixes $|\sampleapi| = 10m$ and varies $|\samplelocal|$ between $10m$ and $400m$.
All test statistics see some increases in power, with the $L_1$ test seeing surprisingly large gains.

\FloatBarrier

\subsection{Permutation procedure}\label{app:permutation}
\begin{figure}[h]
    \centering
    \includegraphics[width=\textwidth]{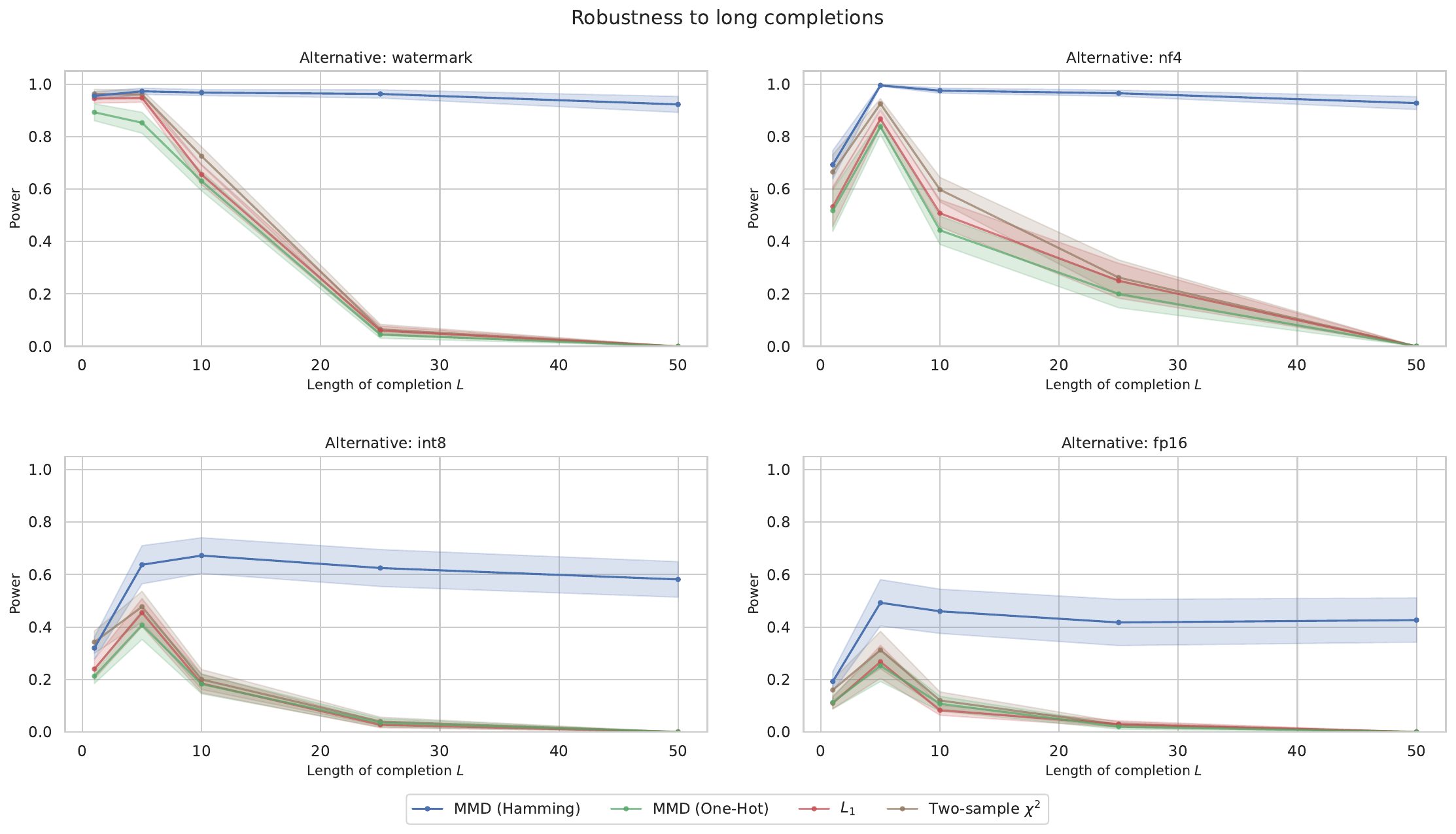}
    \caption{
        Simulated powers for different completion lengths $L$, stratified by the alternative distribution $\api$. 
        This figure parallels Figure \ref{fig:length_stratified} but uses the permutation procedure, rather than repeated sampling from $\local$, to estimate p-values.
    }
    \label{fig:length_permutation}
\end{figure}

\begin{figure}[h]
    \centering
    \includegraphics[width=\textwidth]{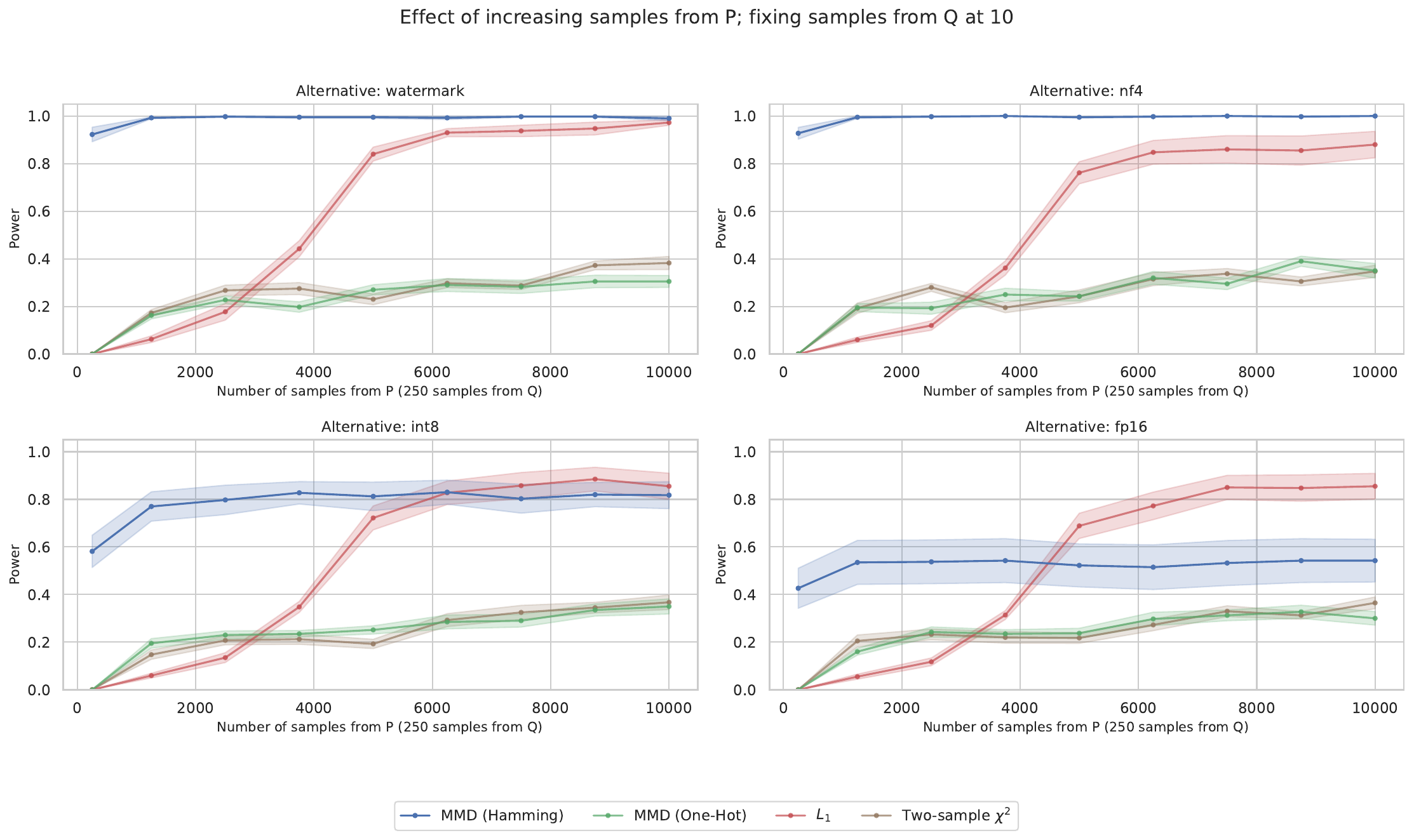}
    \caption{
        Simulated power when the sample size from $\local$ increases, when the sample size from $\api$ is fixed to 250.
        This figure parallels Figure \ref{fig:asymmetric} but uses the permutation procedure, rather than repeated sampling from $\local$, to estimate p-values.
    }
    \label{fig:asymmetric_permutation}
\end{figure}

Results in the main text simulate the test statistic's empirical distribution under the null by sampling datasets $\sampleapi$ and $\samplelocal$ both from $\local$.
Here we validate these trends by conducting the same tests using a permutation procedure to estimate p-values (see Appendix \ref{app:pvalues}).
Figures \ref{fig:sample_complexity_permutation}, \ref{fig:length_permutation}, and \ref{fig:asymmetric_permutation} repeat the sample complexity, length, and asymmetric sampling cost experiments, but use the permutation procedure to estimate p-values. 
Because of the computational complexity of this step, we use \num{100} permutations, estimate power using \num{20} simulations,
and only test Mistral 7B and \llama-3 8B.
We observe that the permutation procedure maintains similar power levels to the bootstrap method, 
and trends from the previous figures are replicated.

\begin{figure}[h]
    \centering
    \includegraphics[width=\textwidth]{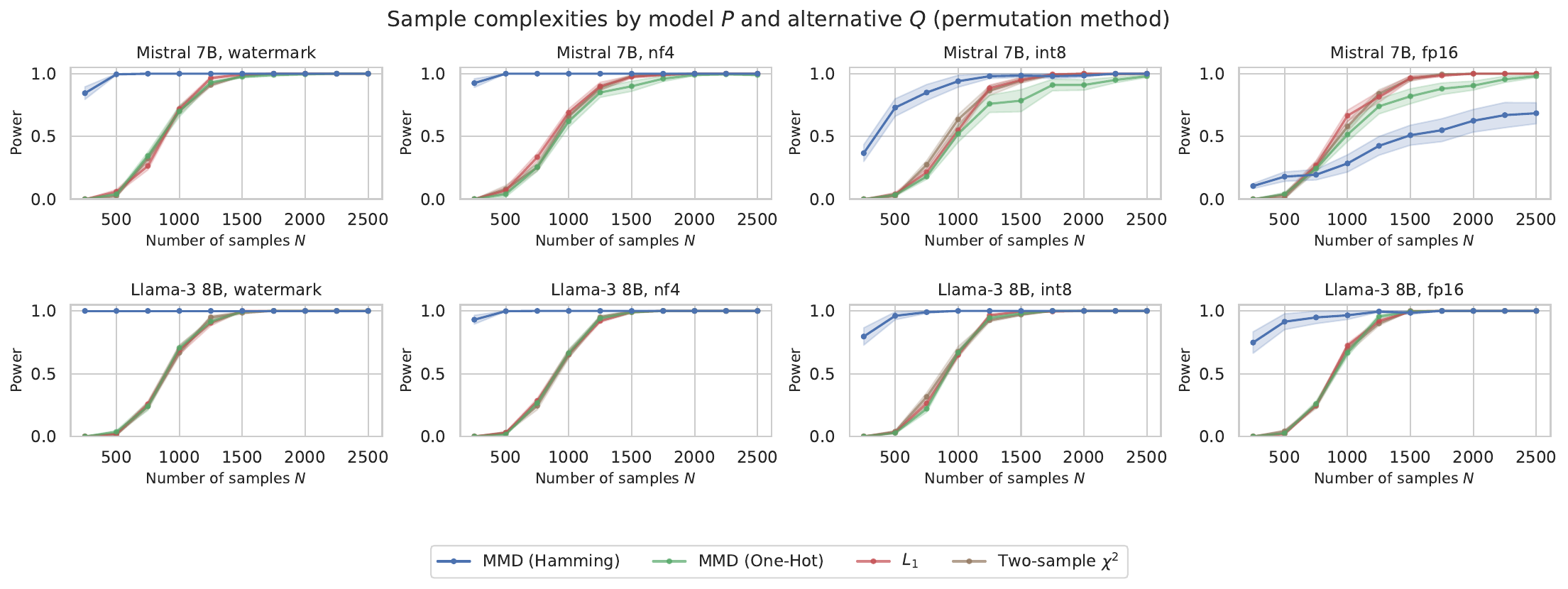}
    \caption{
        Sample complexities for different two-sample tests, stratified by the alternative distribution $\api$, but averaged over five language models and ten prompt distributions $\prior$. This figure parallels Figure \ref{fig:sample_complexity} but uses the permutation procedure, rather than repeated sampling from $\local$, to estimate p-values.
    }
    \label{fig:sample_complexity_permutation}
\end{figure}

\FloatBarrier
\subsection{Additional results from \S\ref{sec:apis}}

\begin{table}[h]
    \centering
    \small
    \begin{tabular}{lllllllll}
        \toprule
         & \multicolumn{4}{l}{Wikipedia} & \multicolumn{2}{l}{\humaneval} & \multicolumn{2}{l}{\ultrachat} \\
         & 3 8B & 3.1 8B & 3 70B & 3.1 70B & 3 8B & 3.1 8B & 3 8B & 3.1 8B \\
        \midrule
        Amazon & 0.07 & 1.00 & 0.32 & 1.00 & 0.48 & 1.00 & 1.00 & 1.00 \\
        Anyscale & 0.02 & --- & --- & --- & --- & --- & --- & --- \\
        Azure & 0.01 & 0.00 & 0.01 & 0.01 & 0.01 & 0.29 & 0.00 & 0.13 \\
        Deepinfra & 0.04 & 0.00 & 0.04 & 0.00 & 0.08 & 0.19 & 0.04 & 0.09 \\
        Fireworks & 0.04 & 0.04 & 0.01 & 0.01 & 0.01 & 1.00 & 0.03 & 0.90 \\
        Groq & 0.03 & 0.07 & 0.02 & 0.59 & 0.01 & 0.98 & 0.05 & 0.35 \\
        Perplexity & 1.00 & 1.00 & 1.00 & 1.00 & 1.00 & 1.00 & 1.00 & 1.00 \\
        Replicate & 0.12 & --- & 0.33 & --- & 0.07 & --- & 0.06 & --- \\
        Together & 0.01 & 0.00 & 0.00 & 0.00 & 0.00 & 0.27 & 0.01 & 0.07 \\
        watermark & 0.30 & 0.00 & 0.07 & 0.03 & 0.40 & 0.85 & 0.06 & 0.38 \\
        int8 & 0.07 & 0.00 & 1.00 & 0.96 & 0.01 & 0.45 & 0.01 & 0.03 \\
        nf4 & 0.09 & 1.00 & 1.00 & 1.00 & 1.00 & 1.00 & 1.00 & 1.00 \\
        \half & \textit{0.02} & \textit{0.01} & \textit{0.00} & \textit{0.01} & \textit{0.00} & \textit{0.02} & \textit{0.00} & \textit{0.04} \\
        \full & \textit{0.01} & \textit{0.00} & \textit{0.00} & \textit{0.00} & \textit{0.00} & \textit{0.10} & \textit{0.01} & \textit{0.04} \\
        \bottomrule
    \end{tabular}
    \caption{
        Power against alternatives for the full audit.
        Table \ref{tab:api_pass_fail} in the main text thresholds power at 0.5 for the APIs.
        The FPRs against the full- and half-precision nulls are italicized.
    }
    \label{tab:audit_power}
\end{table}

\begin{table}[h]
    \centering
    \small
    \begin{tabular}{lllll}
        \toprule
         & 3 8B & 3.1 8B & 3 70B & 3.1 70B \\
        \midrule
        Amazon & \xmark & \xmark & \xmark & \xmark \\
        Anyscale & \cmark & --- & --- & --- \\
        Azure & \cmark & \cmark & \cmark & \cmark \\
        Deepinfra & \cmark & \cmark & \cmark & \cmark \\
        Fireworks & \cmark & \xmark & \cmark & \cmark \\
        Groq & \cmark & \xmark & \cmark & \xmark \\
        Perplexity & \xmark & \xmark & \xmark & \xmark \\
        Replicate & \cmark & --- & \cmark & --- \\
        Together & \cmark & \cmark & \cmark & \cmark \\
        \bottomrule
    \end{tabular}
    \caption{
        Overall audit results, as also copied in Figure \ref{fig:hero}.
        The 8B models are tested on three prompt distributions (Wikipedia, \humaneval, \ultrachat), while the 70B models are tested on Wikipedia.
        Tests from different prompt distributions are combined using a Bonferroni correction.
    }
    \label{tab:audit_bonferroni_pass_fail}
\end{table}

\begin{table}[h]
    \centering
    \small
    \begin{tabular}{lllll}
        \toprule
        & 3 8B & 3.1 8B & 3 70B & 3.1 70B \\
        \midrule
        Amazon & 1.00 & 1.00 & 0.58 & 1.00 \\
        Anyscale & 0.02 & --- & --- & --- \\
        Azure & 0.02 & 0.10 & 0.01 & 0.01 \\
        Deepinfra & 0.01 & 0.04 & 0.04 & 0.01 \\
        Fireworks & 0.00 & 1.00 & 0.03 & 0.00 \\
        Groq & 0.01 & 0.75 & 0.00 & 0.59 \\
        Perplexity & 1.00 & 1.00 & 1.00 & 1.00 \\
        Replicate & 0.02 & --- & 0.48 & --- \\
        Together & 0.01 & 0.09 & 0.00 & 0.00 \\
        nf4 & 1.00 & 1.00 & 1.00 & 1.00 \\
        int8 & 0.00 & 0.12 & 1.00 & 0.92 \\
        watermark & 0.12 & 0.45 & 0.50 & 0.06 \\
        \half & \textit{0.00} & \textit{0.00} & \textit{0.00} & \textit{0.00} \\
        \full & \textit{0.00} & \textit{0.06} & \textit{0.00} & \textit{0.00} \\
        \bottomrule
    \end{tabular}
    \caption{
        Power against alternatives for the full audit with the Bonferroni correction.
        The FPRs against the full- and half-precision nulls are italicized.
    }
    \label{tab:audit_bonferroni_power}
\end{table}

Table \ref{tab:audit_power} shows the overall audit results.
The 8B models are tested on three prompt distributions (Wikipedia, \humaneval, \ultrachat), while the 70B models are tested on Wikipedia.
Tests from different prompt distributions are combined using a Bonferroni correction in Table \ref{tab:audit_bonferroni_power}.

\paragraph{Discussion: the composite null reduces power.}
In \S\ref{sec:apis}, we use a composite null hypothesis that combines the \full and \half distributions.
Table \ref{tab:composite} shows the power of the Hamming MMD test in this composite null setting, stratified by model and prompt distribution.
In general, power is reduced using the composite null.
Power is generally highest on \humaneval, which collects longer completions than Wikipedia (250 vs. 50 tokens).

\begin{table}[h]
    \centering
    \small
    \begin{tabular}{lllllllll}
        \toprule
         & \multicolumn{4}{l}{Wikipedia} & \multicolumn{2}{l}{\humaneval} & \multicolumn{2}{l}{\ultrachat} \\
         & 3 8B & 3.1 8B & 3 70B & 3.1 70B & 3 8B & 3.1 8B & 3 8B & 3.1 8B \\
        \midrule
        watermark & 0.30 & 0.00 & 0.07 & 0.03 & 0.40 & 0.85 & 0.06 & 0.38 \\
        int8 & 0.07 & 0.00 & 1.00 & 0.96 & 0.01 & 0.45 & 0.01 & 0.03 \\
        nf4 & 0.09 & 1.00 & 1.00 & 1.00 & 1.00 & 1.00 & 1.00 & 1.00 \\
        fp16 & \textit{0.02} & \textit{0.01} & \textit{0.00} & \textit{0.01} & \textit{0.00} & \textit{0.02} & \textit{0.00} & \textit{0.04} \\
        \full & \textit{0.01} & \textit{0.00} & \textit{0.00} & \textit{0.00} & \textit{0.00} & \textit{0.10} & \textit{0.01} & \textit{0.04} \\
        \bottomrule
    \end{tabular}
    \caption{
        Power against local alternatives for the composite null setting (\S\ref{sec:apis}) in character space ($L=1000$), stratified by model and prompt distribution.
    }
    \label{tab:composite}
\end{table}

\paragraph{Discussion: the Hamming MMD is correlated with absolute differences in \humaneval average accuracy.}
In Figure \ref{fig:api_mmd_humaneval} in the main text, we show that the Hamming MMD is correlated with the \textit{absolute} differences in \humaneval average accuracy.
However, the direction of this difference is inconsistent across APIs: sometimes APIs with nonzero MMDs have higher \humaneval accuracy than the local model, and sometimes they have lower accuracy (see Table \ref{tab:humaneval_expanded}).
Our main argument in the main text is that regardless of direction, one can predict the \textit{magnitude} of change using the MMD. This is important in its own right; for example, it suggests that when the effect size returned by our test is large, users should do a manual examination for quality. 
For tasks with an automated verifier, like \humaneval, this two-step procedure may seem roundabout; however, for tasks without an automated accuracy metric, the MMD effect size provides users signal on when to invest manual resources into reviewing if the API is affecting performance.

\begin{table}[h]
    \centering
    \small
    \begin{tabular}{lcc}
        \toprule
        Version & Llama-3 8B Instruct & Llama-3.1 8B Instruct \\
        \midrule
        \full & 0.17 & 0.67 \\
        fp16 & 0.17 & 0.39 \\
        Amazon & 0.19 & 0.52 \\
        Azure & 0.19 & 0.72 \\
        Deepinfra & 0.18 & 0.64 \\
        Fireworks & 0.19 & 0.68 \\
        Groq & 0.14 & 0.26 \\
        int8 & 0.15 & 0.53 \\
        nf4 & 0.00 & 0.00 \\
        Perplexity & 0.00 & 0.06 \\
        Replicate & 0.22 & --- \\
        Together & 0.19 & 0.66 \\
        watermark & 0.08 & 0.55 \\
        \bottomrule
    \end{tabular}
    \caption{
        Raw \humaneval average accuracy for the Llama-3 8B Instruct and Llama-3.1 8B Instruct models.
    }
    \label{tab:humaneval_expanded}
\end{table}

\FloatBarrier
\subsection{Comparing APIs to each other}
\begin{figure}[h]
    \centering
    \includegraphics[width=\textwidth]{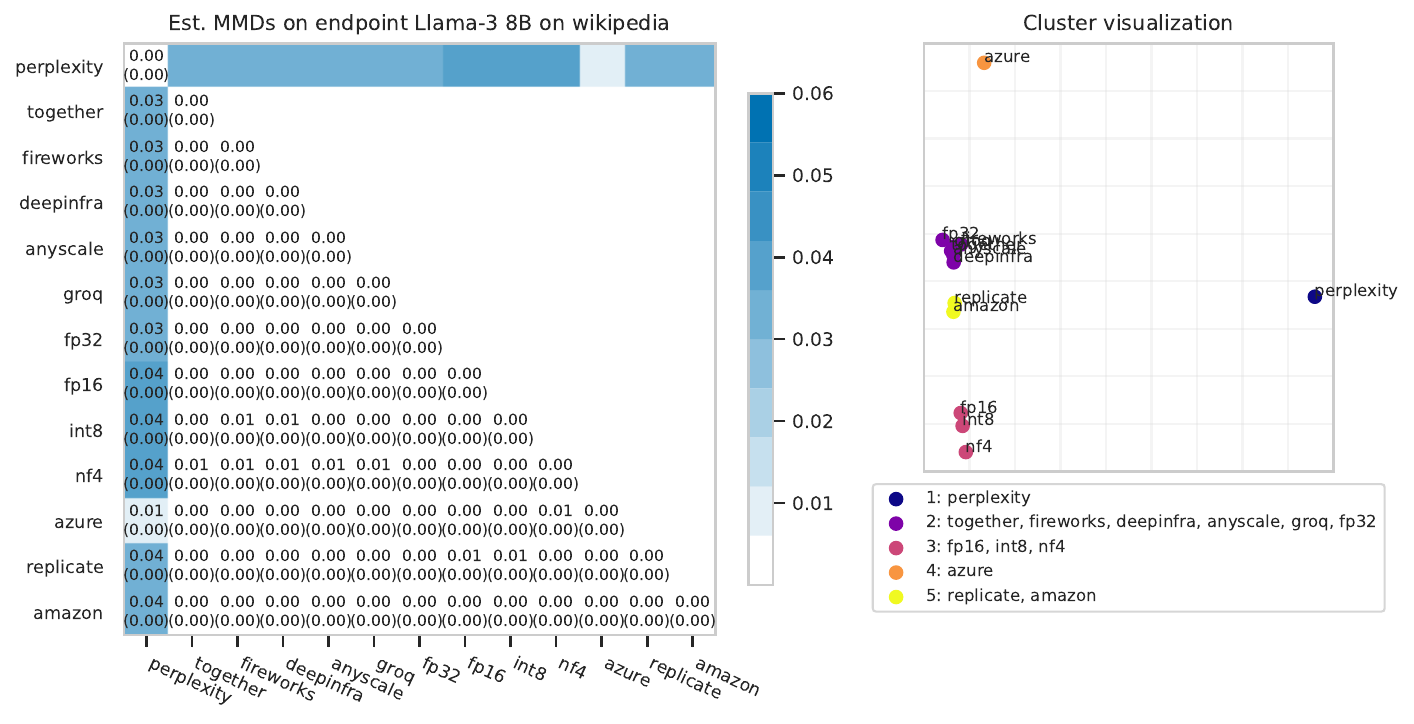}
    \caption{
        \textit{(Left)} Hamming MMDs between APIs for the Llama-3 8B model on the Wikipedia prompt distribution.
        \textit{(Right)} Visualization of the 2D spectral clustering components. Clusters are colored together.
    }
    \label{fig:api_api_Meta-Llama-3-8B-Instruct_wikipedia}
\end{figure}

\begin{figure}[h]
    \centering
    \includegraphics[width=\textwidth]{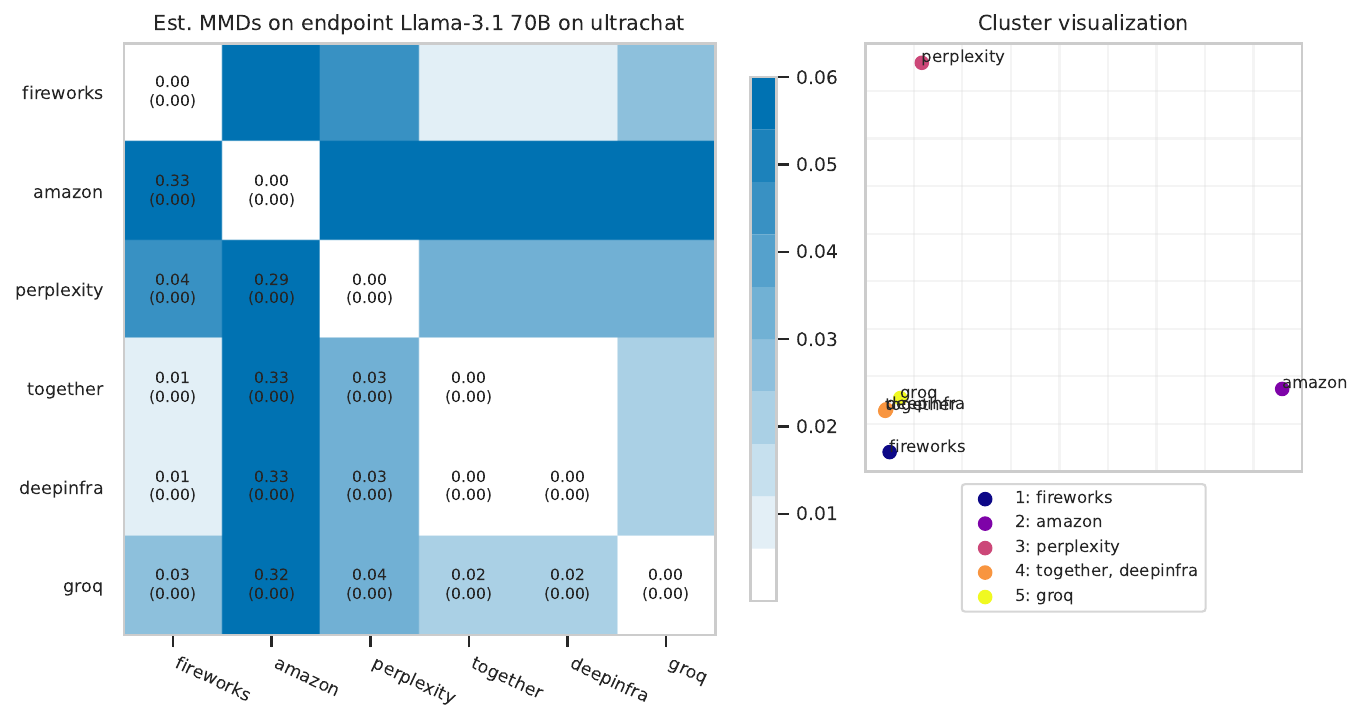}
    \caption{
        \textit{(Left)} Hamming MMDs between APIs for the Llama-3.1 70B model on the \ultrachat prompt distribution.
        \textit{(Right)} Visualization of the 2D spectral clustering components. Clusters are colored together.
    }
    \label{fig:api_api_Meta-Llama-3.1-70B-Instruct_ultrachat}
\end{figure}

In Figures \ref{fig:api_api_Meta-Llama-3-8B-Instruct_wikipedia} -- \ref{fig:api_api_Meta-Llama-3.1-70B-Instruct_ultrachat},
we compute the pairwise MMDs between APIs and quantized model weights for all prompt distributions (Wikipedia, \humaneval, \ultrachat) and available models (\llama-3 8B, \llama-3.1 8B, \llama-3 70B, \llama-3.1 70B).
We use spectral clustering with two components to discover groups of implementations.
Providers that pass the audit in Table \ref{tab:api_pass_fail} are typically clustered with the null distributions $\half$ and $\full$, reflecting that they are distributionally close to these nulls.

\begin{figure}[tb]
    \centering
    \includegraphics[width=\textwidth]{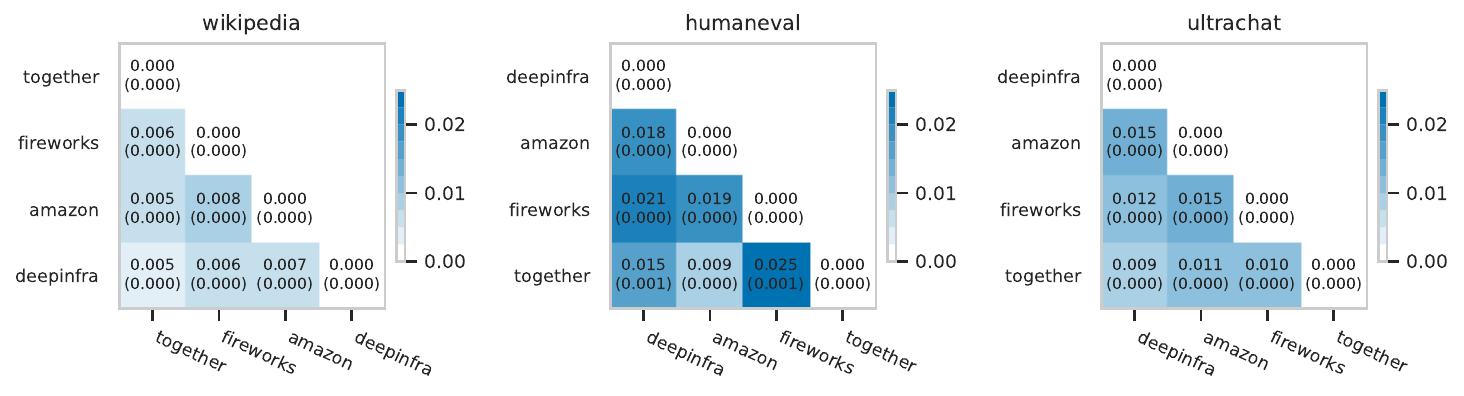}
    \caption{Estimated MMDs between \llama-3.1 405B APIs on three prompt distributions. Deepinfra publishes that they serve \llama-3.1 405B at fp8 precision. (Source: \href{https://deepinfra.com/meta-llama/Meta-Llama-3.1-405B-Instruct}{Deepinfra website})}
    \label{fig:api_api_405b}
\end{figure}

Additionally, Figure \ref{fig:api_api_405b} shows the estimated MMDs between APIs for each of the three prompt distributions for the \llama-3.1 405B model.
Due to their size, we could not sample from the released weights for this large model directly, but we can still estimate the distances between APIs for this model.


\begin{figure}[h]
    \centering
    \includegraphics[width=\textwidth]{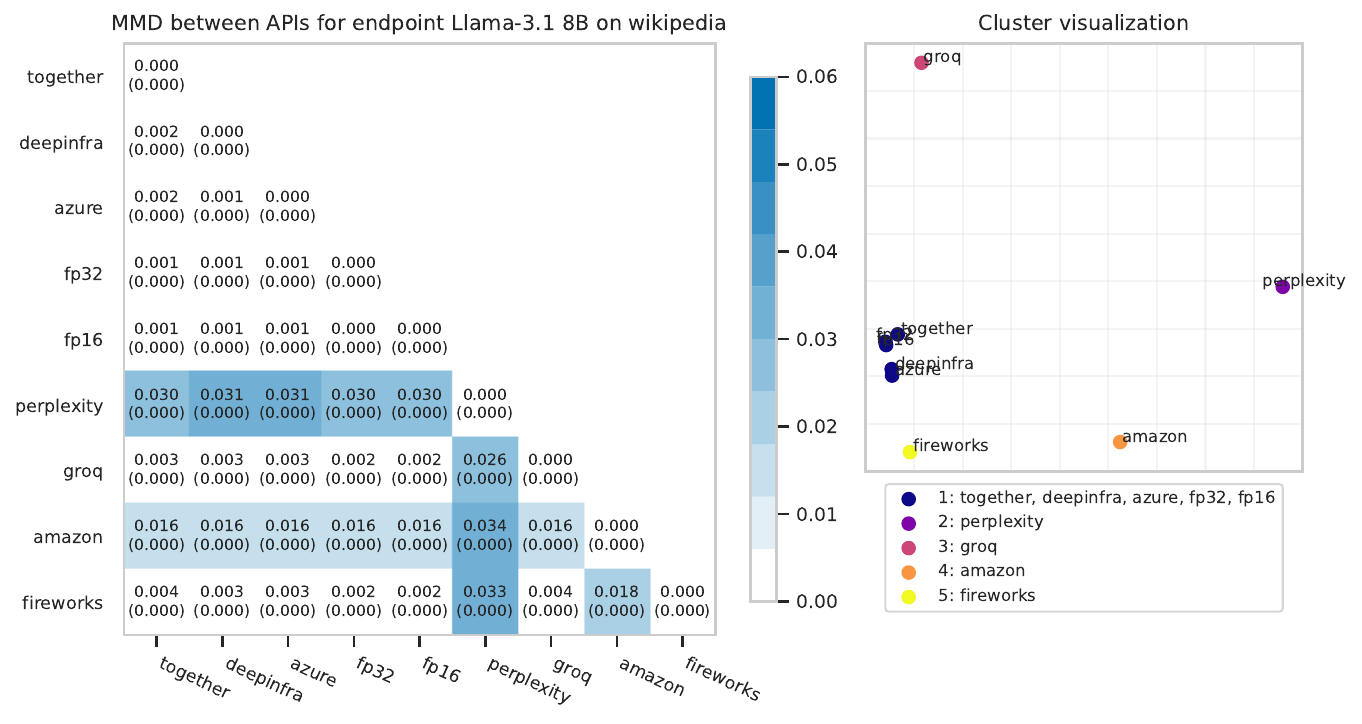}
    \caption{
        \textit{(Left)} Hamming MMDs between APIs for the Llama-3.1 8B model on the Wikipedia prompt distribution.
        \textit{(Right)} Visualization of the 2D spectral clustering components. Clusters are colored together.
    }
\end{figure}

\begin{figure}[h]
    \centering
    \includegraphics[width=\textwidth]{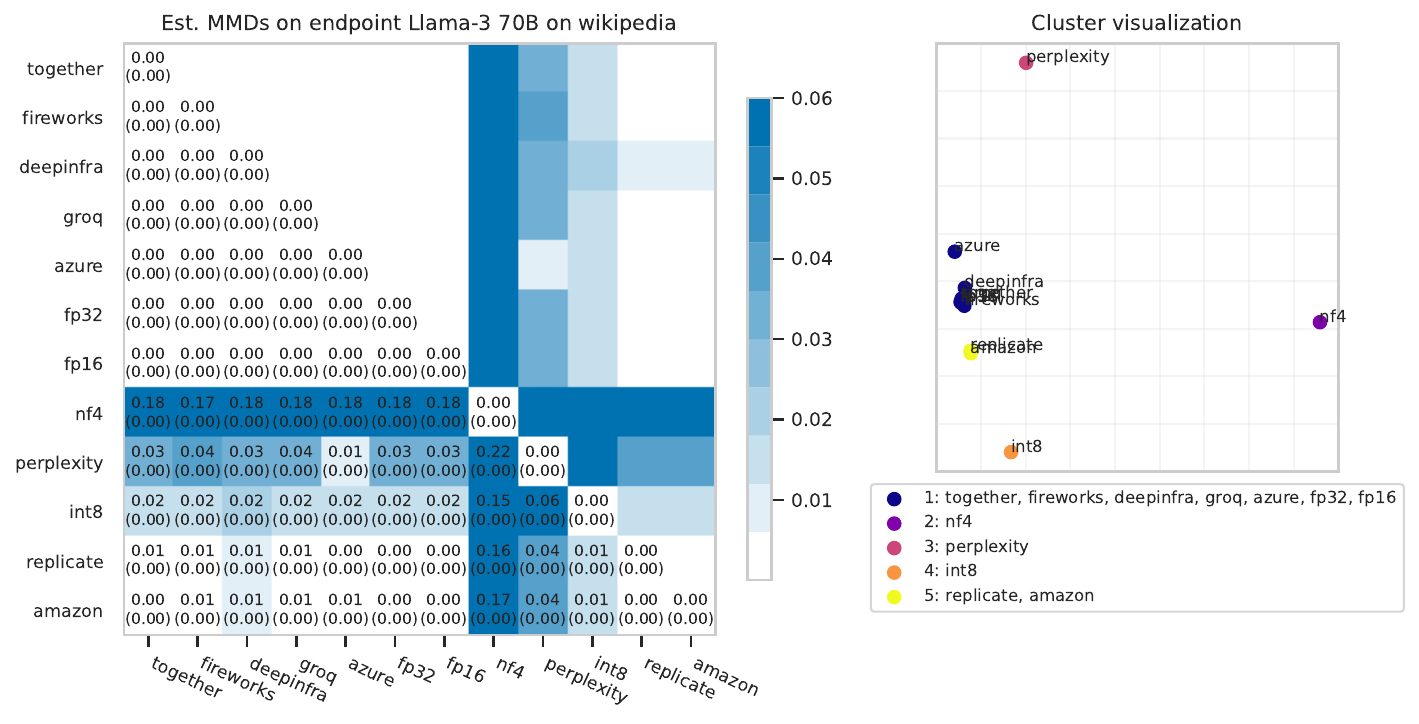}
    \caption{
        \textit{(Left)} Hamming MMDs between APIs for the Llama-3 70B model on the Wikipedia prompt distribution.
        \textit{(Right)} Visualization of the 2D spectral clustering components. Clusters are colored together.
    }
\end{figure}

\begin{figure}[h]
    \centering
    \includegraphics[width=\textwidth]{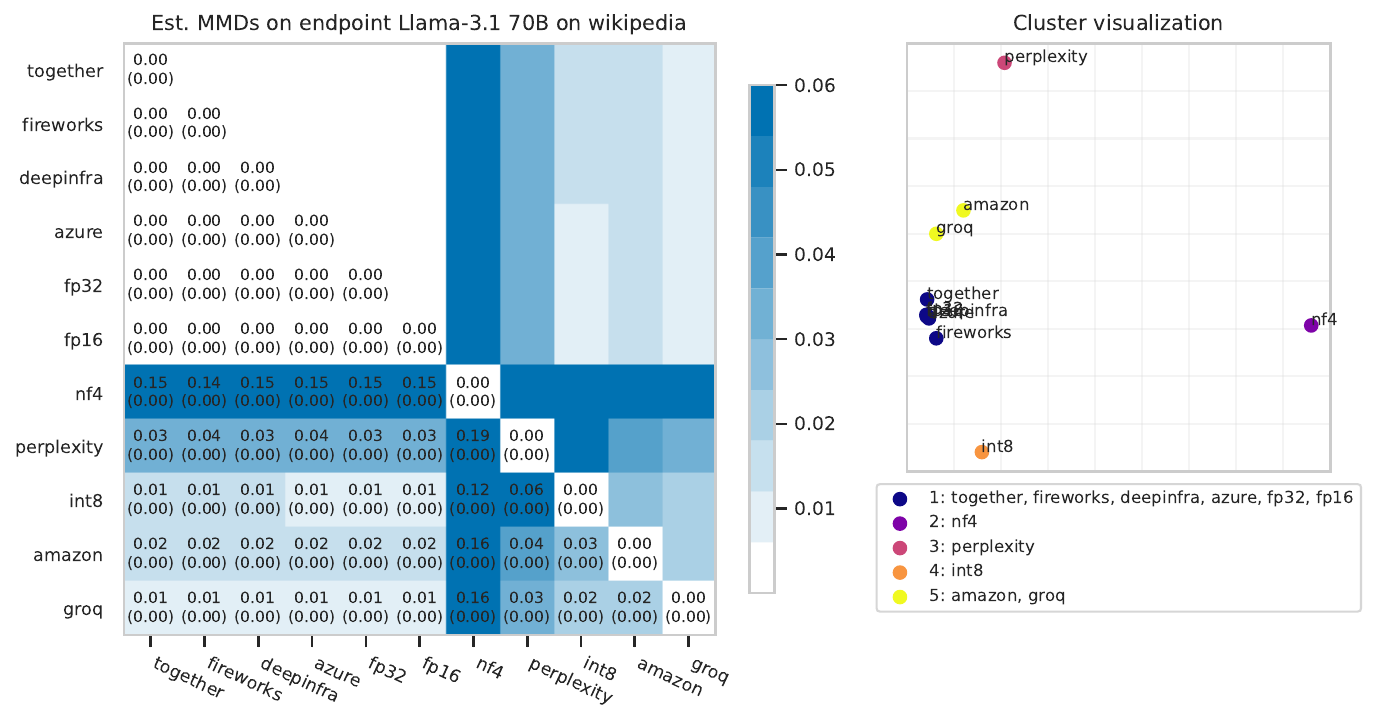}
    \caption{
        \textit{(Left)} Hamming MMDs between APIs for the Llama-3.1 70B model on the Wikipedia prompt distribution.
        \textit{(Right)} Visualization of the 2D spectral clustering components. Clusters are colored together.
    }
\end{figure}


\begin{figure}[h]
    \centering
    \includegraphics[width=\textwidth]{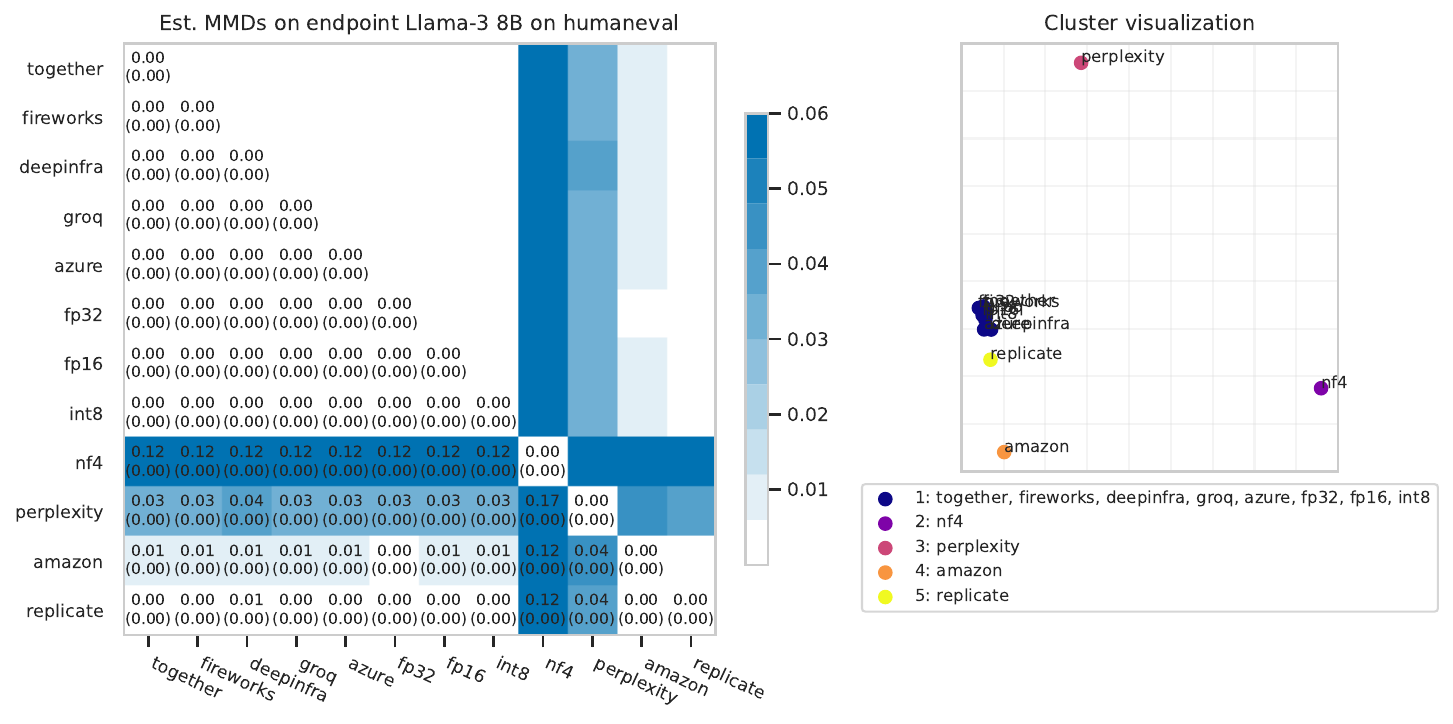}
    \caption{
        \textit{(Left)} Hamming MMDs between APIs for the Llama-3 8B model on the \humaneval prompt distribution.
        \textit{(Right)} Visualization of the 2D spectral clustering components. Clusters are colored together.
    }
\end{figure}

\begin{figure}[h]
    \centering
    \includegraphics[width=\textwidth]{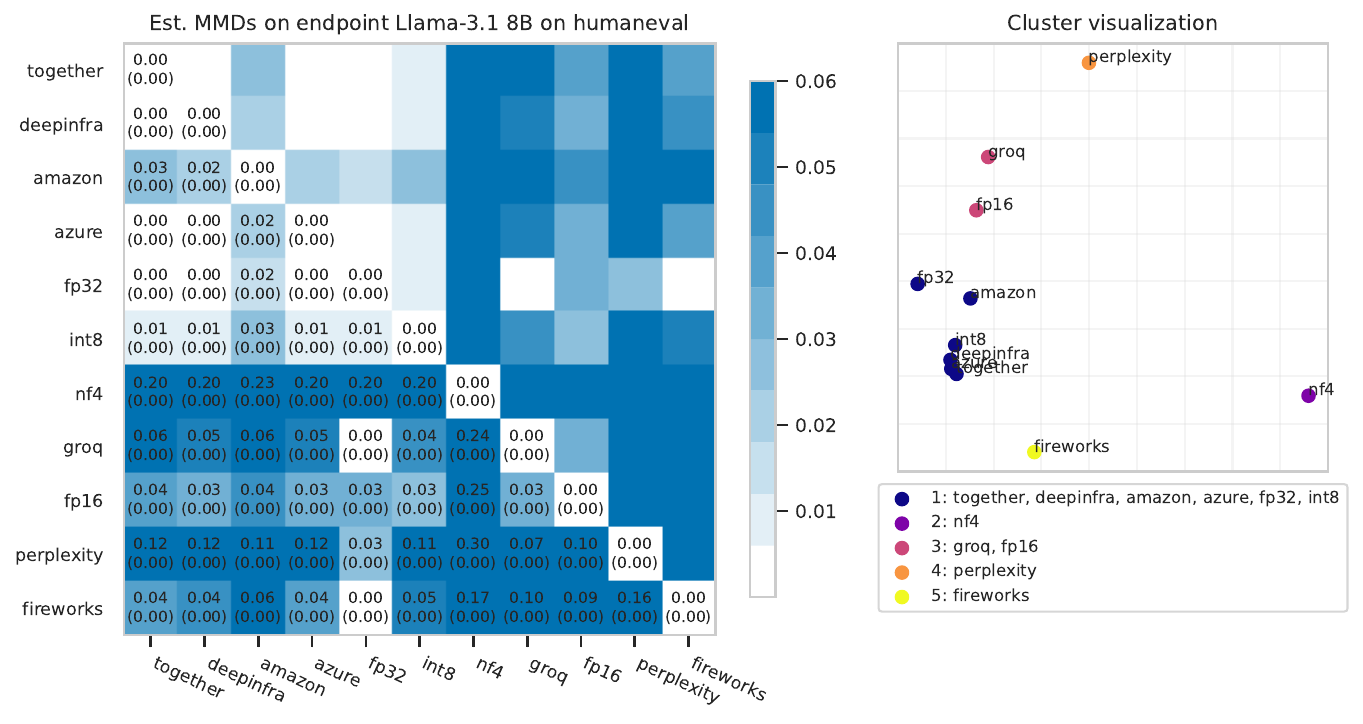}
    \caption{
        \textit{(Left)} Hamming MMDs between APIs for the Llama-3.1 8B model on the \humaneval prompt distribution.
        \textit{(Right)} Visualization of the 2D spectral clustering components. Clusters are colored together.
    }
\end{figure}

\begin{figure}[h]
    \centering
    \includegraphics[width=\textwidth]{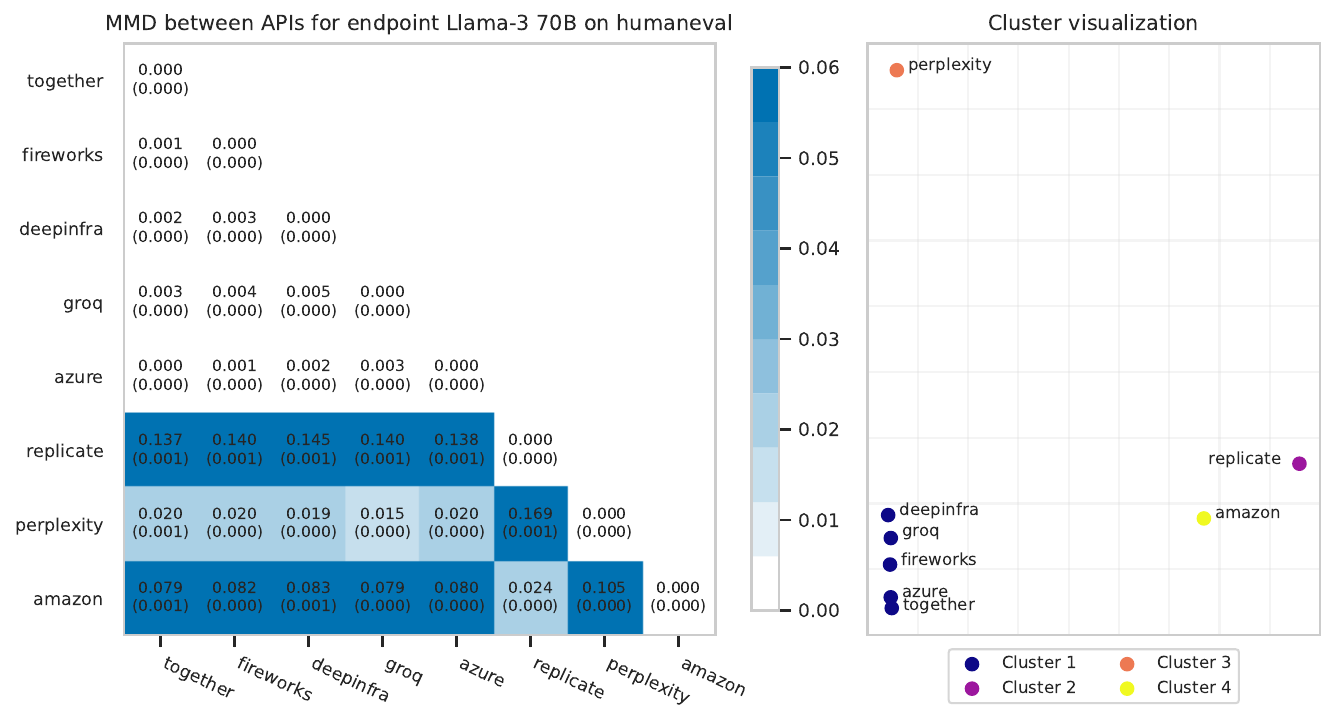}
    \caption{
        \textit{(Left)} Hamming MMDs between APIs for the Llama-3 70B model on the \humaneval prompt distribution.
        \textit{(Right)} Visualization of the 2D spectral clustering components. Clusters are colored together.
    }
\end{figure}

\begin{figure}[h]
    \centering
    \includegraphics[width=\textwidth]{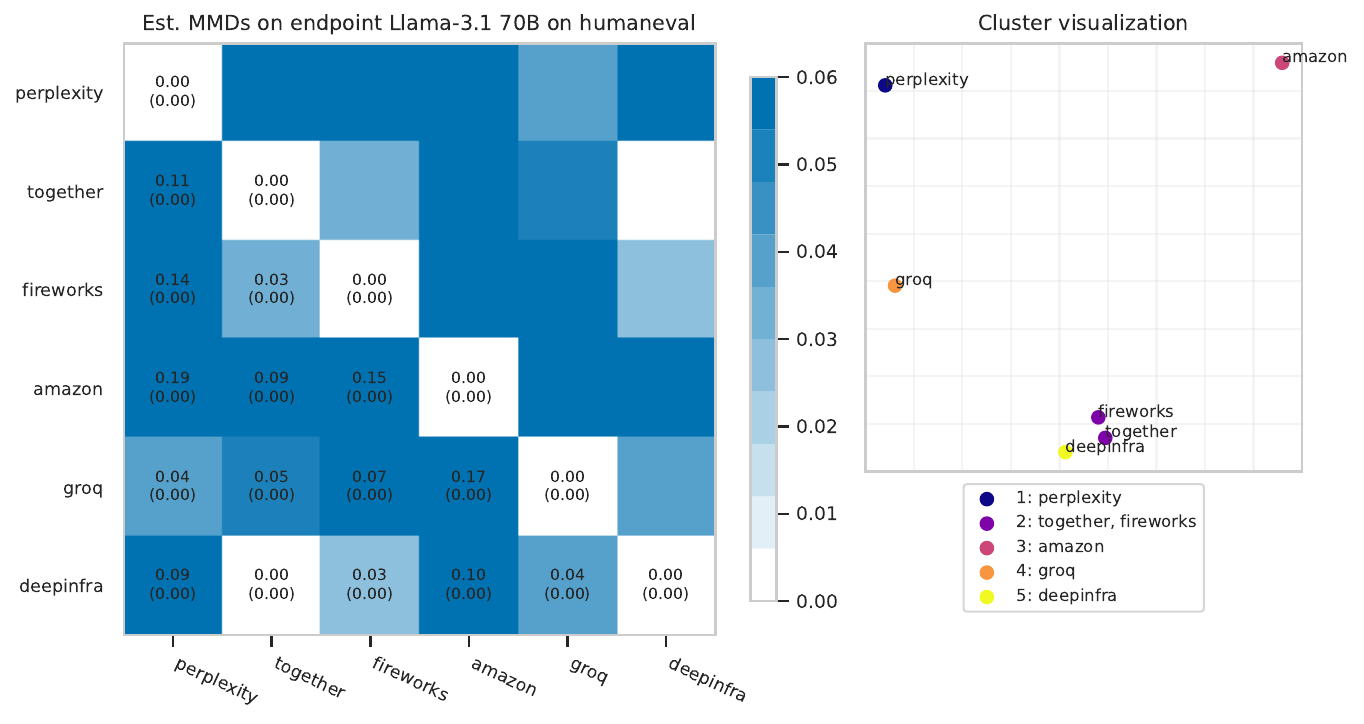}
    \caption{
        \textit{(Left)} Hamming MMDs between APIs for the Llama-3.1 70B model on the \humaneval prompt distribution.
        \textit{(Right)} Visualization of the 2D spectral clustering components. Clusters are colored together.
    }
\end{figure}

\begin{figure}[h]
    \centering
    \includegraphics[width=\textwidth]{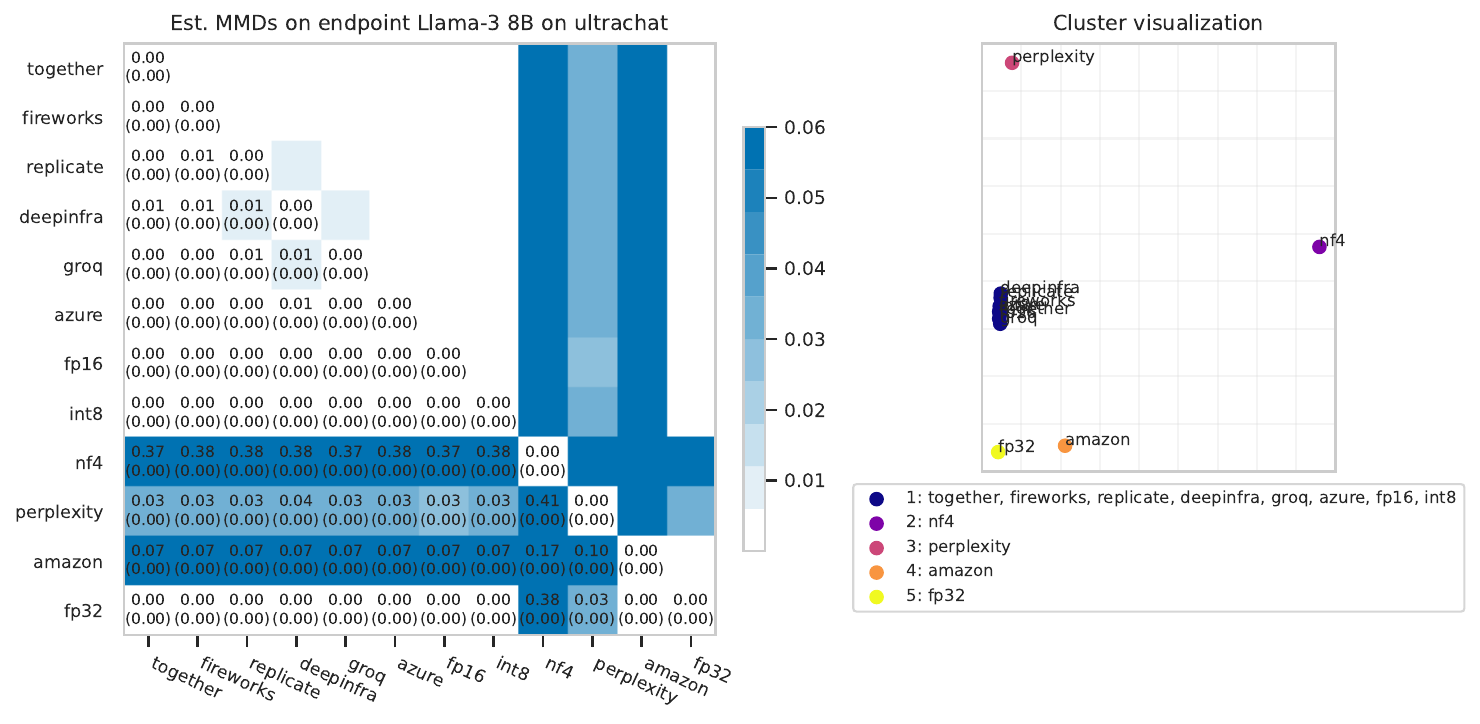}
    \caption{
        \textit{(Left)} Hamming MMDs between APIs for the Llama-3 8B model on the \ultrachat prompt distribution.
        \textit{(Right)} Visualization of the 2D spectral clustering components. Clusters are colored together.
    }
\end{figure}

\begin{figure}[h]
    \centering
    \includegraphics[width=\textwidth]{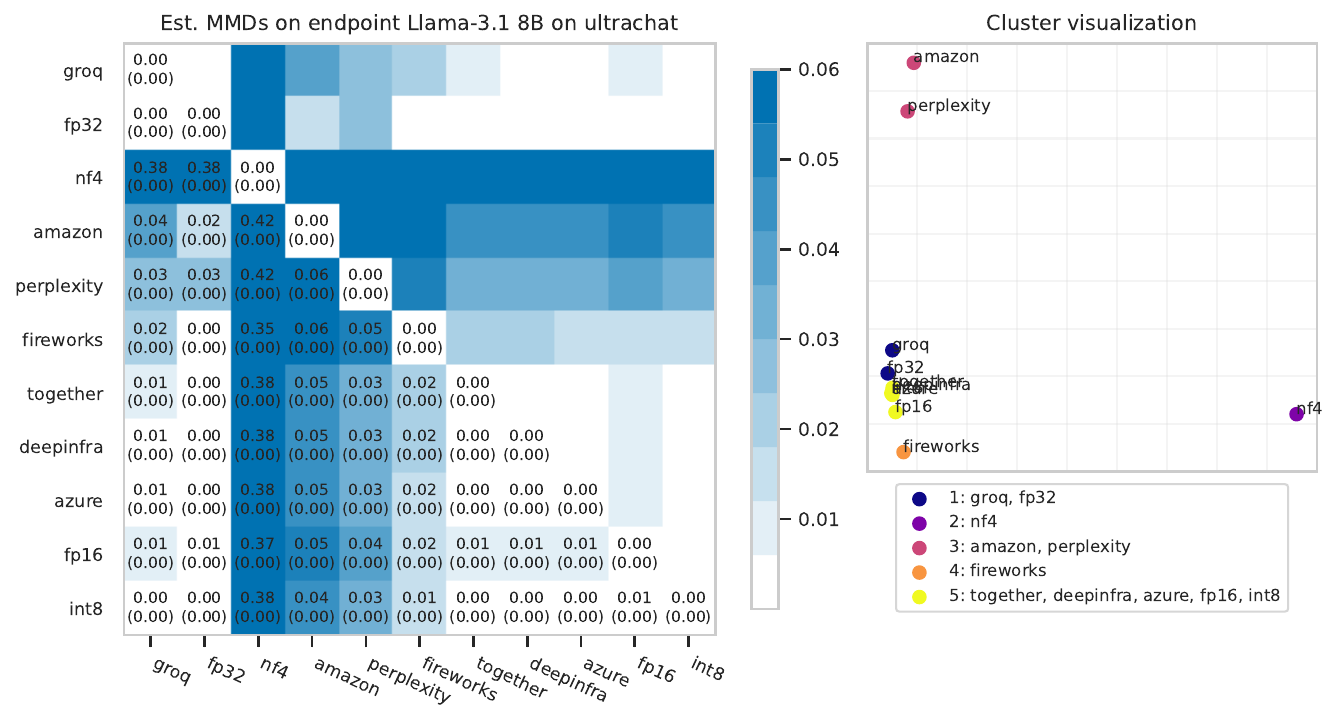}
    \caption{
        \textit{(Left)} Hamming MMDs between APIs for the Llama-3.1 8B model on the \ultrachat prompt distribution.
        \textit{(Right)} Visualization of the 2D spectral clustering components. Clusters are colored together.
    }
\end{figure}

\begin{figure}[h]
    \centering
    \includegraphics[width=\textwidth]{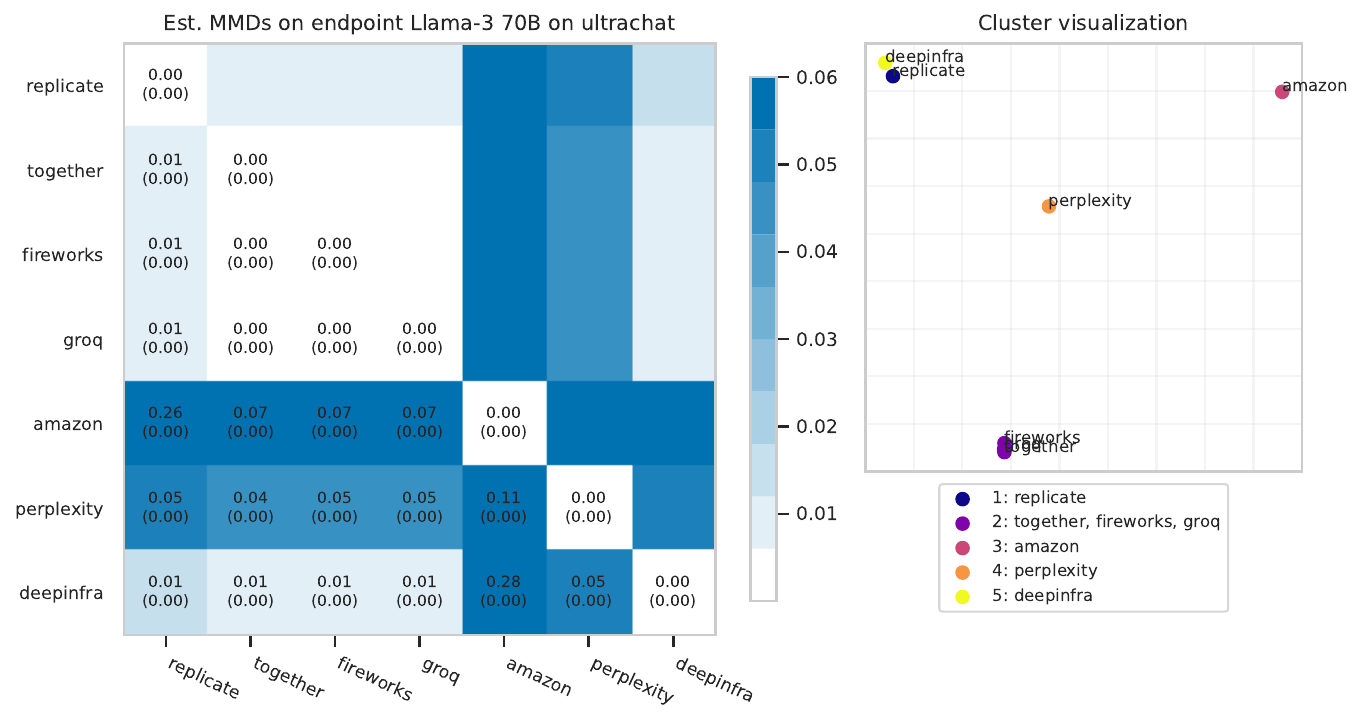}
    \caption{
        \textit{(Left)} Hamming MMDs between APIs for the Llama-3 70B model on the \ultrachat prompt distribution.
        \textit{(Right)} Visualization of the 2D spectral clustering components. Clusters are colored together.
    }
\end{figure}

\FloatBarrier
\section{Data samples}\label{app:samples}
In this appendix, we randomly sample completions from pairs of distributions to illustrate the differences between them.
In general, we observe that it is challenging to visually distinguish between samples from different distributions, especially for longform tasks.
This is because each distribution produces diverse outputs.
Formal statistical tests that we describe in the main text are necessary to detect these differences.

In a few cases, detected differences between distributions are also visually obvious.
For example, we observe that the \llama 70B-scale models quantize poorly, and their \nf completions are degenerate.
Different language models also often differ in how they begin completions.
We also observe that some APIs (in particular, Perplexity) seem to be producing lower-entropy completions than the reference distribution, suggesting
some form of caching or incorrect implementation of the temperature parameter.

\subsection{Comparing models to quantized or watermarked versions}\label{app:samples_models_vs_quantized}

\begin{samplebox}{Samples from \full vs. \half \llama-3 70B (Instruct)}
    \textbf{Prompt:} Continue the paragraph. Do not output anything except the continuation to the paragraph. Start the continuation immediately.
    ``Calectasia is a genus of about fifteen species of flowering plants in the family Dasypogonaceae and ...''\\

    {\small
    \def\arraystretch{1.5}
    \begin{tabularx}{\linewidth}{>{\raggedright\arraybackslash}X>{\raggedright\arraybackslash}X}
        \textit{\full} & \textit{\half} \\
        \arrayrulecolor[gray]{0.9}\hline
        ``...are native to eastern Australia and New Zealand, with one species, Calectasia articulata, also occurring in Lord Howe Island. The genus is thought to have evolved from a single ancestral species in the mid-Tertiary period. Calectasia		 & ``...is endemic to Australia, with a majority of the species found in the southern and western parts of the country. The plants in this genus are shrubs or small trees with simple or compound leaves, and they produce small yellow or white flowers in clusters
        \\
        \arrayrulecolor[gray]{0.9}\hline
        ``...is native to the Pacific region, with species found in Australia, New Zealand, and the Pacific Islands, including New Caledonia and the Solomon Islands. The genus is characterised by its unique inflorescences, which are racemes	 & ``...are native to Australia and New Caledonia, which are characterized by their club-shaped inflorescences and solitary, axillary flowers. The flowers are typically yellow or mauve in color and are often highly fragrant, attracting a range

        \\
        \arrayrulecolor[gray]{0.9}\hline
        ``...are native to the coastal regions of southeastern Australia, southern New South Wales, Victoria, Tasmania, and the coastal islands of Southern and Southwestern Australia, typically preferring well-drained sandy soils and full sun to partial shade."	 & ``... is endemic to northern and eastern Australia, including Tasmania and southern New Guinea. They are spinifex grasses of shrub-like habit, typically ranging in size from 0.5 to 4 metres in height. The leaves are strap-shaped

        \\
        \arrayrulecolor[gray]{0.9}\hline
        ``...is endemic to Australia and New Guinea, with most species being restricted to the southwestern corner of Western Australia. The plants are typically small to medium in size, and have leaves that are often reduced in size or absent, producing stems and inflores	 &``...is endemic to Australia, with species found in the coastal and mountainous regions of eastern New South Wales, the Australian Capital Territory, and southeastern Queensland. The plants are characterised by their simple, solitary, terminal flowers, which appear in late winter
        \\
        \arrayrulecolor[gray]{0.9}\hline
        ``...is found in Australia and New Zealand, where they grow in coastal areas with sandy soils. They are also known as calectases or lomandras in some regions, and are commonly encountered in dry, sandy areas and scrublands. The	 & ``... native to eastern Australia and New Caledonia. They are shrub to small tree species, often growing up to 2-6 metres in height, with simple serrated leaves that are usually 10-40 mm long and a thick cop
        \\
        \arrayrulecolor[gray]{0.9}\hline
    \end{tabularx}
    }
\end{samplebox}

\begin{samplebox}{Samples from \full vs. \half \llama-3 70B (Instruct)}
    \textbf{Prompt:} Continue the paragraph. Do not output anything except the continuation to the paragraph. Start the continuation immediately.\\
    ``The agreement provides for training teams from the Libyan Stock Exchange in Tripoli and in London to...''\\

    {\small
    \def\arraystretch{1.5}
    \begin{tabularx}{\linewidth}{>{\raggedright\arraybackslash}X>{\raggedright\arraybackslash}X}
        \textit{\full} & \textit{\half} \\
        \arrayrulecolor[gray]{0.9}\hline
        ``...enhance their skills in various aspects of securities trading and exchange management, including market regulation, risk management, and investor protection. The training program is designed to equip the Libyan Stock Exchange with the necessary expertise to operate efficiently and effectively, promoting economic growth		 & ``...enable them to develop their technical skills and knowledge in areas such as market regulation, surveillance, and settlement systems. Additionally, the partnership will facilitate the exchange of best practices and expertise between the two exchanges, with the aim of enhancing the Libyan capital market
        \\
        \arrayrulecolor[gray]{0.9}\hline
        ``...enhance their skills in areas such as market regulation, trading systems, and investor relations, with the goal of increasing the efficiency and transparency of the Libyan capital market. Additionally, the London Stock Exchange will provide technical assistance to help the Libyan Stock Exchange	 & ``...gain expertise in modern trading practices, risk management, and regulatory compliance, with the aim of enhancing the Libyan capital market's infrastructure and attracting foreign investment to support the country's economic growth and development. The partnership is also expected to facilitate the listing of
        \\
        \arrayrulecolor[gray]{0.9}\hline
        ``...enhance their skills and knowledge in modern stock exchange operations, with a focus on trading, settlement, and market surveillance systems. Additionally, the London Stock Exchange will provide technical assistance to help the Libyan Stock Exchange develop its regulatory framework and operational infrastructure,	& ``...to enhance their skills in trading, regulation, and market operations, with a focus on implementing international best practices and standards. The training programs will cover a range of topics, including trading rules, risk management, and corporate governance, and will be designed
        \\
        \arrayrulecolor[gray]{0.9}\hline
        ``exchange knowledge and best practices in areas such as market operations, listing requirements, and investor relations, with the goal of strengthening the Libyan Stock Exchange's capacity to attract foreign investment and promote economic growth in Libya.''	&``...to enhance their skills and knowledge in areas such as market regulation, trading systems, and risk management. Additionally, the deal will facilitate the exchange of information and expertise between the two countries, with the LSE providing guidance on developing a more efficient and
        \\
        \arrayrulecolor[gray]{0.9}\hline
        ``...to enhance their skills in areas such as market regulation, securities trading, and risk management, with the ultimate goal of strengthening the Libyan capital market and promoting economic development in the region.''	 & ``...exchange information and best practices on securities regulation, market surveillance, and trading operations. The training will focus on enhancing the capacity of the Libyan Stock Exchange to effectively regulate and monitor the Libyan capital market, as well as to improve its operational efficiency and competitiveness
        \\
        \arrayrulecolor[gray]{0.9}\hline
    \end{tabularx}
    }
\end{samplebox}

\begin{samplebox}{Samples from \full vs. \inteight \llama-3 70B (Instruct)}
    \textbf{Prompt:} Continue the paragraph. Do not output anything except the continuation to the paragraph. Start the continuation immediately.\\
    ``Weaver, Tom. I Talked with a Zombie: Interviews with 23 Veterans of Horror and Sci-Fi Films and Tele...''\\

    {\small
    \def\arraystretch{1.5}
    \begin{tabularx}{\linewidth}{>{\raggedright\arraybackslash}X>{\raggedright\arraybackslash}X}
        \textit{\full} & \textit{\inteight} \\
        \arrayrulecolor[gray]{0.9}\hline
        ``...vision, from the 1930s to the 1980s. Jefferson, NC: McFarland, 2009. Print. This book provides a unique perspective on the history of horror and science fiction, as it is told through	 & ``...including Max Brooks,ed Troma Nightmares and Brian De Palma's Mission: Impossible III,ed Stan Winston,ed Edward Scissorhands,ed...''
        \\
        \arrayrulecolor[gray]{0.9}\hline
        ``...vision. McFarland, 2011. This book provides a unique perspective on the horror and science fiction genres by offering firsthand accounts from individuals who played a significant role in their development. The interviews, which span several decades, offer insight into	 & ``...including legendary actors like Bruce Campbell,...''
        \\
        \arrayrulecolor[gray]{0.9}\hline
        ``...vision About Their Most Memorable Roles. McFarland, 2011. Print. This book offers a unique perspective on the horror and sci-fi genre, as it is comprised of interviews with actors and actresses who have played iconic roles in various	 & ``...Famous Monsters,ed by Tom Weaver,ed features in-depth interviews with ed ed legends of horror and sci-fi,ed films and television,ed. Amonged the ed ed is Bruce Dern,ed,ed horror
        \\
        \arrayrulecolor[gray]{0.9}\hline
        ``...vision. McFarland, 2011. This collection of interviews with professionals from the horror and science fiction genres offers valuable insights into the production of classic films and television shows. The book includes discussions with actors, writers, directors, and special	 &``...and TV,ed with the cast and crew of George Romero's Diary of the Dead,ing to recall the good old days of horror movie making. We caught up with Tippi Hedren,...''
        \\
        \arrayrulecolor[gray]{0.9}\hline
        ``...vision, McFarland, 2008. This book provides a unique insight into the experiences of actors, directors, and special effects artists who worked on classic horror and science fiction films and television shows. The interviews offer a behind-the-scenes look	 &``...Famous Monsters and Others.''
        \\
        \arrayrulecolor[gray]{0.9}\hline
    \end{tabularx}
    }
\end{samplebox}

\begin{samplebox}{Samples from \full vs. \nf \mistral 7B (Instruct)}
    \textbf{Prompt:} Continue the paragraph. Do not output anything except the continuation to the paragraph. Start the continuation immediately.\\
    ``The agreement provides for training teams from the Libyan Stock Exchange in Tripoli and in London to...''\\

    {\small
    \def\arraystretch{1.5}
    \begin{tabularx}{\linewidth}{>{\raggedright\arraybackslash}X>{\raggedright\arraybackslash}X}
        \textit{\full} & \textit{\nf} \\
        \arrayrulecolor[gray]{0.9}\hline
        ``collaborate on developing market infrastructure, regulatory frameworks, and trading systems. Additionally, the partnership allows for knowledge exchange on best practices in financial services, risk management, and market surveillance. This cooperation aims to foster a stable, thriving,	 & ``ensure the seamless exchange of knowledge and best practices in trading, market surveillance, and regulatory compliance. This transatlantic collaboration aims to prepare the Libyan Stock Exchange for a more robust and resilient financial market, integrating international standards
        \\
        \arrayrulecolor[gray]{0.9}\hline
        ``work collaboratively on developing and strengthening the market infrastructure, adhering to international standards, and facilitating the growth of the Libyan capital market.''	 &``ensure compliance with international best practices and regulations, as well as enhancing capacity-building in various areas such as trading systems, surveillance, and market data management.''
        \\
        \arrayrulecolor[gray]{0.9}\hline
        facilitate the exchange and sharing of knowledge in financial markets and regulations, while also ensuring the proper implementation of international best practices and standards. This collaborative effort aims to strengthen the Libyan Stock Exchange, enhance its efficacy, and promote sustainable economic growth		 & ``ensure that they are equipped with the latest financial market trading techniques and best practices. This exchange of knowledge is aimed at modernizing the Libyan Stock Exchange and making it more competitive on an international level. The teams will undergo intensive training programs
        \\
        \arrayrulecolor[gray]{0.9}\hline
        ``facilitate the development of modern trading practices and administrative systems, ensuring a harmonious integration with global financial standards.''	&``to share best practices and enhance the technical competencies of the teams, ensuring the smooth operation and sustained growth of the exchange.''
        \\
        \arrayrulecolor[gray]{0.9}\hline
        ...further enhance their capabilities and gain in-depth knowledge of modern trading practices, regulatory frameworks, and technology implementations. This collaboration aims to strengthen the Libyan Stock Exchange's position in the international financial market, contribute to economic stability,	 &``share their expertise in various areas, including market analysis, trading practices, regulatory compliance, and technology infrastructure. This exchange of knowledge aims to equip the Libyan Stock Exchange with the necessary tools and skills to effectively manage their market, improve its transparency
        \\
        \arrayrulecolor[gray]{0.9}\hline
    \end{tabularx}
    }
\end{samplebox}

\begin{samplebox}{Samples from \full vs. \nf \llama-3 70B (Instruct)}
    \textbf{Prompt:} Continue the paragraph. Do not output anything except the continuation to the paragraph. Start the continuation immediately.\\
    ``The agreement provides for training teams from the Libyan Stock Exchange in Tripoli and in London to...''\\

    {\small
    \def\arraystretch{1.5}
    \begin{tabularx}{\linewidth}{>{\raggedright\arraybackslash}X>{\raggedright\arraybackslash}X}
        \textit{\full} & \textit{\nf} \\
        \arrayrulecolor[gray]{0.9}\hline
        ``...enhance their skills in areas such as market regulation, surveillance, and settlement systems, with the aim of improving the overall efficiency and transparency of the Libyan capital market. Additionally, the agreement outlines plans for the development of a new trading platform, which	 & Course!!!!!!!!!!!!!!!!!!!!!!!!!!!!!!!!!!!!!!!!!!!!!!!!!
        \\
        \arrayrulecolor[gray]{0.9}\hline
        ``familiarize themselves with the operational systems and regulatory frameworks of the London Stock Exchange, with the aim of enhancing the capacity and competitiveness of the Libyan capital market. Additionally, the partnership will facilitate the exchange of best practices and expertise in areas such as risk		 & Earth!!!!!!!!!!!!!!!!!!!!!!!!!!!!!!!!!!!!!!!!!!!!!!!!!
        \\
        \arrayrulecolor[gray]{0.9}\hline
        ``...enhance their knowledge and skills in areas such as market regulation, trading operations, and risk management, thus enabling them to develop a more efficient and transparent capital market in Libya. The cooperation is expected to facilitate the integration of the Libyan Stock Exchange into	 &  certain!!!!!!!!!!!!!!!!!!!!!!!!!!!!!!!!!!!!!!!!!!!!!!!!!
        \\
        \arrayrulecolor[gray]{0.9}\hline
        ``...gain expertise in modern stock exchange operations, including trading, settlement, and regulatory practices. This training will be conducted by experienced staff from the London Stock Exchange and will take place over a period of six months. Additionally, the agreement includes the provision of		& healthy!!!!!!!!!!!!!!!!!!!!!!!!!!!!!!!!!!!!!!!!!!!!!!!!!
        \\
        \arrayrulecolor[gray]{0.9}\hline
        ``...design and implement a new trading system, which will enable the Libyan Stock Exchange to operate more efficiently and in line with international standards. The training will cover all aspects of the new system, including its technical capabilities, operational procedures, and risk management strategies	 & Reverse!!!!!!!!!!!!!!!!!!!!!!!!!!!!!!!!!!!!!!!!!!!!!!!!!
        \\
        \arrayrulecolor[gray]{0.9}\hline
    \end{tabularx}
    }
\end{samplebox}

\begin{samplebox}{Samples from \full vs. watermarked \llama-3.1 8B (Instruct)}
    \textbf{Prompt:} Continue the paragraph. Do not output anything except the continuation to the paragraph. Start the continuation immediately.\\
    ``The agreement provides for training teams from the Libyan Stock Exchange in Tripoli and in London to...''\\

    {\small
    \def\arraystretch{1.5}
    \begin{tabularx}{\linewidth}{>{\raggedright\arraybackslash}X>{\raggedright\arraybackslash}X}
        \textit{\full} & \textit{Watermark} \\
        \arrayrulecolor[gray]{0.9}\hline
        ``...further develop and refine their skills in the areas of corporate governance, risk management, and financial markets operations, with the goal of strengthening the Libyan Stock Exchange's capacity to effectively play a central role in the country's economic development.''	 & ``...enable them to develop the needed expertise to list Libyan companies on both exchanges, with the ultimate goal of increasing access to international finance for these companies and promoting economic growth in Libya.''
        \\
        \arrayrulecolor[gray]{0.9}\hline
        ``...learn about international best practices, listing rules, and compliance regulations, as well as to discuss ways to enhance the Libyan Stock Exchange's capacity and increase investor confidence.''	 & ``...enable the implementation of international best practices and improve market operations, assessment, and enforcement. The project also aims to develop the institutional capacity of the Libyan Stock Exchange to list and regulate Sharia-compliant securities, and to establish a robust framework for corporate governance
        \\
        \arrayrulecolor[gray]{0.9}\hline
        ``...enhance their capacity and knowledge in areas such as corporate governance, risk management, and trading system implementation, ultimately contributing to the development of a more robust and efficient Libyan capital market.''	 & ``learn best practices for listing and disclosure standards, as well as the use of the London Stock Exchange's ELITE programme to support the growth of Libyan companies.''
        \\
        \arrayrulecolor[gray]{0.9}\hline
        ``...familiarize them with international best practices in exchange operations, market making, and listing requirements. The teams will also undergo negotiation and stakeholder management training to enhance their capability to engage effectively with investors and issuers. Additionally, the agreement includes a	 &``...familiarise them with the best practices of international exchange organisations, and for regular monitoring of Libya's progress towards regulatory compliance with international standards.''
        \\
        \arrayrulecolor[gray]{0.9}\hline
        ``...facilitate greater integration of Libyan listed companies into the international capital market and to enhance the exchange's technical capabilities. The London Stock Exchange's internationally recognized standards and rules will be shared with the Libyan Stock Exchange to improve the efficiency and transparency of trading	 &``...optimize trading and listings systems, and also allocated a sum for implementing a state-of-the-art trading platform at the Libyan Stock Exchange in Tripoli.''
        \\
        \arrayrulecolor[gray]{0.9}\hline
    \end{tabularx}
    }
\end{samplebox}

\subsection{Comparing models to other models}\label{app:samples_models_vs_models}
\begin{samplebox}{Samples from \llama-3 8B (Instruct) vs. Phi-3 Mini (Instruct)}
    \textbf{Prompt:} Continue the paragraph. Do not output anything except the continuation to the paragraph. Start the continuation immediately.\\
    ``Gung Ho! (full title: Gung Ho!: The Story of Carlson\'s Makin Island Raiders) is a 1943 American war ...''\\

    {\small
    \def\arraystretch{1.5}
    \begin{tabularx}{\linewidth}{>{\raggedright\arraybackslash}X>{\raggedright\arraybackslash}X}
        \textit{\llama-3 8B} & \textit{Phi-3 Mini} \\
        \arrayrulecolor[gray]{0.9}\hline
        ``...drama film directed by Ray Enright, starring Randolph Scott and James Cagney. The story is based on the true events of Carlson's Makahiki Raiders, a group of U.S. Marine Raiders who, under the command of Lt	 & war film directed by Robert Wise and produced by Pandro S. Berman for Samuel Goldwyn Productions. It is based on the true story of the Makin Island raid during World War II. The film tells the tale of U
        \\
        \arrayrulecolor[gray]{0.9}\hline
        ``...drama film directed by Robert Wise, produced by Samuel G. Engel, and starring Randolph Scott. The movie is based on the true story of Alamo Scout Company B, 2nd Battalion, 1st Marine Regiment, during the Pacific	 & ...comedy-drama film directed by Mel Ferrer. Set during the Second Battle of the Philippines, the story is based on the true events of World War II and follows the 21st Marine Raider Battalion, led by Captain Howard
        \\
        \arrayrulecolor[gray]{0.9}\hline
        ...``comedy-drama film directed by Sam Wood, starring Randolph Scott, James Gleason, and Robert Ryan. The film is based on the true story of the United States Marine Corps' 2nd Marine Raider Battalion, who were tasked with the		& war biographical film directed by Robert Lewis and John Ford. Set during World War II, the film focuses on the exploits of Lieutenant Commander Evans Carlson and his specialized American task force in Japan, highlighting the innovative guerrilla
        \\
        \arrayrulecolor[gray]{0.9}\hline
        ...``film directed by Ray Enright and starring Randolph Scott, Alan Hale, Jr., and Noah Beery Jr. The film is based on a true story about a group of United States Marine Raiders who launched a successful raid on the Japanese occupied Makin		&war film directed by John Sturges, written by Ward Elliott, and adapted from the book ``Gung Ho!'' by E.R. Hewett. This film chronicles the daring and harrowing experiences of the Makin
        \\
        \arrayrulecolor[gray]{0.9}\hline
        ``...film directed by Ray Enright and starring Randolph Scott, Barbara Britton, and Frank Mitchell. The movie is based on the true story of the formation of the Makin Island Attack Unit, a group of United States Marine Raiders who were tasked with		 & ...decion film directed by Robert Hendrickson and produced by William Alland. The film stars Victor Mature, Betty Field, and Sabu in the lead roles. Set during World War II, it chronicles the story of a group of
        \\
        \arrayrulecolor[gray]{0.9}\hline
    \end{tabularx}
    }
\end{samplebox}

\begin{samplebox}{Samples from \llama-3.1 70B (Instruct) vs. Gemma-2 9B (Instruct)}
    \textbf{Prompt:} Continue the paragraph. Do not output anything except the continuation to the paragraph. Start the continuation immediately.\\
    ``Gung Ho! (full title: Gung Ho!: The Story of Carlson\'s Makin Island Raiders) is a 1943 American war ...''\\

    {\small
    \def\arraystretch{1.5}
    \begin{tabularx}{\linewidth}{>{\raggedright\arraybackslash}X>{\raggedright\arraybackslash}X}
        \textit{\llama-3.1 70B} & \textit{Gemma-2 9B} \\
        \arrayrulecolor[gray]{0.9}\hline
        ``... film directed by Ray Enright and starring Randolph Scott. The movie is based on the views attributed to the commander of the 2nd Raider Battalion, Evans Carlson, and the Battle of Makin Island (also known as the Battle of Butar	&  film directed by Ray Enright and written by  .'''''' 
        \\
        \arrayrulecolor[gray]{0.9}\hline
        ``... film directed by Ray Enright and starring Randolph Scott. The film is largely fictional, but is based on the true story of the Carlson's Raiders, the 2nd Marine Raider Battalion, which was activated on June 19, 1942	&film, directed by Ray Enright and starring  Wallace Ford.  The film is based on a 1942 book by T.R. Fehrenbach about the famous amphibious raid of Makin Island in 1942
        \\
        \arrayrulecolor[gray]{0.9}\hline
        ``... film directed by Ray Enright, starring Randolph Scott, and based on the true story of the Carlson's Raiders, a special military unit set up by President Franklin D. Roosevelt as an American counterpart to the British Commandos, that took part in	&film about the real-life exploits of United States Marine Corps scout-snipers on Makin Island during World War II.  The film was directed by Ray Enright and stars Jon Hall as Lt. Alexander Bonnyman, a stern Marine officer
        \\
        \arrayrulecolor[gray]{0.9}\hline
        ``... film directed by Ray Enright that follows the story of Lieutenant Colonel Evans Carlson's 2nd Marine Raider Battalion and their 30-day occupation of Makin Island in the Gilbert Islands during August 1942. The film stars Randolph Scott, Robert	&film, directed by Henry H. Arbitell, narrated by Philip J. Freeman, with music by William O. Steward and appeared in the following excerpt from the Encarta Encyclopedia:``Gung Ho! (full title: Gung
        \\
        \arrayrulecolor[gray]{0.9}\hline
        film directed by Ray Enright, starring Randolph Scott, and based on the true story of the Carlson's Raiders and the Makin Island raid. The screenplay was written by W. S. Burnett and data collected from the USMC. The film was	 & film based on the real-life United States Marine Corps raid on Makin Island during World War II.  
        \\
        \arrayrulecolor[gray]{0.9}\hline
    \end{tabularx}
    }
\end{samplebox}

\begin{samplebox}{Samples from \llama-3 8B (Instruct) vs. \llama-3.1 70B (Instruct)}
    \textbf{Prompt:} Continue the paragraph. Do not output anything except the continuation to the paragraph. Start the continuation immediately.\\
    ``Gung Ho! (full title: Gung Ho!: The Story of Carlson\'s Makin Island Raiders) is a 1943 American war ...''\\

    {\small
    \def\arraystretch{1.5}
    \begin{tabularx}{\linewidth}{>{\raggedright\arraybackslash}X>{\raggedright\arraybackslash}X}
        \textit{\llama-3 8B} & \textit{\llama-3.1 70B} \\
        \arrayrulecolor[gray]{0.9}\hline
        ...film directed by Robert Wise and starring Randolph Scott. The movie is based on the true story of the Carlson's Patrol, also known as the Makin Island Raiders, a small group of U.S. Marines who conducted a daring raid on Japanese-held	 & ``... film directed by Ray Enright, starring Randolph Scott. The film's plot is based on the real-life World War II Makin Island raid led by Lieutenant Colonel Evans Carlson's 2nd Marine Raider Battalion. Scott plays a fictional commander of the

        \\
        \arrayrulecolor[gray]{0.9}\hline
        ``...drama film directed by Ray Enright, starring Alan Ladd as John W. ``Moe'' Smith, a Marine raider who takes part in a special operations mission on Wake Island during the early years of World War II. The film	 & film directed by Ray Enright, starring Randolph Scott, and produced by Walter Wanger. The screenplay was written by W. S. [Walker] Crowther and Joseph Hoffman, corrected to Hoffman's dictation, based on the book of the same

        \\
        \arrayrulecolor[gray]{0.9}\hline
        ``...drama film directed by Ray Enright and starring Randolph Scott, James Wong Howe, and Robert Ryan. The film is based on the true story of the U.S. Marine Raider Battalion, led by Major Evans Carlson, who captured the strategic Mak	& film directed by Ray Enright, starring Randolph Scott and Alan Curtis. The film is based on the story of the Carlson's Raiders and their famous raid on Makin Island during World War II. It attempts to convey the message that battle is not just

        \\
        \arrayrulecolor[gray]{0.9}\hline
        ...film directed by Ray Enright and starring Alan Ladd, Robert Preston, and Veronica Lake. The movie is based on a true story about a group of U.S. Marines stationed on Makin Island, also known as Butaritari,	 & film directed by Ray Enright and starring Randolph Scott. The film is based on the true story of the Carlson's Patrol, a 30-day raid on Japanese-held Makin Island between August 17 and September 25, 1942, by

        \\
        \arrayrulecolor[gray]{0.9}\hline
        ``...drama film directed by Ray Enright and starring Randolph Scott, Alan Curtis, and James Ellison. The movie is based on the true story of the Carlson's Raiders, a group of United States Marine Raiders who seized the strategic Makin Island from	&   film directed by Ray Enright and starring Randolph Scott, based on the true story of the 2nd Marine Raider Battalion's World War II raid on Makin Island, which was led by Lieutenant Colonel Evans Carlson.
        \\
        \arrayrulecolor[gray]{0.9}\hline
    \end{tabularx}
    }
\end{samplebox}

\begin{samplebox}{Samples from \llama-3.1 8B (Instruct) vs. OLMo 7B (Instruct)}
    \textbf{Prompt:} Continue the paragraph. Do not output anything except the continuation to the paragraph. Start the continuation immediately.\\
    ``Gung Ho! (full title: Gung Ho!: The Story of Carlson\'s Makin Island Raiders) is a 1943 American war ...''\\

    {\small
    \def\arraystretch{1.5}
    \begin{tabularx}{\linewidth}{>{\raggedright\arraybackslash}X>{\raggedright\arraybackslash}X}
        \textit{\llama-3.1 8B} & \textit{OLMo 7B} \\
        \arrayrulecolor[gray]{0.9}\hline
        ``... film directed by Ray Enright. The film is based on a true story about the 2nd Marine Raider Battalion, under the command of Lieutenant Colonel Evans F. Carlson, who occupied the Makin Island (now known as Butaritari	&  films, follows the inspiring true story of Colonel Carl L. Smith's 4,000-man Corps of famous Japanese-American Battle Ground Warriors, who triumphantly embarked on a 50-day, 90-mile Hawaiian desert race to build a trenches and
        \\
        \arrayrulecolor[gray]{0.9}\hline
        ...film directed by Ray Enright, starring Robert Hutton and Preston Foster. The movie is based on the book of the same name, If You Want to Look for Treasure by Maj. Carl G. Carlson, USMCR, which tells the	& drama film directed by Ray DeFinis and featuring an ensemble cast that includes William “Dub” Jones, Harold Gobber, Tom Tully, Alan Cotton, and Philip Merivale. Set during World War II, the film follows the
        \\
        \arrayrulecolor[gray]{0.9}\hline
        ... film directed by Ray Enright, starring Randolph Scott, Jameson Thomas and Carole Landis. The film is an adaptation of the 1943 non-fiction book of the same name by Robert P. Landry and Frank R. McCoy,	& mercenaries, mostly African-American, who are tasked with building and managing a new humangeous village for the local Malayan>'s (the indigenous people) and teaching them modern agriculture and forensics. observe the orderly and disciplined
        \\
        \arrayrulecolor[gray]{0.9}\hline
        ...film directed by Ray Enright, starring Wallace Beery and Carl Esmond. The screenplay was written by Jo Swerling and Joseph Hoffman, based on the 1943 book ``Gung Ho!'' by Robert P. Henderson.		&film, directed by Henry H. Arbitell, narrated by Philip J. Freeman, with music by William O. Steward and appeared in the following excerpt from the Encarta Encyclopedia: ``Gung Ho! (full title: Gung
        \\
        \arrayrulecolor[gray]{0.9}\hline
        ``...buddy film directed by Ray Enright, starring Robert Hutton as Major Jim Curry and Randolph Scott as Major Carl Entschede. The film is loosely based on a true story of Carlson's Makin Island Raiders, a United States Marine Corps	 & animation by Tatsuo Stella, with uneorologisticatley accurate meteorological animations by Chica Tokiji, the film celebrates the heroic deeds of a devoted team of outcasts led by frontiersman Sam Dwier Brown, whose words
        \\
        \arrayrulecolor[gray]{0.9}\hline
    \end{tabularx}
    }
\end{samplebox}

\subsection{Comparing APIs}\label{app:samples_models_vs_apis}

\begin{samplebox}{Samples from \full vs. Perplexity for \llama-3 8B (Instruct)}
    \textbf{Prompt:} Continue the paragraph. Do not output anything except the continuation to the paragraph. Start the continuation immediately.
    ``Calectasia is a genus of about fifteen species of flowering plants in the family Dasypogonaceae and ...''\\

    {\small
    \def\arraystretch{1.5}
    \begin{tabularx}{\linewidth}{>{\raggedright\arraybackslash}X>{\raggedright\arraybackslash}X}
        \textit{\full} & \textit{Perplexity} \\
        \arrayrulecolor[gray]{0.9}\hline
        ``...are native to the southern and eastern regions of Australia, New Zealand, and surrounding islands. The genus is characterized by its unique flowers, which are often yellow or orange in color and have five or six petals. Calectasia plants are often small	 & ``...is native to Australia, where they are commonly found in dry, open forests and woodlands. The plants are shrubs or small trees, typically growing to a height of 1-5 meters, with a spreading habit and a rounded crown.
        \\
        \arrayrulecolor[gray]{0.9}\hline
        ``...is endemic to Australia, with most species found in the states of Western Australia and Northern Territory. The plants are generally woody shrubs or small trees that grow up to 10 meters tall, with leaves that are typically divided into three lobes		 & ``...are native to Australia, where they are commonly found in dry, open forests and woodlands. The plants are shrubs or small trees that typically grow to a height of 1-5 meters, with a spread of around 1-3
        \\
        \arrayrulecolor[gray]{0.9}\hline
        ``...is primarily native to Australia, with some species also found in New Guinea and the Pacific Islands. The plants are generally shrubs or small trees, typically growing up to 2-5 meters tall, with simple or lobed leaves that are often		 & ``...is native to Australia, where they are found in a variety of habitats, including forests, woodlands, and grasslands. The plants are characterized by their distinctive, often brightly colored flowers, which are typically arranged in clusters or spikes. The leaves
        \\
        \arrayrulecolor[gray]{0.9}\hline
        ``...are native to the temperate and subtropical regions of Australia. They are small to medium-sized shrubs or trees, typically growing up to 5 meters tall, with a distinctive, layered or shrubby habit. The leaves are simple, alternate	 &``...are native to Australia, where they are commonly found in dry, open forests and woodlands. The plants are shrubs or small trees, typically growing to a height of 1-5 meters, and have a distinctive, sprawling habit. The

        \\
        \arrayrulecolor[gray]{0.9}\hline
        ``...is native to Australia and New Guinea. The plants are shrubs or small trees that typically grow to a height of around 1-5 meters, with a spread of up to 2-3 meters. The leaves are simple, alternate,	 &``...are native to Australia, where they are found in a variety of habitats, including forests, woodlands, and grasslands. The plants are characterized by their distinctive, often brightly colored flowers, which are typically arranged in clusters or spikes, and their

        \\
        \arrayrulecolor[gray]{0.9}\hline
    \end{tabularx}
    }
\end{samplebox}

\begin{samplebox}{Samples from \full vs. Amazon Bedrock for \llama-3 70B (Instruct)}
    \textbf{Prompt:} Continue the paragraph. Do not output anything except the continuation to the paragraph. Start the continuation immediately.\\
    ``Weaver, Tom. I Talked with a Zombie: Interviews with 23 Veterans of Horror and Sci-Fi Films and Tele...''\\

    {\small
    \def\arraystretch{1.5}
    \begin{tabularx}{\linewidth}{>{\raggedright\arraybackslash}X>{\raggedright\arraybackslash}X}
        \textit{\full} & \textit{Amazon Bedrock} \\
        \arrayrulecolor[gray]{0.9}\hline
        ``...visions. Jefferson, NC: McFarland, 2004. Weldon, Michael J. The Psychotronic Video Guide. New York: St. Martin's Press, 1996. Weldon, Michael J. The Psychot	 & ``...vision. McFarland \& Co, 2004.  This unique collection brings together a group of people who have been behind the scenes of some the most beloved and enduring horror and science fiction films and TV series, providing insight into how their
        \\
        \arrayrulecolor[gray]{0.9}\hline
        ...vision. Jefferson, NC: McFarland \& Company, Inc., 2004. ISBN 0-786-48181-0; Weaver, Tom. Interviews with B Science Fiction and Horror Movie Makers: Writers, Producers,	 & ``...vision. McFarland, 2004. ISBN 0-7864-1684-5. Weaver, Tom. They Fought in the Creature Features: A History of Makeup Special Effects Artists. McFarland, 2010

        \\
        \arrayrulecolor[gray]{0.9}\hline
        ``...vision. Jefferson, NC: McFarland, 2004. ISBN 0-7864-1686-1. Willemen, Paul. ``The Fantastic.'' In Film Comment 14, no. 6 (1978):	 & ``...Films and Television Shows. McFarland \& Company, 2009'' The author delVES into conversations with behind-the-scenes veterans, revealing lesser-known stories from beloved franchises such as Star Trek and Friday the 13th, providing insight into
        \\
        \arrayrulecolor[gray]{0.9}\hline
        ``...visions. Jefferson, NC: McFarland \& Co., 2011, p112-114, ISBN 0786444923; Weaver, Tom. Monsters, Mutants and Heavenly Creatures: Confessions of 14 Classic Sci-Fi/H	 & ``...vision. McFarland, 2004. ISBN 0-7864-1684-5. Weaver, Tom. They Fought in the Creature Features: A History of Makeup Special Effects Artists. McFarland, 2010

        \\
        \arrayrulecolor[gray]{0.9}\hline
        ``...vision. Applying an informal, conversational style to put his subjects at ease, Weaver elicits candid, often humorous, and occasionally revelatory discussions about the highs and lows of life in the genre movie trenches. Among the emotive chat-weaving	 &``...vision. New York: McFarland, 2004. Wells, Paul (ed.). Neo Noir. Short Films by Contemporary Film-makers. UK: Wallflower/Columbia University Press, 1999.''
        \\
        \arrayrulecolor[gray]{0.9}\hline
    \end{tabularx}
    }
\end{samplebox}

\begin{samplebox}{Samples from \full vs. Deepinfra for \llama-3.1 70B (Instruct)}
    \textbf{Prompt:} Continue the paragraph. Do not output anything except the continuation to the paragraph. Start the continuation immediately.
    ``Litzman served on the Knesset's Internal Affairs Committee from 1999 to 2001, and as the Deputy Chai...''
    
    {\small
    \def\arraystretch{1.5}
    \begin{tabularx}{\linewidth}{>{\raggedright\arraybackslash}X>{\raggedright\arraybackslash}X}
        \textit{\full} & \textit{Deepinfra} \\
        \arrayrulecolor[gray]{0.9}\hline
        ``...ordinator of the Second Netanyahu government from 1999 to 2001. He was appointed Minister of Housing and Construction by Ariel Sharon in 2001, a position he held until 2002, and has served as the Deputy Minister of Health		 & ``...Committee chairman. With the formation of Ariel Sharon's coalition government in 2001, Litzman was appointed Minister of Health, a position within which he continues to be until June 2011.''

        \\
        \arrayrulecolor[gray]{0.9}\hline
        ``...rman of the Knesset from 2009 to 2013. In 2001, he was appointed as the Minister of Health in Ariel Sharon's unity government, a position he held until Ehud Olmert became the leader		 &``per of the Knesset from 1999 to 2001. In 2001, he was appointed to the Minister of Religious Services by Ariel Sharon. Litzman served as Minister of Labor and Welfare, and Deputy Minister of Defense

        \\
        \arrayrulecolor[gray]{0.9}\hline
        ``ra in the same committee from 2001 to 2004, serving under the chairmanship of Gideon Ezra. In 2003, Litzman was appointed the head of the committee's subcommittee on emergencies. His electoral district covered	& ``...Chairman of the Knesset's Health Committee from 1999 to 2005. In 2001, he was appointed by Prime Minister Ariel Sharon as the Ministry of Health's chief rabbi, a post he holds to this day

        \\
        \arrayrulecolor[gray]{0.9}\hline
        ``rman of the Knesset from 2001 to 2008. In 2008, Litzman became Deputy Health Minister of Israel, a position he held until February 2019 when he resigned over allegations that he had obstruct	 &``...rector of the Ministry of Health from 2001 to 2006. During his tenure, he was responsible for implementing the Ministry's plans to improve the health services in Israel, particularly in the area of hospital management, health insurance reform,

        \\
        \arrayrulecolor[gray]{0.9}\hline
        ``...Chairman of the committee. He was also a member of the Knesset's Economics Committee, the Committee on the Status of Women and the Committee on the Rights of the Child.''		 &``...rman of the Committee on the Status of Women from 2001 to 2004. In 2003, he was appointed to serve on the Special Committee on the Rights of Women and the Family. Litzman also served as a
        \\
        \arrayrulecolor[gray]{0.9}\hline
    \end{tabularx}
    }
\end{samplebox}

\begin{samplebox}{Samples from \full vs. Amazon Bedrock for \llama-3 70B (Instruct)}
    \textbf{Prompt:} Continue the paragraph. Do not output anything except the continuation to the paragraph. Start the continuation immediately.
    ``He formally resigned as health minister on 26 November 2017, in protest of railroad repair work happ...''
    
    {\small
    \def\arraystretch{1.5}
    \begin{tabularx}{\linewidth}{>{\raggedright\arraybackslash}X>{\raggedright\arraybackslash}X}
        \textit{\full} & \textit{Amazon Bedrock} \\
        \arrayrulecolor[gray]{0.9}\hline
        ``ening on the railway line near his hometown of Likoni, which he believed was a threat to the ferry service that many locals relied on for transportation. His resignation was seen as a bold move, as he had previously been a close ally of the & ``ening on a major bridge in the town of Temerín, which was causing significant disruptions to comm...''

        \\
        \arrayrulecolor[gray]{0.9}\hline
        ``ening on a major rail line without adequate alternative transportation arrangements being made, which he felt would disproportionately affect the poor and vulnerable members of society. His decision sparked widespread criticism of the government's handling of the situation, and many saw it as a significant & ening a major downtown road, which was scheduled to coincide with a large international gathering of health experts in the city, and would likely cause significant traffic disruptions, all while the city's main hospital was still recovering from a recent devastating flood.
        \\
        \arrayrulecolor[gray]{0.9}\hline
        ``...ening simultaneously on multiple sections of track, which he believed would cause unnecessary disruptions to commuters. Shortly after, he announced his candidacy for the opposition leader position, citing his desire to bring about change to the government's handling of infrastructure development projects. His	& ening on a section of tracks that run close to the hospital where he worked as a doctor, which he claimed posed a health risk to patients and staff.
        \\
        \arrayrulecolor[gray]{0.9}\hline
        ``ening on a major railway line, which he contended would lead to widespread disruptions and negatively impact the country's already strained healthcare system, a stance widely seen as a bold move by many in the medical community, who had long been calling for greater	 & ``ening on a key rail line without consulting him, which he believed was a vital healthcare transportation route for patients and medical personnel.''

        \\
        \arrayrulecolor[gray]{0.9}\hline
        ``ening on a major railway line without proper planning and communication, which caused disruptions to healthcare services and put patients' lives at risk. This decision sparked widespread debate and criticism from various stakeholders, including medical professionals, politicians, and citizens, who saw it	 & ...ening on a Saturday, which he said was an "attack on the Jewish Sabbath" and would cause ``massive disruptions'' to the country's transportation system.
        \\
        \arrayrulecolor[gray]{0.9}\hline
    \end{tabularx}
    }
\end{samplebox}

\begin{samplebox}{Samples from \full vs. Together for \llama-3 70B (Instruct)}
    \textbf{Prompt:} Continue the paragraph. Do not output anything except the continuation to the paragraph. Start the continuation immediately.
    ``He formally resigned as health minister on 26 November 2017, in protest of railroad repair work happ...''
    
    {\small
    \def\arraystretch{1.5}
    \begin{tabularx}{\linewidth}{>{\raggedright\arraybackslash}X>{\raggedright\arraybackslash}X}
        \textit{\full} & \textit{Together} \\
        \arrayrulecolor[gray]{0.9}\hline
        ening on the main railway line that connects Nairobi to the port city of Mombasa, which he claimed was being done without proper planning and would cause disruptions to the transportation of medical supplies, and potentially harm the healthcare system. His resignation was seen as	 & ening on a track that ran close to his home in the upscale Mumbai suburb of Bandra, which he claimed was causing noise pollution and disturbing the peace in his neighborhood.
        \\
        \arrayrulecolor[gray]{0.9}\hline
        ``ening on a nearby site, which he claimed would cause noise pollution and disrupt the operations of a children's hospital that was being built on adjacent land, a project he had championed during his tenure as minister. The controversy surrounding the railroad repair work	& ``ening on the same day as the national memorial service for the victims of the Hwaseong capsizing disaster, which he felt was a grave mistake and showed a lack of respect for the deceased and their families.''
        \\
        \arrayrulecolor[gray]{0.9}\hline
        ``ening on a rail line that runs near the Prime Minister's residence, which he believed was unfairly prioritized over more urgent infrastructure projects in other parts of the country. His decision sparked widespread outrage and accusations of government corruption, leading to a series of		& ``ening on a key railway line linking the country's major urban centers, which he claimed was a risky and unnecessary project that would burden the taxpayers and was pushed through without adequate discussion or transparency. His resignation was seen as a major blow to the government
        \\
        \arrayrulecolor[gray]{0.9}\hline
        ``ening on a major rail line that connects Moscow to the city of Kazan, which he believed would be detrimental to the already precarious healthcare system in the region. His resignation was seen as a bold move, as he was one of the most prominent	 & ``ening on a main commuter line during rush hour, which he claimed would cause undue disruptions to the daily lives of citizens. His decision was met with a mix of surprise and admiration from the public, with many praising his willingness to take a stand against
        \\
        \arrayrulecolor[gray]{0.9}\hline
        ``ening on a main highway in his constituency, which he claimed was not properly coordinated with local authorities and caused unnecessary disruptions to the daily lives of his constituents.''		 &``...ening on a rail line near his hometown, which he believed would cause environmental damage and disrupt the local ecosystem. His resignation was met with widespread support from environmental groups and local residents, who saw him as a champion of their cause. Despite his departure
        \\
        \arrayrulecolor[gray]{0.9}\hline
    \end{tabularx}
    }
\end{samplebox}

\begin{samplebox}{Samples from \full vs. Fireworks for \llama-3.1 80B (Instruct)}
    \textbf{Prompt:} Complete the code. Do not output anything except the completion. Start the continuation immediately.
\begin{verbatimcode}
```
from typing import List
def intersperse(numbers: List[int], delimeter: int) -> List[int]:
    """ Insert a number 'delimeter' between every two consecutive elements of input list `numbers'
    >>> intersperse([], 4)
    []
    >>> intersperse([1, 2, 3], 4)
    [1, 4, 2, 4, 3]
    """
\end{verbatimcode}
\bigskip
{\small
\textbf{\full: } 
\begin{verbatimcode}
from typing import List
def intersperse(numbers: List[int], delimeter: int) -> List[int]:
    result = []
    for num in numbers:
        result.append(num)
        if numbers.index(num)!= len(numbers) - 1:
            result.append(delimeter)
    return result
\end{verbatimcode}
{\noindent \color[rgb]{0.9,0.9,0.9} \rule{\textwidth}{0.5pt}}
\begin{verbatimcode}
from typing import List
def intersperse(numbers: List[int], delimeter: int) -> List[int]:
    """ Insert a number 'delimeter' between every two consecutive elements of input list `numbers'
    >>> intersperse([], 4)
    []
    >>> intersperse([1, 2, 3], 4)
    [1, 4, 2, 4, 3]
    """
    if not numbers:
        return []
    result = [numbers[0]]
    for num in numbers[1:]:
        result.extend([delimeter, num])
    return result
\end{verbatimcode}
\bigskip
\textbf{Fireworks:}\begin{verbatimcode}
from typing import List
def intersperse(numbers: List[int], delimiter: int) -> List[int]:
    """ Insert a number 'delimiter' between every two consecutive elements of input list `numbers'
    >>> intersperse([], 4)
    []
    >>> intersperse([1, 2, 3], 4)
    [1, 4, 2, 4, 3]
    """
    return [number for i, number in enumerate(numbers) for _ in range(2 - i % 2)] + [delimiter] * (int(len(numbers) / 2) if len(numbers) != 0 else 0)
\end{verbatimcode}
{\noindent \color[rgb]{0.9,0.9,0.9} \rule{\textwidth}{0.5pt}}
\begin{verbatimcode}
from typing import List
def intersperse(numbers: List[int], delimiter: int) -> List[int]:
    """ Insert a number 'delimiter' between every two consecutive elements of input list `numbers'
    >>> intersperse([], 4)
    []
    >>> intersperse([1, 2, 3], 4)
    [1, 4, 2, 4, 3]
    """
    result = []
    for i, num in enumerate(numbers):
        result.append(num)
        if i < len(numbers) - 1:
            result.append(delimiter)
    return result
\end{verbatimcode}
    }
\end{samplebox}

\begin{samplebox}{Samples from Replicate vs. Perplexity for \llama-3 70B (Instruct)}
    \textbf{Prompt:} Complete the code. Do not output anything except the completion. Start the continuation immediately.
\begin{verbatimcode}
```
from typing import List
def has_close_elements(numbers: List[float], threshold: float) -> bool:
    """ Check if in given list of numbers, are any two numbers closer to each other than
    given threshold.
    >>> has_close_elements([1.0, 2.0, 3.0], 0.5)
    False
    >>> has_close_elements([1.0, 2.8, 3.0, 4.0, 5.0, 2.0], 0.3)
    True
    """
\end{verbatimcode}
\bigskip
{\small
\textbf{Replicate: } Here is the completed code:
\begin{verbatimcode}
```
from typing import List
def has_close_elements(numbers: List[float], threshold: float) -> bool:
    """ Check if in given list of numbers are any two numbers closer to each other than
    given threshold.
    """
    numbers.sort()  # sort the list in ascending order
    for i in range(len(numbers) - 1):  # iterate over the list
        if numbers[i + 1] - numbers[i] < threshold:
            return True  # if it is, return True
    return False  # if no such pair is found, return False
```
\end{verbatimcode}
{\noindent \color[rgb]{0.9,0.9,0.9} \rule{\textwidth}{0.5pt}}
Here is the completed code:
\begin{verbatimcode}
```
from typing import List
def has_close_elements(numbers: List[float], threshold: float) -> bool:
    for i in range(len(numbers)):
        for j in range(i + 1, len(numbers)):
            if abs(numbers[i] - numbers[j]) < threshold:
                return True
    return False
```
\end{verbatimcode}
\bigskip
\textbf{Perplexity:}\begin{verbatimcode}
```
for i in range(len(numbers)):
    for j in range(i + 1, len(numbers)):
        if abs(numbers[i] - numbers[j]) < threshold:
            return True
return False
```
\end{verbatimcode}
{\noindent \color[rgb]{0.9,0.9,0.9} \rule{\textwidth}{0.5pt}}
\begin{verbatimcode}
```
for i in range(len(numbers)):
    for j in range(i + 1, len(numbers)):
        if abs(numbers[i] - numbers[j]) < threshold:
            return True
return False
```
\end{verbatimcode}
    }
\end{samplebox}

\begin{samplebox}{Samples from \nf vs. Perplexity for \llama-3.1 8B (Instruct)}
    \textbf{Prompt:} Complete the code. Do not output anything except the completion. Start the continuation immediately.
\begin{verbatimcode}
```
from typing import List
def intersperse(numbers: List[int], delimeter: int) -> List[int]:
    """ Insert a number 'delimeter' between every two consecutive elements of input list `numbers'
    >>> intersperse([], 4)
    []
    >>> intersperse([1, 2, 3], 4)
    [1, 4, 2, 4, 3]
    """
\end{verbatimcode}
\bigskip
{\small
\textbf{\nf: } simulation!!!!!!!!!!!!!!!!!!!!!!!!!!!!!!!!!!!!!!!!!!!!!!!!!!!!!!!!!!!!!!!!!!!!!!!!!!!!!!!!!!!!!!
{\noindent \color[rgb]{0.9,0.9,0.9} \rule{\textwidth}{0.5pt}}
XCTAssertEqual!!!!!!!!!!!!!!!!!!!!!!!!!!!!!!!!!!!!!!!!!!!!!!!!!!!!!!!!!!!!!!!!!!!!!!!!!!!!!!!!!!!!!!!!!!!!
\\
\bigskip
\textbf{Perplexity:}
Here's the completion of the code:
\begin{verbatimcode}
```python
from typing import List

def intersperse(numbers: List[int], delimeter: int) -> List[int]:
    """ Insert a number 'delimeter' between every two consecutive elements of input list `numbers'
    >>> intersperse([], 4)
    []
    >>> intersperse([1, 2, 3], 4)
    [1, 4, 2, 4, 3]
    """
    if not numbers:
        return numbers
    result = [numbers[0]]
    for num in numbers[1:]:
        result.extend([delimeter, num])
    return result
```
\end{verbatimcode}
{\noindent \color[rgb]{0.9,0.9,0.9} \rule{\textwidth}{0.5pt}}
Here is the completion of the code:
\begin{verbatimcode}
```python
from typing import List


def intersperse(numbers: List[int], delimeter: int) -> List[int]:
    """ Insert a number 'delimeter' between every two consecutive elements of input list `numbers'
    >>> intersperse([], 4)
    []
    >>> intersperse([1, 2, 3], 4)
    [1, 4, 2, 4, 3]
    """
    if not numbers:
        return []
    result = [numbers[0]]
    for num in numbers[1:]:
        result.extend([delimeter, num])
    return result
```
\end{verbatimcode}
    }
\end{samplebox}

\begin{samplebox}{Samples from Amazon Bedrock vs. Perplexity for \llama-3 70B (Instruct)}
    \textbf{Prompt:} Complete the code. Do not output anything except the completion. Start the continuation immediately.
\begin{verbatimcode}
```
from typing import List
def intersperse(numbers: List[int], delimeter: int) -> List[int]:
    """ Insert a number 'delimeter' between every two consecutive elements of input list `numbers'
    >>> intersperse([], 4)
    []
    >>> intersperse([1, 2, 3], 4)
    [1, 4, 2, 4, 3]
    """
\end{verbatimcode}
\bigskip
{\small
\textbf{Amazon Bedrock: } Here is the completed code:
\begin{verbatimcode}
```
from typing import List

def intersperse(numbers: List[int], delimiter: int) -> List[int]:
    result = []
    for i, num in enumerate(numbers):
        result.append(num)
        if i < len(numbers) - 1:
            result.append(delimiter)
    return result
```
\end{verbatimcode}
{\noindent \color[rgb]{0.9,0.9,0.9} \rule{\textwidth}{0.5pt}}
Here is the completed code:
\begin{verbatimcode}
```
from typing import List
def intersperse(numbers: List[int], delimeter: int) -> List[int]:
    """ Insert a number 'delimeter' between every two consecutive elements of input list `numbers`
    >>> intersperse([], 4)
    []
    >>> intersperse([1, 2, 3], 4)
    [1, 4, 2, 4, 3]
    """
    result = []
    for i in range(len(numbers)):
        result.append(numbers[i])
        if i < len(numbers) - 1:
            result.append(delimeter)
    return result
```
\end{verbatimcode}
\bigskip
\textbf{Perplexity:}\begin{verbatimcode}
```
result = []
for i, num in enumerate(numbers):
    result.append(num)
    if i < len(numbers) - 1:
        result.append(delimeter)
return result
```
\end{verbatimcode}
{\noindent \color[rgb]{0.9,0.9,0.9} \rule{\textwidth}{0.5pt}}
\begin{verbatimcode}
```
result = []
for i, num in enumerate(numbers):
    result.append(num)
    if i < len(numbers) - 1:
        result.append(delimeter)
return result
```
\end{verbatimcode}
    }
\end{samplebox}

\begin{samplebox}{Samples from Amazon Bedrock vs. Together for \llama-3 70B (Instruct)}
    \textbf{Prompt:} Complete the code. Do not output anything except the completion. Start the continuation immediately.
\begin{verbatimcode}
```
from typing import List
def has_close_elements(numbers: List[float], threshold: float) -> bool:
    """ Check if in given list of numbers, are any two numbers closer to each other than
    given threshold.
    >>> has_close_elements([1.0, 2.0, 3.0], 0.5)
    False
    >>> has_close_elements([1.0, 2.8, 3.0, 4.0, 5.0, 2.0], 0.3)
    True
    """
\end{verbatimcode}
\bigskip
{\small
\textbf{Amazon Bedrock: } Here is the completed code:
\begin{verbatimcode}
```
from typing import List

def has_close_elements(numbers: List[float], threshold: float) -> bool:
    """ Check if in given list of numbers are any two numbers closer to each other than
    given threshold.
    """
    for i in range(len(numbers)):
        for j in range(i + 1, len(numbers)):
            if abs(numbers[i] - numbers[j]) < threshold:
                return True
    return False
```
\end{verbatimcode}
{\noindent \color[rgb]{0.9,0.9,0.9} \rule{\textwidth}{0.5pt}}
Here is the completed code:
\begin{verbatimcode}
```
from typing import List

def has_close_elements(numbers: List[float], threshold: float) -> bool:
    for i in range(len(numbers)):
        for j in range(i + 1, len(numbers)):
            if abs(numbers[i] - numbers[j]) < threshold:
                return True
    return False
```
\end{verbatimcode}
\bigskip
\textbf{Together:}\begin{verbatimcode}
```
for i in range(len(numbers)):
    for j in range(i + 1, len(numbers)):
        if abs(numbers[i] - numbers[j]) < threshold:
            return True
return False
```
\end{verbatimcode}
{\noindent \color[rgb]{0.9,0.9,0.9} \rule{\textwidth}{0.5pt}}
\begin{verbatimcode}
```
for i in range(len(numbers)):
    for j in range(i + 1, len(numbers)):
        if abs(numbers[i] - numbers[j]) < threshold:
            return True
return False
```
\end{verbatimcode}
    }
\end{samplebox}
\FloatBarrier

\end{document}